\documentclass[10pt,twocolumn]{article}
\usepackage{comment}
\usepackage[nocompress]{cite}
\usepackage{amsmath,amssymb,amsfonts}
\usepackage{graphicx}
\usepackage{textcomp}
\usepackage[usenames,dvipsnames]{xcolor}
\usepackage{amsmath}
\usepackage{leftidx}
\usepackage{siunitx}
\usepackage{multirow}
\usepackage{footnote}
\usepackage[utf8]{inputenc}
\usepackage{array}
\usepackage{mdwmath}
\usepackage{mdwtab}
\usepackage{eqparbox}
\usepackage{url}
\usepackage{cite}
\usepackage{amsmath,amssymb,amsfonts}
\usepackage{algorithm2e}
\RestyleAlgo{ruled}
\usepackage{textcomp}
\usepackage{lettrine}
\usepackage[font=small]{caption}
\usepackage{subcaption}
\usepackage[export]{adjustbox}
\usepackage{leftidx}
\usepackage{siunitx}
\usepackage{multirow}
\usepackage{balance}
\usepackage{booktabs}
\usepackage{hyperref}
\usepackage{balance}

\usepackage{makecell}
\usepackage{soul}
\usepackage{censor}
\usepackage{tabularx}
\usepackage{pifont} 
\usepackage{silence}
\usepackage[
  a4paper,
  top=1.8cm,
  bottom=1.8cm,
  left=1.7cm,
  right=1.7cm,
  columnsep=0.6cm
]{geometry}

\usepackage{ragged2e}
\newcolumntype{L}[1]{%
  >{\RaggedRight\arraybackslash\hspace{0pt}}p{#1}}
\newcolumntype{C}[1]{%
  >{\Centering\arraybackslash\hspace{0pt}}p{#1}}

\newcommand{\wcell}[1]{%
  \parbox[t]{\linewidth}{\RaggedRight\hspace{0pt}#1}}

\newcommand{\cmark}{\ding{51}}
\newcommand{\xmark}{\ding{55}}

\newcolumntype{Y}{>{\raggedright\arraybackslash}X}

\renewcommand{\arraystretch}{1.00}

\newtheorem{definition}{Definition}

\def\BibTeX{{\rm B\kern-.05em{\sc i\kern-.025em b}\kern-.08em
    T\kern-.1667em\lower.7ex\hbox{E}\kern-.125emX}}

\date{}
\title{A Review Of Robotic World Models For Dynamic Environments Based On Factor And Scene Graphs}
\author{Marco Giberna, 
Miguel Fernandez-Cortizas,
Jose Luis Sanchez Lopez,
and Holger Voos}

\begin{document}

\maketitle

\begingroup
\renewcommand{\thefootnote}{}
\footnotetext{%
\thanks{Authors are with the Automation and Robotics Research Group, Interdisciplinary Centre for Security, Reliability and Trust (SnT), University of Luxembourg. Holger Voos is also associated with the Faculty of Science, Technology and Medicine, University of Luxembourg, Luxembourg.}
\textbf{Acknowledgment:}
This work was partially funded by the Fonds National de la Recherche of Luxembourg (FNR) through the DEUS project (Ref. C22/IS/17387634/DEUS) and the project PCS-GRAPHS (Ref. C25/IS/19613768/PCS-GRAPHS).}
\addtocounter{footnote}{-1}
\endgroup

\begin{abstract}
Models based on graphs have emerged in robotics as a powerful foundation for internal world representations, where factor and scene graphs are among the most prominent model types found in the related literature and in successful robotic solutions. Initially, many of these models were assuming static environments as a simplification.
Herein, factor graphs mainly provide uncertainty-aware geometric estimations while scene graphs enable a structured semantic abstraction.  However, real-world robotic environments are often dynamic, posing severe challenges for purely static world representations. Therefore, this review presents a comprehensive view on how dynamic aspects of real-world environments can be addressed in such graph-based world models. 
We organize our assessments around three main aspects: (I) suitable representations, (II) pipelines to construct and update the representations, and (III) their exploitation for downstream tasks. 
We review approaches that are either based on factor or scene graphs, but put special emphasis on novel approaches that combine both types to form hybrid models. We mainly analyze how different types of dynamics can be modeled herein, and categorize common architectural patterns. 
Finally, emerging trends and open challenges are identified, including uncertainty propagation from learned perception through the representation layers, the observability of dynamic-entity motion and scale under minimal sensing, scalable lifelong maintenance, and the lack of datasets and evaluation protocols that ground world-model quality in downstream task performance under dynamics.
\end{abstract}

\section{Introduction}\label{sec:introduction}
Robotic autonomy relies on the ability to construct, maintain and exploit internal representations of the world.
These representations serve as a robot's mental map, enabling the grounding of perception into action and supporting decision-making and action generation to achieve desired tasks. 
World representations are necessary for mobile robots, even if they only come in a simple static form. Geometric maps such as point clouds, occupancy grids, or meshes provide the spatial substrate required for localization, navigation, and collision-free motion planning \cite{luca_carlone_slam_2025}.
Without such representations, even basic tasks like obstacle avoidance and exploration in any environment would be infeasible.

Beyond low-level geometry, many robotic applications require richer and more structured representations. 
Mobile manipulation and object retrieval, for instance, require identifying not only where the robot is, but also \textit{what} objects exist, \textit{where} they are located, and \textit{how} they relate to one another (e.g. a human \textit{sitting on} the chair \textit{next to} the table \textit{in} the kitchen) \cite{bavle_slam_2023}.
Similarly, long-horizon task planning and human-robot interaction depend on representations that support abstraction and reasoning, aligning the robot and human's understanding of the environment. 
Commands expressed in natural language cannot be grounded directly in metric maps, but require semantic entities, relational structure, and symbolic reasoning over space \cite{maggio_clio_2024} \cite{rana2023sayplangroundinglargelanguage}.
Models based on graph structures provide an expressive framework for constructing such required world representations. In particular, factor graphs and scene graphs have emerged as the two most prominent, yet complementary graph-based approaches that bridge perception, estimation, and reasoning. 

Initially, based on the insight that probabilistic approaches were much better suited to cope with the uncertainty inherent in robotic applications than deterministic ones, Probabilistic Graphical Models (PGMs) and especially factor graphs became very successful in robot perception and especially in simultaneous localization and mapping (SLAM) \cite{luca_carlone_slam_2025}.  
They offer a very suitable mathematically grounded uncertainty-aware representation of the environment, explicitly modeling the dependencies between uncertain robot states, landmarks, objects and noisy sensor measurements into a unified PGM, see e.g.,  \cite{dellaert_factor_2021}, \cite{dellaert_factor_2017}. Their formulation enables scalable inference through optimization, making them well suited for common robotic tasks such as SLAM, sensor fusion in the presence of noise and partial observability \cite{dellaert_factor_2017}, but also probabilistic approaches for planning and control \cite{mukadam2018continuous}. 
Scene graphs, on the other hand, emphasize structure and semantics by representing the environment as entities (nodes, subjects, and objects) and relations (edges, predicates). They capture spatial, functional, and semantic relationships at multiple levels of abstractions by inferring concepts from the raw perception \cite{chang_comprehensive_2023}, \cite{catalano_3d_2025}.
Recent works such as \cite{hughes_hydra_2022}, \cite{bavle_s-graphs_2025} 
increasingly combine these paradigms, yielding hybrid representations in which factor graphs provide a probabilistic mathematical model of the geometric aspects of the environment while scene graphs support semantic abstraction and reasoning in parallel.

Factor and scene graphs were originally widely adopted in static robotic world representations. However, those are fundamentally limited in real-world settings, as most environments for mobile robot applications such as homes, offices, construction sites, or warehouses, are inherently dynamic \cite{wang_-depth_2025}, \cite{bavle_slam_2023}. 
Objects move, humans interact with and alter the environment, furniture is rearranged, doors open and close, and entire structures evolve over time.
All of these dynamic processes can be organized across different temporal scales, ranging from  continuously moving agents to long-term structural changes, as described later in more detail. 
Static maps implicitly assume permanence and consistency, making them less accurate when employed in situations presenting persistent motion or long-term changes \cite{bescos_dynaslam_2018}. 
This mismatch leads to, for example, a decreasing localization accuracy, incorrect data association, spurious landmarks, and eventually the failure in downstream tasks.
Moreover, static representations cannot capture temporal phenomena such as object lifetimes, motion patterns, or recurring routines, essential for future state prediction and long-term robot autonomy \cite{catalano_aion_2025}, \cite{schmid_khronos_2024}.

These limitations have motivated a growing body of research on dynamic world representations, where time, changes, and uncertainty are paramount and the center of attention, and also led to  related extensions of models based on factor and scene graphs. They enable robots not only to estimate \textit{what is currently true}, but also to reason about \textit{what has changed}, \textit{what persists}, and \textit{what is likely to happen next} from a lower geometric and uncertainty grounded space up to higher semantic and abstracted layers\cite{schmid_khronos_2024} \cite{morris_dynosam_2025}. As the number of such solutions is currently quickly expanding, this review aims to organize this landscape by addressing three main research questions:
\begin{enumerate}
    \item \textbf{Which representations (world models) based on factor and scene graphs do exist for robotic autonomy in dynamic environments?} \\
    We analyze the strengths and limitations of the usage of factor graphs, scene graphs and especially hybrid representations with respect to a taxonomy of dynamic aspects of the environment.
    \item \textbf{How can such representations be constructed and maintained?}\\
    We review methods for perception, data association, temporal modeling, uncertainty handling, and inference using graph-based approaches, highlighting how different works generate dynamic world models from sensor data.
    \item \textbf{How can such representations be exploited?}\\
    We examine how dynamic graph-based world models can be used for downstream tasks such as navigation, manipulation, task planning, predictive interaction and semantic reasoning, and how evaluation metrics can be used for an assessment. 
\end{enumerate}
By organizing and assessing existing works around these questions, this review provides a comprehensive perspective on spatio-temporal scene modeling through the specific use of graph-based representations, namely factor graphs and scene graphs, showcasing current trends and identifying open challenges, to provide a foundation for future research in long-term, robust robotic situational awareness in dynamic environments \cite{bavle_slam_2023}. 
This review is not only a catalog of existing approaches, but also discusses a conceptual combination of factor and scene graphs for dynamic world modeling. We intentionally organize prior work through the lens of representational choices, inference structure, and temporal dynamics.

\section{Background and Related Works}\label{sec:background}
\subsection{Mobile Robot Graph-Based World Models}
In this section we first briefly summarize the use of factor graphs, scene graphs and hybrid forms of both for robotic world modeling. We then define \textit{dynamics} in the context of robotics and finally provide an overview of existing surveys and reviews of the use of factor and scene graphs, particularly for modeling \textit{dynamic} robot environments, and identify research gaps herein, which motivate this review article. 

\subsubsection{Factor Graphs}\label{sec:background_factor_graphs}
\begin{figure}
    \centering
    \includegraphics[width=\columnwidth]{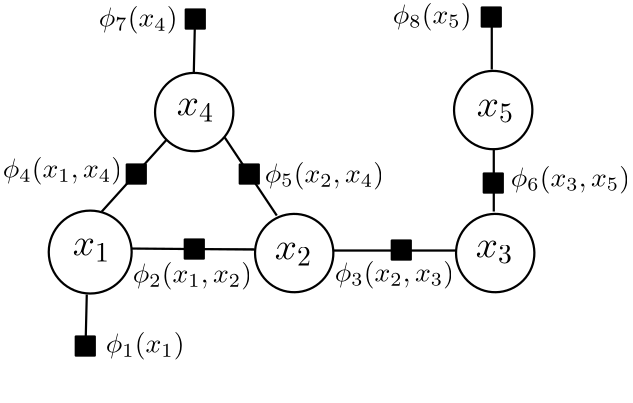}
    \caption{Factor graph example.}
    \label{fig:FG_example}
\end{figure}

Factor graphs have emerged as a primary mathematical framework mainly for geometric world models, providing a flexible approach for integrating various sensor data and prior knowledge into a unified, probabilistic optimization problem\cite{deeb_model-based_2019} \cite{gulati_robust_2017}. In general, a factor graph describes the structure of a factorization of a function. For instance, 
consider a function $J(X)$ of a set of variables $X = \{x_1, \ldots , x_n\}$, that factors into a product of $r$ local functions or ``factors'' $\phi_j(X_j)$ of a subset $X_j$ of $X$, respectively:
\begin{equation}
J(X) = \prod_{j=1}^{r} \phi_j(X_j)
\end{equation}
A factor graph $G = \{X,\mathcal{F},\mathcal{E}\}$ is an undirected bipartite graph that consists of  the set of variable nodes $X$, typically depicted as circles, and the set of factor nodes $\mathcal{F}$, typically depicted as squares. The set $\mathcal{E}$ involves the edges connecting variable nodes and factor nodes. The example in Figure \ref{fig:FG_example} shows a factor graph with
5 variables $X = \{x_1, \ldots, x_5 \}$ and 8 factors $\mathcal{F}=\{\phi_1, \ldots, \phi_8 \}$,  representing the factorization 
\begin{equation}
\begin{aligned}
    J(X) &= \prod_{j=1}^{8} \phi_j(X_j) \\
    &= \phi_1(x_1)\cdot\phi_2(x_1,x_2) \cdot \phi_3(x_2,x_3) \cdot \\
    & \quad \phi_4(x_1,x_4) \cdot \phi_5(x_2,x_4) \cdot \phi_6(x_3,x_5) \cdot \\
    & \quad \phi_7(x_4) \cdot \phi_8(x_5)
    \end{aligned}
\end{equation}
In a probabilistic formulation, factor graphs can be used as a PGM representing the factorization of a joint probability distribution over state variables \cite{gulati_robust_2017}. Within probabilistic robotic world modeling, a factor graph encodes all forms of uncertainty, such as for instance those arising from noisy sensor measurements. In this case, the variables $X$ represent quantities to be estimated, such as  robot poses ($\mathit{X}_k$), static map points ($\mathit{M}_j$) or, crucial in dynamic settings, objects and agent poses ($\mathit{L}_{k,t}$) and motion transformations ($\mathit{H}_{k,t}$)\cite{bescos_dynaslam_2021} \cite{deeb_piecewise-deterministic_2022} \cite{morris_dynosam_2025}. The factors $\phi$ encode constraints or measurements, such as odometry ($\mathit{\phi}_{odom}$), visual reprojection errors ($\mathit{\phi}_{repr}$), point-to-plane ($\mathit{\phi}_{ICP}$), structural priors ($\mathit{\phi}_{pose-plane}$, $\mathit{\phi}_{wall-room}$) or semantic motion priors ($\mathit{\phi}_{mot}$)  \cite{morris_dynosam_2025} \cite{gulati_robust_2017} \cite{bescos_dynaslam_2021} \cite{giberna_dynemo-slam_2025}. 
In the example in Figure \ref{fig:FG_example}, $x_1, x_2, x_3$ might be robot poses, while $x_4, x_5$ might represent two landmark poses. The factors $\phi_1, \phi_7, \phi_8$ represent pose priors, $\phi_2, \phi_3$ could be odometry measurements, and $\phi_4, \phi_5, \phi_6$ represent measurements of the landmark poses from the related robot poses, respectively.

The strength of factor graphs lies in their capacity to combine diverse types of constraints while explicitly handling uncertainty and measurement noise \cite{brink_using_2016}.
Variables are typically modeled under the assumption of zero-mean Gaussian noise, where the information matrix (i.e., the inverse of the covariance matrix) determines the weight of each factor. Factor graphs are often exploited (e.g., in SLAM) to estimate the most likely value of all variables using Maximum A Posteriori (MAP) inference in form of nonlinear least-squares optimization \cite{deeb_model-based_2019}.
\subsubsection{Scene Graphs}\label{sec:background_scene_graphs}
\begin{figure}[b]
    \centering
    \includegraphics[width=\columnwidth]{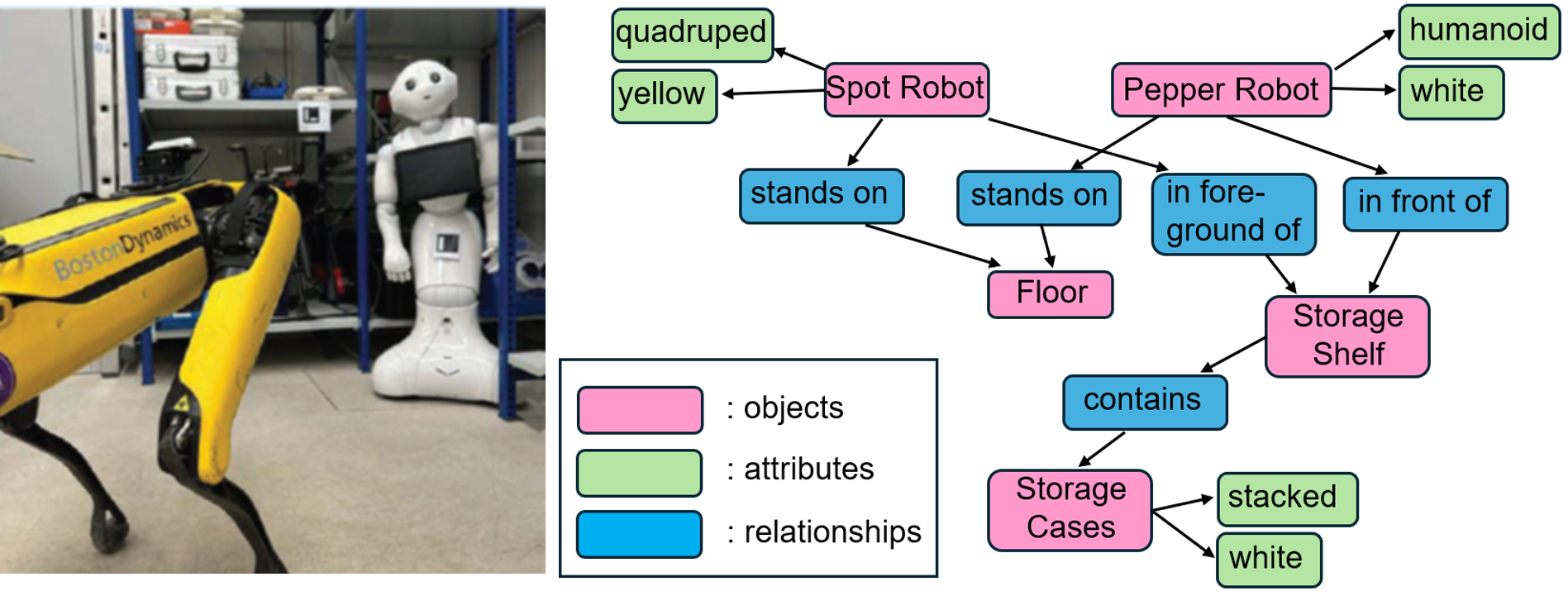}
    \caption{Example image of a scene (left) and related scene graph (right).}
    \label{fig:SG_example}
\end{figure}

Scene Graphs were originally invented in computer vision, based on the intention to describe the detailed semantics of a scene shown in a digital image. A scene graph encodes object instances, attributes of objects, and relationships between objects \cite{Johnson_2015_CVPR}. 
This simple formulation is powerful enough to describe visual scenes in great detail because it places no restriction on the types of objects, attributes, and relationships that can be represented. Figure \ref{fig:SG_example} shows an example of a scene graph (right) of a captured  image (left). This example illustrates that object instances may be physical objects (``robot''), locations (``floor''), things (``storage case'') or parts of other objects. Attributes can describe color (``robot is yellow''), type (``quadruped''), shape or pose. Relationships describe the connections between pairs of objects and can encode geometry (``in front of'') or positions, actions (``stands on''), or object parts (``shelf contains cases''). These relationships are usually expressed as $\langle$ \textit{subject - predicate - object} $\rangle$ triplets. 

Formally, given a set of object classes $\mathcal{C}$, a set of attribute types $\mathcal{A}$, and a set of relationship types $\mathcal{R}$, a scene graph $G$ is defined as a tuple $G = (\mathcal{O}; \mathcal{E})$ where $\mathcal{O} = \{o_1; \ldots ; o_n\}$
is a set of objects and $\mathcal{E} \subseteq \mathcal{O} \times \mathcal{R} \times \mathcal{O}$ is a set of edges.  Each object has the form $o_i = (c_i; A_i)$ where $c_i \in \mathcal{C}$ is the class of the object and $A_i \subseteq \mathcal{A}$ are the attributes of the object\cite{chang_comprehensive_2023}. 

The generation of scene graphs is mainly based on machine learning approaches and can be 
roughly divided into three parts: feature extraction, contextualization, and graph construction and reasoning \cite{li2024scene}. Herein, the process of feature extraction is mainly responsible for encoding objects or object pairs in the image. Contextualization plays the role of associating different
entities and is mainly used to enhance the contextual information between entities. Finally, 
these contextual information are used to predict the predicate and complete the construction and inference of the graph. There are currently two main types of scene graph generation \cite{li2024scene}, \cite{chang_comprehensive_2023}. The top-down methods comprise two sequential stages, namely object detection and pairwise relationship recognition, while the bottom-up method involves jointly detecting and recognizing the objects and their relationships \cite{adaimi2023composite}. 

As scene graphs can abstract objects and the relationships between them in a more explicit and easier-to-understand way, they became quite popular for visual scene understanding in general computer vision, but also in robotics. The conceptual shift from dense, low-level metric maps towards scene graphs represents a significant advancement in robotic world modeling. In robotics, scene graphs offer an abstract, layered representation that bridges the gap between the raw geometric perception and high-level symbolic reasoning \cite{bavle_slam_2023}. By decomposing an environment into a scene graph structure robots can move beyond flat representations, such as occupancy grids or point clouds, to models that capture the semantic and relational architecture of the world.  

As scene graphs might rapidly grow in real-world complex environments, the literature distinguishes between two primary structural paradigms: 
\begin{itemize}
    \item Flat Scene Graphs: these models treat all entities equally at a single level of representation, focusing on object-level nodes and their direct relationships \cite{catalano_3d_2025}. Approaches such as ConceptGraphs\cite{gu_conceptgraphs_2023} and SceneGraphFusion\cite{wu_scenegraphfusion_2021} utilize this approach to manage open-vocabulary object recognition and incremental updates in cluttered scenes where clear structural hierarchies may be emergent rather than predefined. 
    \item Hierarchical Scene Graphs: these represent the environment across multiple levels of abstraction, grouping lower level elements into semantically richer containers, e.g. objects within rooms, rooms within buildings\cite{rosinol_3d_2020} \cite{bavle_s-graphs_2025}. This multi-resolution approach is crucial for scalability in large-scale environments, as it allows robots to perform long-term planning and inference over compact abstractions rather than dense data \cite{catalano_3d_2025}.  
\end{itemize}
Hierarchical frameworks like Hydra \cite{hughes_hydra_2022} or S-Graphs\cite{bavle_s-graphs_2025} adopt a bottom-up stack comprising distinct layers specifically for indoor environments: (1) Metric-Semantic Mesh, Pointcloud, (2) Objects and Agents, (3) Places and Structures, (4) Rooms and Floors, and (5) Building. An example that extends these works to outdoor scenes is  \cite{greve_collaborative_2024} that proposes a hierarchical scene graph of a traffic scenario comprising 6 layers: (1) Keyframes \& Point clouds, (2) Lane Graph, (3) Vehicles, (4) Landmarks, (5) Roads \& Intersections, and (6) Environment.

\begin{figure}[ht]
    \centering
    \includegraphics[width=\columnwidth]{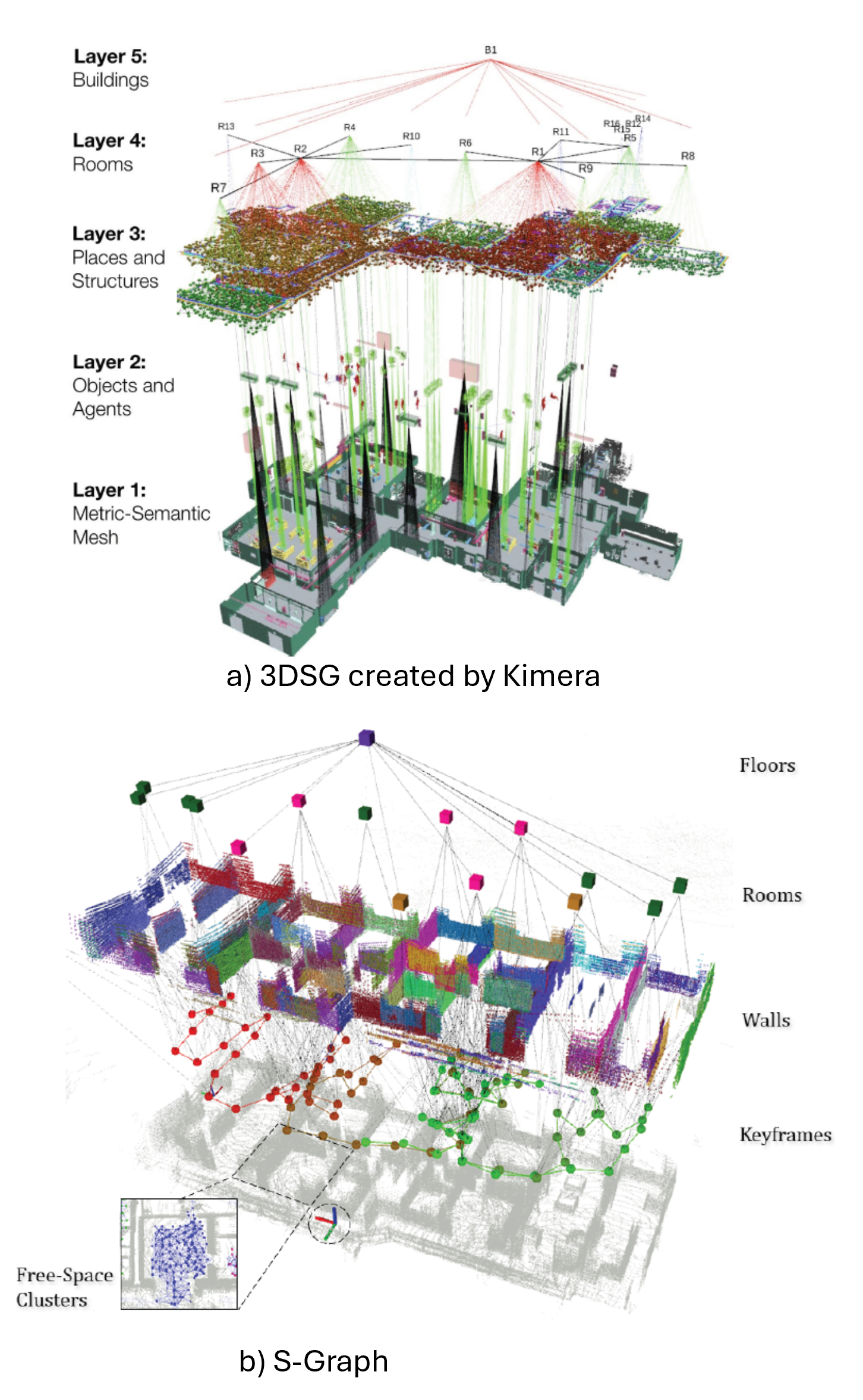}
    \caption{Example of 3DSGs created by (a) Kimera \cite{rosinol_kimera_2021} and (b) S-Graphs \cite{bavle_situational_2022}.}
    \label{fig:Examples_3DSG}
\end{figure}

As real-world scenes are three-dimensional, the idea of scene graphs has been extended to 3D space, leading to 3D scene graphs (3DSG) \cite{armeni_3d_2016} \cite{kim_3-d_2020}. 
Edges in a 3DSG encode the connections within the world model through several relationship types, such as: 
\begin{itemize}
    \item Spatial Relations: These include geometric constraints such as \textit{on}, \textit{in}, \textit{adjacent to} and \textit{support} \cite{catalano_3d_2025} \cite{yan_dynamic_2025}.
    \item Visibility and Proximity: \textit{Co-visibility} edges link objects seen simultaneously, while \textit{close-to} edges often use nearest-neighbourg k-d tree to define spatial proximity \cite{behrens_lost_2025}.
    \item Functional Relations: a newer class of edges representing operational dependencies. OpenFunGraph models these as directed edges between ``interactive elements'', such as switches and handles, and objects that they influence, e.g. \textit{controls} a light, \textit{opens} a cabinet \cite{zhang_open-vocabulary_2025}.
    \item Affordances: attributes or edges that define possible interactions which can change the entities states, such as \textit{sit on}, \textit{opening}, \textit{pulling}, providing the robot with actionable knowledge \cite{catalano_3d_2025}.
    \item Containment and Affiliation: Inter-layer edges such as \textit{part of}, \textit{located in} or \textit{contains} facilitate hierarchical queries \cite{catalano_3d_2025} \cite{behrens_lost_2025}.
\end{itemize}
This structured abstraction allows robots to move beyond simple obstacle avoidance, static mapping and localization, towards actionable spatial perception, where the world model serves as a persistent, queryable memory for complex multi-step tasks in evolving robot-human shared spaces. Two examples of a 3DSG are shown in Figure \ref{fig:Examples_3DSG}.

\subsubsection{Hybrid Models using Factor and Scene Graphs}\label{sec:background_hybrid}
A scene graph answers \textit{``what exists and how it relates''}, while a factor graph answers \textit{``where it is, and how certain, consistent it is with sensor measurements''}. Therefore, it seems promising to combine these two paradigms to a more comprehensive world representation. Recent robotics 3DSG solutions explicitly include a pose-graph layer as the uncertainty-bearing backbone, and the integration of the two paradigms creates a multi-resolution framework which bridges metric-semantic perception with high-level symbolic reasoning \cite{rosinol_kimera_2021} \cite{bavle_situational_2022}.
In this hybrid paradigm, factor graphs serve as the probabilistic geometric backbone, while scene graphs provide the relational architecture necessary for task-level autonomy in dynamic environments \cite{catalano_3d_2025} \cite{rosinol_3d_2020}.

The literature converges to a small set of recurring architectural patterns for coupling the two paradigms, each representing a different trade-off between integration tightness and modularity.
The most prevalent pattern is the \textbf{layered stack with factor graph backbone}: the factor graph simultaneously optimizes the robot's trajectory and the metric-semantic mesh, whose globally and temporally consistent output is then fed to a separate scene graph parser that builds the relational structure on top. 
Hydra \cite{hughes_hydra_2022} and Kimera \cite{rosinol_kimera_2021} support this approach, where a dedicated parsing thread incrementally constructs the scene graph from the optimized metric layer. 
The advantage is modularity, where the scene graph logic can be developed independently of the SLAM back-end, but the disadvantage is a one-directional information flow: the scene graph is informed by the factor graph, but semantic information rarely propagates back to improve localization.

A tighter alternative is the \textbf{single unified optimizable graph}, in which scene graph entities are incorporated directly as variables and factors within the same factor graph that optimizes robot poses. 
S-Graphs \cite{bavle_s-graphs_2025} and its visual counterpart vS-Graphs \cite{tourani2026vs} introduced this pattern, merging SLAM and scene graph into a single multi-layered jointly optimizable factor graph where walls, rooms, and floors appear as semantic variables constrained by geometric factors.
This enables bidirectional information flow: geometric observation constrain semantic entity positions, but semantic structural priors in turn constrain and regularize ego-trajectory estimation, demonstrably reducing drift compared to purely geometric approaches.

A third pattern, the \textbf{loosely coupled parallel architecture}, maintains the factor graph and scene graph as separate optimization problems connected only through shared node positions, avoiding the computational cost of joint optimization at the expense of weaker consistency guarantees \cite{rosinol_3d_2020}.

\subsection{Dynamics in the Context of Robotics}\label{sec:background_dynamics_definition}
It is necessary to fix what we mean by \textit{dynamics}, since the term is used inconsistently across robotics and adjacent fields. 
In classical mechanics, dynamics concerns forces and torques and their effect on motion; in dynamical system theory, it describes the time-evolution of a point in a geometric state space; in computer science, a ``dynamic`` data structure refers to memory organization rather than physical change. 
Within mobile robotics, the terms is used more narrowly and typically refers to external, uncontrolled moving entities, such as pedestrians and vehicles, that a robot must perceive and account for during localization and mapping\cite{kucner_survey_2023}. 

This robotics-specific usage, however, is itself too narrow for our purpose.
Hahnel et al. \cite{hahnel2003map} offered one of the first attempts to define dynamics as an environmental property, stating that an environment is dynamic if it changes over time or if moving objects are present within it. 
This definition is useful but indirect, and, being framed around discrete macroscopic objects, does not make explicit how motion that is never directly observed, e.g. a chair found in a new location, should be treated. Therefore, we adopt the definition of dynamics given in \cite{kucner_survey_2023}:
\vspace{6pt}
\begin{definition}[Dynamics]
\label{def:dynamics}
    Dynamics are any directly or indirectly perceivable motions.
\end{definition}
\vspace{6pt}
This formulation is deliberately broad. \textit{Direct} perceivability covers motion observed as it occurs given an appropriate sensor.
\textit{Indirect} perceivability covers semi-static entities, such as relocated furniture, whose motion is never itself observed but whose resulting change of state is.
This distinction between direct and indirect perceivability, rather than any specific list of entity types or timescales, is what we take to be the essential content of Definition \ref{def:dynamics}, and it is this distinction, not a fixed taxonomy, that Kucner et al. \cite{kucner_survey_2023} leave open for further specialization within a given application domain.

Definition \ref{def:dynamics} tells us \textit{what} qualifies as dynamics. It does not by itself prescribe how a world model should organize or respond to difference instances of it.
In Section \ref{sec:taxonomy_dynamics}, we build on this definition to introduce a taxonomy of dynamics tailored to graph-based robotic world models, organized primarily around the temporal horizon at which a perceivable motion or change occurs.

\subsection{Assessment of Previous Surveys and Reviews}
Probabilistic Graphical Models provide the foundational framework for modeling large-scale inference problems in robotics, with factor graphs serving as a bipartite representation where variables denote unknown states and factors represent local constraints derived from sensor data \cite{dellaert_factor_2021} \cite{dellaert_factor_2017}.
Traditional reviews trace the maturity of SLAM from early formulations based on Kalman filters and Bayesian estimation toward modern graph-based optimization paradigms \cite{mascaro_scene_2025} \cite{luca_carlone_slam_2025}.
Systematic surveys of spatial perception frameworks have categorized scene modeling techniques into three primary layers: metric, metric-semantic, and metric-semantic-topological \cite{mascaro_scene_2025}. 
This highlights a trend shifting away from purely geometric representations like point clouds or meshes, toward advanced data structures that incorporate high-level concepts such as instances and places \cite{bavle_slam_2023}.

Complementary to reviews of metric frameworks, existing reviews of scene graphs define them as structured representations that encode semantic entities and their interrelationships to facilitate human-like reasoning \cite{chang_comprehensive_2023} \cite{amuche_advancements_2025} \cite{lai_review_2022}. 
Early literature in this domain primarily focused on offline image retrieval and 2D scene graph generation, though more recent reviews highlight advancements in 3D scene graphs capable of capturing metrics, semantics and objects (also including dynamic agents) within a unified actionable model \cite{chang_comprehensive_2023} \cite{lai_review_2022}.

Notably, these two types of reviews have developed largely independently: reviews of factor graphs assess uncertainty-aware geometric estimation, while reviews of scene graphs assess semantic and relational abstraction, but very few works survey both representations \emph{jointly} as complementary layers of a single world model. A first recent attempt in this direction is the review of Catalano et al. \cite{catalano_3d_2025}, which focuses on works that combine geometry, semantics, and action within unified representations; however, this overview  does not center on dynamics, nor does it systematically analyze the coupling patterns, uncertainty propagation, or construction pipelines that arise when factor and scene graphs are combined specifically to handle non-stationary environments.

The landscape of SLAM in dynamic environments has been analyzed as a progressive evolution from purely geometric methods to advanced semantic-based paradigms \cite{wang_survey_2024} \cite{chen_semantic_2025} \cite{wen_semantic_2021}.
Existing surveys categorize solution approaches into those that maintain localization stability by excluding dynamic elements as outliers and those that perform active Simultaneous Localization, Mapping and Moving Object Tracking (SLAMMOT) \cite{wang_-depth_2025} \cite{xu_survey_2021}. 
Furthermore, dedicated reviews on feature requirements in dynamic SLAM weight the trade-offs between low-level primitives, such as points or lines, and high-level semantic labels, suggesting that robust robotic autonomy may require the integrated usage of both feature levels \cite{xu_survey_2021}.
These dynamic-SLAM surveys, however, largely set aside the scene-graph literature, and consequently do not address how semantic-relational structure should be maintained under dynamics, nor how it interacts with the probabilistic backbone of the factor graph.

A separate line of work addresses dynamics as a phenomenon in its own right, independent of any specific representation. Kucner et al.\cite{kucner_survey_2023} survey Maps of Dynamics (MoDs) and propose a general definition and characterization of dynamics for mobile robotics, which we adopted and extended in Section \ref{sec:background_dynamics_definition} and organize into a temporal-horizon taxonomy in Section~\ref{sec:taxonomy_dynamics}. 
While their survey provides a rigorous treatment of \textit{what} dynamic phenomena are and how they can be statistically modeled over space and time, it is agnostic to the underlying world-model representation, and does not address factor graphs, scene graphs, or their hybrid forms.

Despite the depth of reviews in individual subfields, critical gaps in the existing review or survey literature can be identified.  
Key surveys often focus on isolated areas like perception or planning while overlooking the practical integration of these components \cite{khan_foundation_2025} \cite{bavle_slam_2023}.
Gaps also remain in analyzing the long-term usability of maps in complex dynamic scenes and handling the non-rigid deformations inherent in highly dynamic environments \cite{chen_semantic_2025} \cite{luca_carlone_slam_2025}.
Taken together, these limitations reveal a fragmented landscape in which probabilistic estimation (factor graphs), semantic scene understanding (scene graphs), and the characterization of dynamics have each matured in relative isolation, with limited cross-pollination between the three. 
To the best of our knowledge, no existing survey or review jointly addresses the full pipeline from probabilistic spatio-temporal estimation to structured semantic reasoning especially in \textit{dynamic} environments, nor does any work systematically analyze how these two paradigms can be combined to achieve robust, long-term or even lifelong situational awareness.

This review addresses these gaps directly.
We first establish a shared characterization of dynamics (Section~\ref{sec:taxonomy_dynamics}), building on and extending the definition of Kucner et al. \cite{kucner_survey_2023} and based on the surveyed literature, and then assess existing works with respect to three questions: which graph-based representations are suited to dynamic environments (Section ~\ref{sec:dyn_world_model_representations}), how they are constructed and maintained (Section ~\ref{sec:construction_pipeline}), and how they are exploited for downstream tasks (Sections~\ref{sec:querying_reasoning_action}--\ref{sec:applications}). 
Additionally,we present the commonly used evaluation protocols, datasets, and metrics for tesing in ~\ref{sec:evaluation_metrics}, and a comprehensive comparison of existing related works in ~\ref{sec:taxonomy}. Finally, we conclude this review by highlighting the open challenges identified and outlining future research directions in ~\ref{sec:challenges_future_works}. 
In doing so, this review provides a unified perspective on the synergy between factor graphs and scene graphs as complementary and mutually reinforcing tools for long-term robotic autonomy under dynamics.

\section{Taxonomy of Dynamic Aspects in Robotic World Models}\label{sec:taxonomy_dynamics}
Before analyzing how factor graphs and scene graphs represent, construct and exploit non-stationary environment, we first set the vocabulary this review adopts to describe \textit{dynamics themselves}. 
The literature is not consistent here: the same underlying phenomenon, e.g., a chair being moved between visits, is variously described as a \textit{semi-static change} \cite{gomez_object-based_2020}, a \textit{movable} object \cite{brink_using_2016} or a node with a \textit{persistence} attribute\cite{behrens_lost_2025}.
Establishing a shared characterization upfront allows us to classify representations, pipelines, and applications along consistent axes for the remainder of the review.

We organize dynamics along three axes: 
\emph{when} a change occurs relative to observation (Temporal Horizon),
\emph{what} kind of entity undergoes it (Entity Typology),
and \emph{how} the change manifests in the representation (State Expression).
A fourth property, the \emph{predictability} of a dynamic entity's behavior, is briefly noted at the end of this section since it is only sparsely addressed by the surveyed literature.

\subsection{Primary Axis: Temporal Horizon}\label{sec:taxonomy_temporal_horizon}

\begin{table}[t]
\centering
\scriptsize
\setlength{\tabcolsep}{3pt}
\renewcommand{\arraystretch}{1.15}
\caption{Primary temporal-horizon axis used to organize dynamics in graph-based robot world models.}\label{tab:temporal_horizon_taxonomy}
    \begin{tabularx}{\columnwidth}{@{}p{0.15\columnwidth} p{0.12\columnwidth} X@{}}
    \toprule
    \textbf{Tier} & \textbf{Timescale} & \textbf{Description} \\
    \midrule
    Short-Term Dynamics 
    & Continuous / Now 
    & Independently moving entities whose pose changes during observation, such as humans, vehicles, or robots \\ 
    \midrule
    Semi-Static Changes 
    & Minutes--Days 
    & Entities that are static during a single observation but may be relocated, added, removed, or change state between observations \\ 
    \midrule
    Long-Term Structural Changes 
    & Days--Years 
    & Persistent modifications to the built environment itself, such as layout changes, renovations, or seasonal structural variations  \\
    \bottomrule
    \end{tabularx}
\end{table}

The dimension that structures most consistently both individual models and this review is \textit{when} a change occurs relative to the robot's observation stream, since it directly governs the memory allocation, data association strategy, and update frequency a world model must support. 
We distinguish three tiers:

\begin{itemize}
    \item \textbf{Short-Term Dynamics} (continuous motion, now): humans, moving vehicles, and other independently moving entities whose pose changes within the timescale of a single operating session, requiring real-time tracking and motion-aware data association \cite{bescos_dynaslam_2021} \cite{morris_dynosam_2025} \cite{schmid_khronos_2024}.
    \item \textbf{Semi-Static Changes} (minutes to days): objects that are stationary during any single observation but relocated, added, removed, or state-transitioned between observations, e.g., furniture rearrangement or a door's open/closed state \cite{gomez_object-based_2020} \cite{giberna_dynemo-slam_2025} \cite{schmid_khronos_2024}.
    \item \textbf{Long-Term Structural Changes} (days to years): modifications to the built environment itself, e.g., renovations or seasonal layout shifts, requiring inter-session alignment and lifelong map maintenance \cite{walcott-bryant_dynamic_2012} \cite{kim2022lt}.
\end{itemize}
This tiering is not merely descriptive: it maps directly onto the mechanism a world model must implement, from the tightly-coupled SLAMMOT formulations used for Short-Term Dynamics, through the persistence and containment reasoning used for Semi-Static Changes, to multi-session change detection used for Long-Term Structural Changes. 
We adopt this three-tier distinction as the primary organizing axis for the remainder of the paper.
A summary is depicted in Table \ref{tab:temporal_horizon_taxonomy}.

\subsection{Secondary Axes: Entity Typology and State Expression}\label{sec:taxonomy_secondary_axes}
Within each temporal tier, two further properties are useful for characterizing \textit{what} is changing and \textit{how} that change is expressed in the graph. 
Table \ref{tab:dynamics_taxonomy} summarizes both.

\textbf{Entity Typology} distinguishes dynamic elements by the structural assumption their motion model requires. 
\textit{Rigid} entities undergo $SE(3)$ pose change with constant geometry and dominate the surveyed factor-graph formulation \cite{bescos_dynaslam_2021} \cite{qiu2022airdos}.
\textit{Articulated} entities are constrained by internal kinematic chains, relevant both to manipulable objects such doors and drawers \cite{zhang_open-vocabulary_2025} and to parametric human body models \cite{henning_bodyslam_2022} \cite{henning_bodyslam_2023}.
\textit{Deformable} entities, whose shape or topology changes non-rigidly, remain largely outside the scope of the probabilistic graph-based systems surveyed here.

\textbf{State Expression} distinguishes \textit{how} a change manifests once represented. 
\textit{Geometric-kinematic} changes are variations in position, velocity, or boundary and are the traditional target of factor graph estimation.
\textit{Semantic-affordance} changes affect an entity's functional or operational state, e.g., a light turning on or off, and are typically encoded as node attributes in 3D scene graphs \cite{catalano_3d_2025} \cite{olivastri_multi-modal_2024}.
\textit{Topological} changes modify the connectivity of the environment itself, e.g., a closed door removing the traversable edge between two rooms, and are of particular relevance to hierarchical scene graph maintenance \cite{catalano_3d_2025}.

\begin{table}[t]
\centering
\scriptsize
\setlength{\tabcolsep}{3pt}
\renewcommand{\arraystretch}{1.15}
\caption{Secondary axes used to characterize dynamic entities within each
temporal tier.}\label{tab:dynamics_taxonomy}
    \begin{tabularx}{\columnwidth}{@{}l l X@{}}
    \toprule
    \textbf{Axis} & \textbf{Category} & \textbf{Description} \\
    \midrule
    \multirow{3}{*}{\shortstack[l]{Entity\\Typology}}
     & Rigid       & $SE(3)$ pose change, constant geometry \\
     & Articulated & Kinematic chain, constrained internal DoF \\
     & Deformable  & Non-rigid shape/topology change \\
    \midrule
    \multirow{3}{*}{\shortstack[l]{State\\Expression}}
     & Geometric-Kinematic  & Position, velocity, boundary change \\
     & Semantic-Affordance  & Functional/operational state change \\
     & Topological          & Connectivity change in the scene graph \\
    \midrule
    \midrule
    \multirow{3}{*}{\shortstack[l]{Behavioral\\Intent}}
     & Deterministic  & Motion strictly governed by fixed laws \\
     & Physics-Based  & Motion governed by physics-based probabilities \\
     & Goal-Driven    & Motion governed by internal decisions or social compliance \\
    \bottomrule
    \end{tabularx}
\end{table}

\subsection{Adopted Characterization}\label{sec:taxonomy_adopted_characterization}
We adopt \textbf{Temporal Horizon} as the primary organizing dimension of this review, using the three tier distinction to structure this paper's analysis.
Entity Typology and State Expression are used as secondary, descriptive dimensions throughout, applied where relevant to characterize individual systems without constituting an independent organizational axis.
A fourth property, \textit{Behavioral Intent}, that is the degree to which a dynamic entity's future motion is deterministic, physics-based or goal-driven, is addressed by only a small subset of the reviewed literature \cite{catalano_aion_2025} \cite{gorlo_long-term_2024} and remains largely decoupled from probabilistic world-model construction.

\section{Graph-Based Robotic World Models For Dynamic Environments}\label{sec:dyn_world_model_representations}
The evolution of world model representation in robotics has undergone a significant paradigm shift, transitioning from rigid, static geometric maps to adaptive, spatio-temporal structure capable of handling and reasoning about dynamic environments \cite{bavle_slam_2023} \cite{rosinol_kimera_2021}.
In this section, we first review factor graphs as geometric probabilistic backbones (Section \ref{sec:dyn_world_model_representations_factor_graphs}), scene graphs as semantic abstractions (Section \ref{sec:dyn_world_model_representations_scene_graphs}) in dynamic environments, and finally discuss how dynamics themselves are represented (Section \ref{sec:dynamics_representations}).

\subsection{Representational Foundations}
\subsubsection{Factor Graphs as Dynamic Geometric World Models}\label{sec:dyn_world_model_representations_factor_graphs}
To mitigate the impact of outliers and non-stationarity in dynamic environments, factor graph-based systems employ a range of probabilistic robustness mechanisms.
A common approach is the use of \textbf{robust cost functions}, such as the Huber loss, which reduce the influence of large residuals during nonlinear least-squares optimization while preserving sensitivity to small, consistent measurement errors\cite{song_rts-slam_2025}. 
This allows the optimizer to tolerate transient violations of the static-world assumption, caused for instance by moving objects or short-time occlusions, without destabilizing the global solution.
Complementary to robust kernels, \textbf{dynamic information reweighting strategies} explicitly modulate the contribution of individual factors based on their consistency with the current state estimate \cite{lingzhu_xiang_robust_2015}. 
Techniques such as Dynamic Covariance Scaling (DCS) or switchable constraints introduce latent scaling variables that adaptively downweight measurements whose residuals are incompatible with the inferred global structure, providing a principled outlier rejection.
Beyond measurement-level robustness, more expressive models introduce \textbf{latent mobility or existence variables} associated with landmarks or objects, which explicitly encode the likelihood that an entity linked to that variable is static, movable, or currently in motion \cite{lingzhu_xiang_robust_2015}.
By scaling the corresponding factor covariances or relaxing constraints when mobility is inferred, these approaches prevent dynamic or semi-static entities from corrupting the static map while still allowing them to be represented within the same probabilistic framework.
Together, these mechanisms transform factor graphs from purely static geometric estimators into uncertainty-aware models capable of reasoning under dynamic, partially observable conditions.
Recently, many works \cite{bescos_dynaslam_2021} \cite{giberna_dynemo-slam_2025} \cite{brink_using_2016} \cite{deeb_model-based_2019} integrated semantic priors transforming the factor graph from a purely geometric map into a context-aware reasoning engine, facilitating the coupling with other graphical models such as scene graphs.  

Traditional SLAM systems are optimized for a static world treating moving objects and agents as outliers to be rejected \cite{bescos_dynaslam_2018} \cite{bavle_s-graphs_2025}. 
Modern dynamic SLAM instead incorporates these dynamic entities into the graph, moving towards a coupled SLAMMOT approach that is examined in detail in Section \ref{sec:dynamics_representations_agents}.

\subsubsection{Scene Graphs as Structured Abstraction in Dynamic World Models}\label{sec:dyn_world_model_representations_scene_graphs}
Dynamics in scene graphs are typically handled through three complementary mechanisms operating at different temporal and semantic levels. 
\textbf{Spatio‑temporal consistency} is addressed by separating short‑term dynamics from long‑term structure: systems such as Khronos maintain a fast \textit{active window} to absorb immediate changes due to moving humans or objects, while a slower global process detects persistent changes and updates the long‑term scene graph accordingly \cite{schmid_khronos_2024}. 
Beyond short‑term consistency, some approaches explicitly model recurring temporal patterns. 
The Aion framework augments hierarchical scene graphs with \textit{Maps of Dynamics} (MoDs), attaching directional, time‑dependent \textbf{dynamic temporal flow} statistics to navigational nodes in order to capture typical or seasonal human motion patterns, while using \textit{Temporal Ownership Transfer} to preserve these dynamics under loop closures and graph reorganization \cite{catalano_aion_2025}. 
Finally, \textbf{belief and persistence reasoning} is used to cope with partial observability and object permanence. 
Belief Scene Graphs introduce probabilistic \textit{blind nodes} to represent entities expected to exist but not currently observed, inferred from learned commonsense co‑occurrence patterns \cite{saucedo_estimating_2025}, while interaction‑driven systems such as DovSG perform localized graph updates during human–object interactions, reflecting state changes without requiring a full scene rescan \cite{yan_dynamic_2025}.

\begin{figure}
    \centering
    \includegraphics[width=0.9\columnwidth]{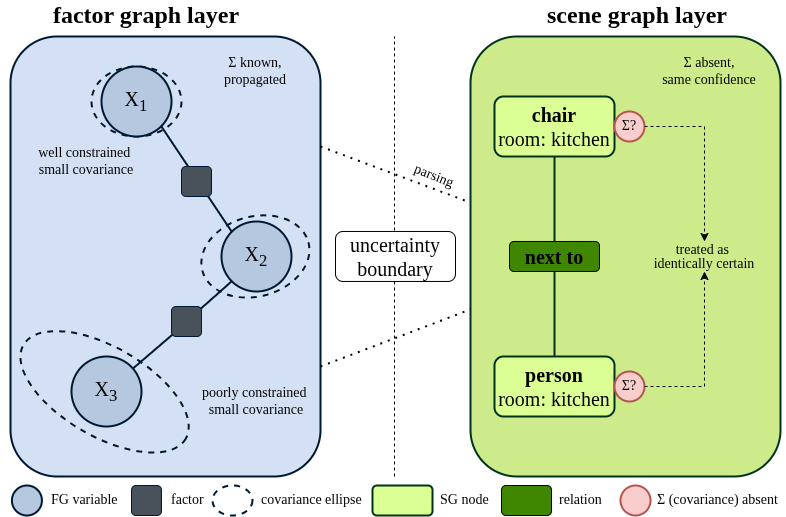}
    \caption{Graphical representation of the uncertainty boundary in the hybrid layered-stack architecture.}
    \label{fig:FG_SG_uncertainty_boundary}
\end{figure}

\subsubsection{Uncertainty Propagation, Failure Modes, Scalability}
Uncertainty management within the factor graph is well-established: variables carry covariance matrices, factors encode information matrices, and the optimizer propagates uncertainty through the Hessian of the nonlinear least-squares problem \cite{dellaert_factor_2017}.
The difficulty arises at the boundary: how does the uncertainty of the pose nodes in a factor graph propagate into the semantic and relational claims made by a scene graph node?
In the layered-stack architecture, this propagation is largely absent, as shown in Figure \ref{fig:FG_SG_uncertainty_boundary}.
Scene graph nodes that are instantiated from a poorly constrained region of the pose graph carries the same confidence as one from a well constrained area. 
In dynamic settings, where uncertainty is generally higher and crucial for spatio-temporal estimation and prediction, this is a large drawback.
Systems that adopt the unified optimizable graph architecture handle this more naturally: by representing semantic entities as factor graph variables, their uncertainty is directly encoded in the Hessian and propagates through joint optimization \cite{bavle_s-graphs_2025}.
Recent work by Millan-Romera et al. \cite{millanromera2026generationuncertaintyawarehighlevelspatial} demonstrates that incorporating learned uncertainty estimates for semantic factors, rather than fixed information matrices, further improves SLAM accuracy, particularly in edge cases where semantic concepts are the only cues preserving geometric consistency.
In dynamic environments the uncertainty is paramount.
A poorly-constrained dynamic entity node produces an unreliable trajectory estimate that, if allowed to influence loop-closure candidates, can corrupt the static map, too\cite{giberna_dynemo-slam_2025}.
Khronos \cite{schmid_khronos_2024} mitigates this by attaching explicit 4D metric uncertainty to its object fragments before they are committed to the global graph, rather than treating every instantiated node as equally confident. 
The broader challenge, i.e., consistent propagation of geometric and temporal uncertainty into semantic relations, (e.g., how certain is the \textit{on} edge between an object and a table, given the covariance on both their pose nodes?) remains largely open and represent one of the key research gaps identified in Section \ref{sec:challenges_future_works}.

A related failure mode in dynamic environments is the corruption of higher-level scene graph abstractions by incorrect dynamic entity handling. 
If a moving person is incorrectly associated with a static wall node during a period of high dynamics, the resulting factor introduces a misleading constraint into the joint optimization that can displace the wall node and corrupt the room topology built on top. 
This cross-layer error propagation is difficult to detect and correct a posteriori, motivating the use of explicit confidence gating to prevent semantically inconsistent observation from entering the graph \cite{zhang_outlier_2023}.

The trade-off between representational richness and real-time inference is a challenge for hybrid factor graph-scene graph systems.
As the environment grows, both the factor graph and the scene graph accumulate nodes and edges, and the computational cost of optimization scales with graph density.
Several strategies have been developed to manage this.
\textit{Hierarchical optimization} exploits the layered structure of scene graphs to scope optimization to the relevant subgraph: S-Graphs 2.0 \cite{bavle_s-graphs_2025} runs room-level local optimization continuously and floor-level global optimization only on loop closure, avoiding full re-optimization of the entire graph at each update.
\textit{Marginalization and sparsification} are applied at the factor graph layer to maintain a bounded-size active window, with older variables marginalized into prior factors. 
DynoSAM \cite{morris_dynosam_2025} and its hybrid world-object centric formulation \cite{morris_online_2025} use embedded-frame anchoring (Section \ref{sec:construction_pipeline_backend}) to preserve sparsity as dynamic object maps grow.
\textit{Task-driven graph pruning}, detailed in Section \ref{sec:querying_reasoning_action_action}, reduces the relational superstructure to a minimal actionable subgraph relevant to the current task, as in Clio \cite{maggio_clio_2024} and SayPlan \cite{rana2023sayplangroundinglargelanguage}, preventing the planning complexity from growing with the full map. 
Together these strategies allow hybrid systems like Khronos \cite{schmid_khronos_2024} and DynEMo-SLAM \cite{giberna_dynemo-slam_2025} to achieve real-time performance in complex large indoor environments. However, the scalability of unified optimizable graphs to large-scale, highly dynamic outdoor scenes where both the agent population and the structural complexity are greater, remains an active research direction and will be discussed in more detail in Section \ref{sec:construction_pipeline_backend}.

\begin{figure}
    \centering
    \includegraphics[width=\columnwidth]{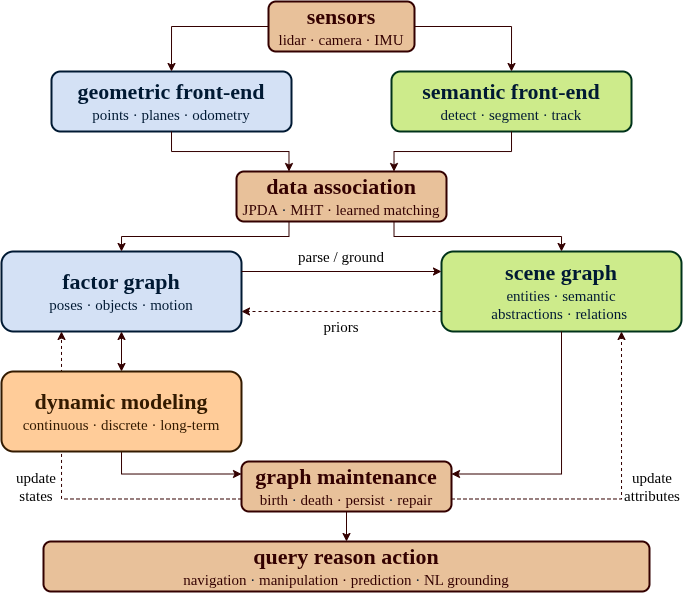}
    \caption{Conceptual architecture of a hybrid probabilistic graph-based dynamic world model.}
    \label{fig:unifying_view_SG_FG}
\end{figure}

Figure \ref{fig:unifying_view_SG_FG} summarizes this hybrid architecture, showing how the factor graph backbone and scene graph superstructure connect through shared dynamic modeling and graph-maintenance stages to support downstream querying, reasoning and action.
This unified representation enables actionable spatial perception, where the world model is not merely a passive map but a persistent, query-able substrate for planning, interaction, and scene prediction.

\subsection{Representing Dynamics in Graph-Based World Models}\label{sec:dynamics_representations}
\begin{figure}
    \centering
    \includegraphics[width=0.9\columnwidth]{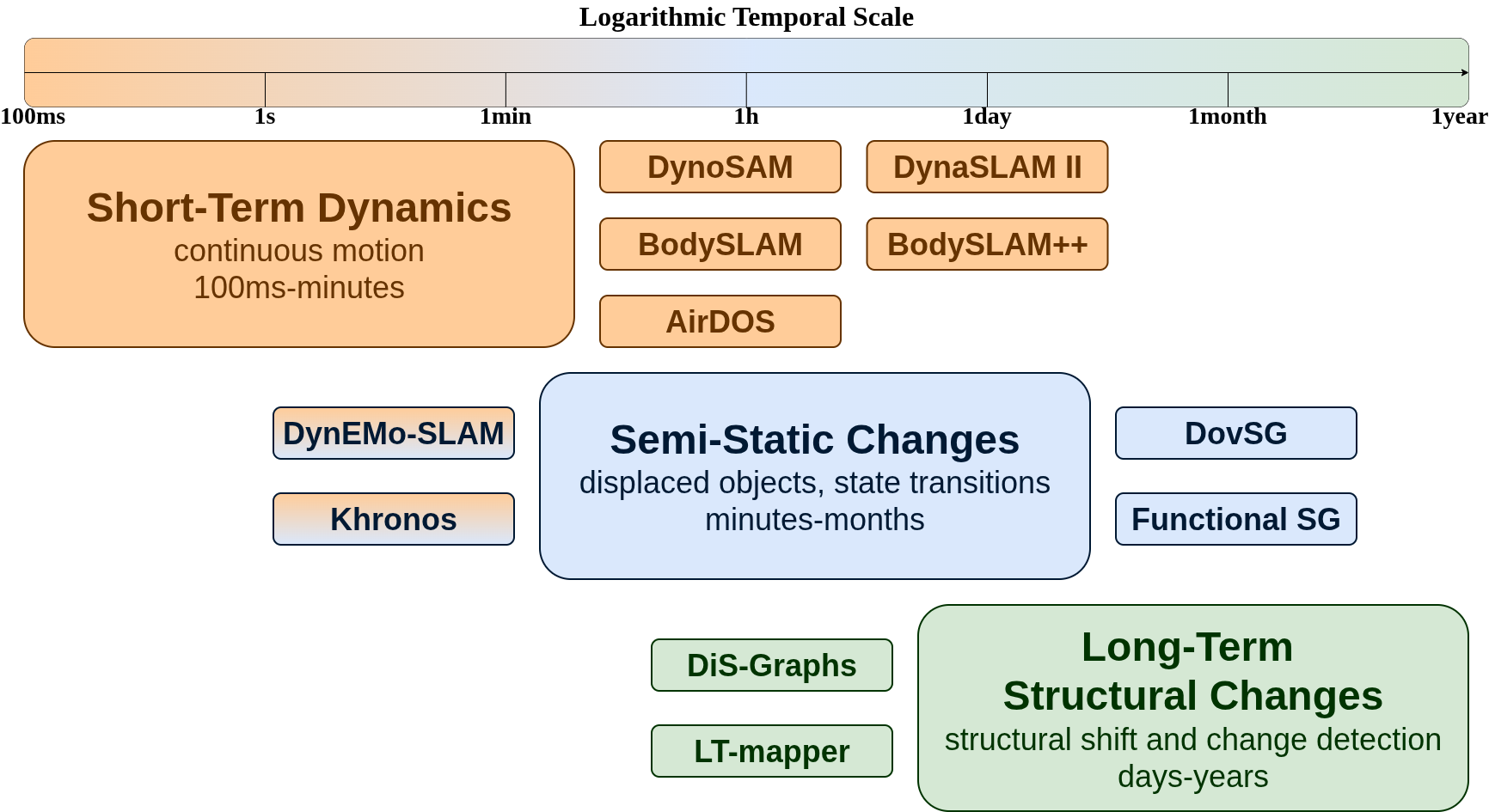}
    \caption{Diagram showing the adopted dynamic distinction ontology, with each group spanning different timescale. Relevant works are reported for each, or group of, type of dynamics.}
    \label{fig:dynamics_timescale}
\end{figure}

The evolution of world model representations for robotics in non-stationary scenarios has progressed from treating environmental change as outlier noise toward a systematic, relational and probabilistic characterization of dynamics across multiple temporal scales \cite{rosinol_3d_2020} \cite{schmid_khronos_2024}.
This section presents the most common characterization of dynamic entities in dynamic world modeling. Specifically, we emphasize the integration of factor graphs and 3DSGs to handle short-term dynamics (Section \ref{sec:dynamics_representations_agents}), semi-static changes (Section \ref{sec:dynamics_representations_objects}), and long-term structural evolution (Section \ref{sec:dynamics_representations_longterm}). 
Figure \ref{fig:dynamics_timescale} depicts the dynamic distinction we adopt, based on the type and temporal scale of expected motion, highlighting some relevant works focusing on each of them.

\subsubsection{Short-Term Dynamics}\label{sec:dynamics_representations_agents}
In environment populated by humans and/or moving objects, the world model must resolve the ``motion ambiguity'' where pixel or point displacement is a compound effect of the simultaneous robot's ego motion and the independent motion of other entities \cite{zhang_semantic-integrated_2026}. 

Early paradigms typically treated short-term dynamic entities as outliers to be masked or rejected \cite{bescos_dynaslam_2018} to preserve the consistency of purely static map. 
Contemporary loosely coupled systems often maintain these as separate threads, using the camera pose as a prior for multi object tracking (MOT) but without allowing object measurements to refine the ego-localization\cite{tian_dl-slot_2024} \cite{tian_dl-slot_2024} \cite{gou_visual_2024}. 
In this way, a uni-directional flow of information is generated, allowing the optimized static map data, output of the SLAM process, to enhance MOT but not the other way around.

Modern representations employ a tightly-coupled SLAMMOT approach, where the factor graph back-end jointly optimizes camera poses, static landmarks, and the trajectories of dynamic objects or agents within a unified optimization manifold \cite{zhang_semantic-integrated_2026}.
Systems like DynaSLAM II \cite{bescos_dynaslam_2021}, DynEMo-SLAM \cite{giberna_dynemo-slam_2025}, SMMF-SLAMMOT \cite{zhang_semantic-integrated_2026} support a tightly-coupled approach where robot poses, static landmarks, and dynamic entity trajectories are jointly optimized in a unified factor graph.
This paradigm requires spatio-temporal reprojection factors to introduce cross-terms between camera poses and object motions, ensuring that the estimation of self and the surrounding is mutually beneficial. 
To address the computational effort of joint optimization, DynoSAM\cite{morris_dynosam_2025} and Hybrid iSAM2 \cite{morris_online_2025} formulations anchor rigid body motions to ``embedded frames'', preserving graph sparsity as each object map grows while enabling efficient incremental inference by a Bayes Tree. 


Humans represent a particularly important and challenging sub-class of short-term dynamics, often referred to as \textit{dynamic agents}: they are simultaneously the most common source of dynamics in real-world environments and the most structurally complex entity to model. This stems from the fact that humans fall into the challenging \textit{articulated} and \textit{goal-driven} categories along the typology and behavioral intent axes, respectively.
Systems like Kimera \cite{rosinol_kimera_2021} and Khronos \cite{schmid_khronos_2024} represent humans with dedicated node types combining 3D pose graph trajectories with non-rigid models, ensuring the world models account for both articulated motion and volumetric presence. 
BodySLAM \cite{henning_bodyslam_2022} and its successor BodySLAM++ \cite{henning_bodyslam_2023} push this further, pioneering the joint optimization of camera trajectory and dense human body shape and pose within a single factor graph, where each human observation is a parametric body model linked to the camera pose graph via reprojection factors. 
BodySLAM++ \cite{henning_bodyslam_2023} extends this to multiple simultaneous human instances, integrating inertial factors in a real-time fashion, improving robustness to occlusion in crowded scenes. 
In both cases, the structured prior provided by the body model regularizes an otherwise underconstrained problem, the same role played by rigid-body kinematic priors for object-class short-term dynamics, like for instance in \cite{qiu2022airdos} accounting for dynamic articulated objects.

The choice of the body model determines the degrees of freedom exposed to the optimizer and the computational costs of inference. 
Parametric models dominate in SLAM-integrated systems: SMPL \cite{loper_smpl_2015} represents the body as a compact, differentiable parameter vector via linear blend skinning, making it naturally compatible with factor graph optimization \cite{henning_bodyslam_2022}\cite{henning_bodyslam_2023}.
SMPL-X \cite{SMPL-X:2019} extends this to hands and face, while STAR \cite{STAR:2020} reduces blend skinning artifacts at extreme poses. 
For monocular initialization without depth sensors, video-based regressors such as VIBE \cite{kocabas2019vibe}, FrankMocap \cite{rong2021frankmocap}\cite{joo2020eft} and PromptHMR \cite{wang2025prompthmr} provide practical SMPL estimates. 
At the opposite end, non-parametric approaches such as METRO \cite{lin2021end-to-end} and neural representations such as InstantAvatar \cite{jiang2022instantavatar} offer greater shape fidelity but sacrifice the compact differentiable structure required for real-time factor graph integration. 
The prevailing trade-off is therefore between the structured compactness of SMPL-family models and the expressiveness of learned representations, summarized in Table~\ref{tab:body_models}.

\begin{table*}[t]
\centering
\caption{Comparison of human body representations relevant to robotic world modeling.
\checkmark~=~supported, \texttimes~=~not supported, $\sim$~=~partial support.
RT: real-time capable.}
\label{tab:body_models}
\resizebox{\textwidth}{!}{
\begin{tabular}{lllllllll}
\toprule
\textbf{Model} & \textbf{Input} & \textbf{RT} & \textbf{Hardware} & \textbf{Training} & \textbf{Output} & \textbf{Body Parts} & \textbf{Realism} & \textbf{Use Case} \\
\midrule

SMPL \cite{loper_smpl_2015}
& Shape/pose params
& \checkmark
& CPU/GPU
& Offline (scan-based)
& Parametric mesh
& Body only
& Low
& SLAM, MOT, animation \\

SMPL-X \cite{SMPL-X:2019}
& Shape/pose/expr.\ params
& $\sim$
& GPU
& Offline (scan-based)
& Parametric mesh
& Body + hands + face
& Low
& Expressive HRI, animation \\

STAR \cite{STAR:2020}
& Shape/pose params
& \checkmark
& CPU/GPU
& Offline (scan-based)
& Parametric mesh
& Body only
& Low
& SLAM, extreme pose recovery \\

Anny \cite{bregier2025humanmeshmodelinganny}
& Phenotype params
& \checkmark
& CPU/GPU
& None (scan-free)
& Parametric mesh
& Body + hands + face
& Low
& Diverse pop.\ modeling, HMR \\

VIBE \cite{kocabas2019vibe}
& Monocular video
& $\sim$
& GPU
& Offline (video data)
& SMPL params/frame
& Body only
& Low
& Video-based pose init. \\

FrankMocap \cite{rong2021frankmocap} \cite{joo2020eft}
& RGB image
& \checkmark
& GPU
& Offline (images)
& SMPL-X params
& Body + hands
& Low
& Real-time monocular HMR \\

METRO \cite{lin2021end-to-end}
& RGB image
& \texttimes
& GPU (high-end)
& Offline (images)
& Raw mesh vertices
& Body + hands
& Medium
& Accurate shape recovery \\

InstantAvatar \cite{jiang2022instantavatar}
& Monocular video
& \texttimes
& GPU (high-end)
& Per-subject online
& Neural field + mesh
& Full body
& High
& Avatar reconstruction \\

\bottomrule
\end{tabular}
}
\end{table*}


\subsubsection{Semi-Static Changes}\label{sec:dynamics_representations_objects}
Semi-static dynamics involve objects that are stationary during observation but relocated over time, ending up displaced in future observations \cite{brink_using_2016} \cite{giberna_dynemo-slam_2025}.
This requires the model to distinguish between ``moving'' and ``movable'' states. 
Additionally, it is needed to learn what kinds of changes a specific entity can undergo and how to handle object disappearance. 

Nodes in modern 3DSGs are annotated with dynamic attributes such as \textit{open/closed} or \textit{on/off} \cite{olivastri_multi-modal_2024}. 
These properties, often inferred via Vision-Language Models \cite{shao_large_2025}, define \textit{functional scene graphs} \cite{rotondi_fungraph_2025} \cite{zhang_open-vocabulary_2025} where interactive elements like handles are linked to the objects they trigger, ending up to change their states. 

The key question of permanence is handled via hierarchy and containment edges (for instance an object in a drawer) \cite{behrens_lost_2025} \cite{yan_dynamic_2025}.
System like Lost \& Found \cite{behrens_lost_2025} maintain object permanence by updating these relations online. 
If a human hides an item in a drawer, the \textit{contains} edge ensures the robot \textit{remembers} the location despite total occlusion. 

Decisions to prune the graph are often governed by probabilistic decay rates \cite{gomez_object-based_2020} or explicit geometric verification \cite{schmid_khronos_2024}.
In DPG-SLAM \cite{walcott-bryant_dynamic_2012}, sectors of a node are turned off if they intersect with evidence of removal. 
A node is marked inactive if a threshold of sectors is off, but it is only deleted if it no longer provides useful topological information, thereby managing complexity without losing historical data. 

\subsubsection{Long-Term Structural Changes} \label{sec:dynamics_representations_longterm}
Lifelong autonomy necessitates a world model that can align multiple sessions and detect structural shifts in the environment \cite{deeb_model-based_2019} \cite{catalano_3d_2025}, precluding the assumption of staticity in entities or parts of the map which would otherwise be considered unmovable in shorter or single-session campaigns.

At geometric level, structural change detection compares observations from different sessions to classify map regions as stable, changed or newly appeared. 
DPG-SLAM \cite{walcott-bryant_dynamic_2012} addresses this by continuously comparing current sensor data against submaps from previous sessions.
Regions where contradictions occur are marked as changed, and associated nodes are pruned.
LT-mapper \cite{kim2022lt} extends this to large-scale outdoor LiDAR SLAM through anchor-node-based inter-session loop factors that align drifted session pose graphs into a shared coordinate frame, applying then set-difference operations to identify positive and negative changes at keyframe level.

Merging observations from multiple sessions into a globally consistent factor graph requires solving both the inter-session pose offset and the data association between observation of potentially changed structures.
The standard mechanism is the introduction of inter-session loop closure factors, i.e.,  constraints between nodes from different sessions that share stable landmarks \cite{kim2022lt}.
Memory management is critical to keep this tractable. As shown in \cite{labbe2014online}, a hierarchical short-term/working/long-term memory structure can bound the number of loop closure candidates, enabling multi-session operation under computational constraints. 

A complementary approach to multi-session alignment uses prior structural knowledge, typically in form of Building Information Models (BIM) or architectural plans, to anchor the factor graph to a known coordinate frame and constrain the estimation of structural elements. 
BIM-SLAM \cite{vega_torres_bim-slam_2023} models both the robot's SLAM graph and the architectural plan as hierarchical factor graph, then detects inter-session loop closure between them. 
DiS-Graphs \cite{shaheer_tightly_2025} extends this to jointly estimate deviations between the planned and the built environments, incorporating BIM-derived structural priors as soft constraints in the factor graph to remain accurate even when the real building deviates from its original plans.
The key insight in both works lies in the hierarchical scene graph structure, e.g., walls, rooms, floors, which provides a natural alignment target that is more robust to partial scene change than point-level descriptors.
The graph update policy in these systems therefore distinguishes between permanent structural entities, even in case of deviations or changes, which are retained and used for inter-session alignment, and semi-static or movable entities.

\section{Dynamic World Model Construction}\label{sec:construction_pipeline}
Dynamic world model construction refers to the end-to-end process by which raw sensor observations are transformed into a persistent, probabilistically grounded, and semantically structured representation of a non-stationary environment. 
In hybrid representations combining factor graphs and scene graphs, this process spans multiple tightly interacting stages, from front-end perception and data association to back-end inference and long-term graph maintenance.

\begin{figure}
    \centering
    \includegraphics[width=\columnwidth]{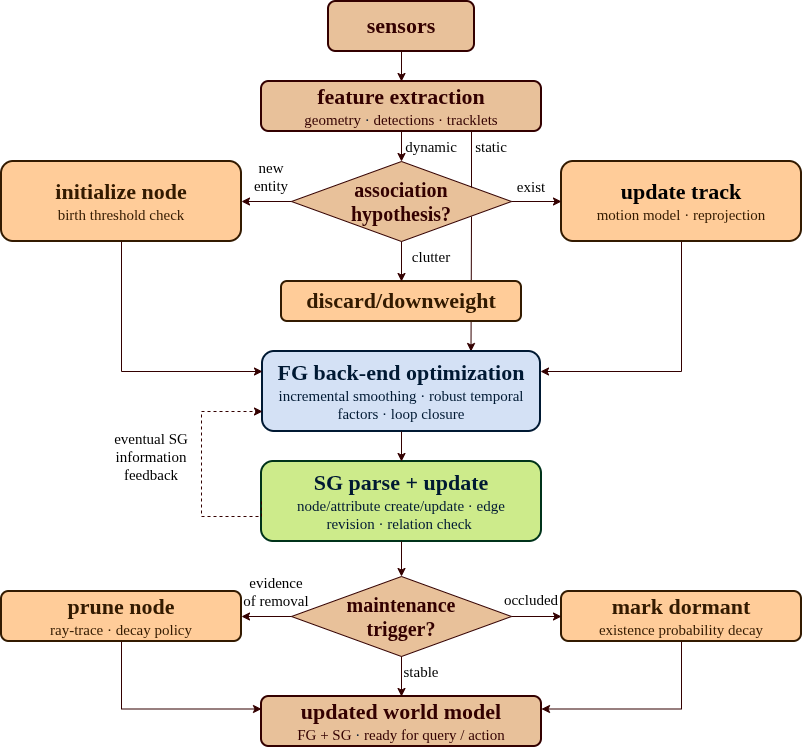}
    \caption{Per-frame operational pipeline of the hybrid world model.}
    \label{fig:FG_SG_world_model_construction_maintenance_workflow}
\end{figure}

Rather than a monolithic pipeline, modern systems adopt modular yet interdependent components, where design choices at early stages (e.g., detection granularity or association strategy) strongly influence the stability, scalability, and expressiveness of the resulting world model under dynamics.
This section organizes existing approaches around four core stages that recur across the literature: front-end perception (Section \ref{sec:construction_pipeline_frontend}), data association (Section \ref{sec:construction_pipeline_data_association}), back-end inference and optimization (Section \ref{sec:construction_pipeline_backend}), and graph maintenance (Section \ref{sec:construction_pipeline_graph_maintenance}).
A general pipeline overview is depicted in Figure \ref{fig:FG_SG_world_model_construction_maintenance_workflow}.

\subsection{Front-End Perception}\label{sec:construction_pipeline_frontend}
The front-end perception stage is responsible for extracting candidate entities, attributes and relations from raw sensor data.
In dynamic environments, this stage must operate under partial observability, occlusion and motion ambiguity, while providing sufficiently rich primitives for downstream probabilistic inference. 

\subsubsection{Geometric and Semantic Primitives}
At the geometric level, front-ends extract low-level features such as points, lines, planes or surfels, which remain the backbone of factor graph-based SLAM systems \cite{dellaert_factor_2017} \cite{rosinol_kimera_2021}. 
However, dynamic scenes increasingly require object- and agent-level primitives, motivating the integration of instance segmentation, 3D bounding volumes, or parametric object models \cite{bescos_dynaslam_2021} \cite{morris_dynosam_2025} \cite{schmid_khronos_2024}.

Semantic perception pipelines typically produce:
\begin{itemize}
    \item \textbf{Object detections and instances}, often class-labeled and possibly open-vocabulary \cite{yan_dynamic_2025} \cite{gu_conceptgraphs_2023};
    \item \textbf{Agent hypotheses}, including humans or vehicles, sometimes with articulated or rigid-body structure \cite{henning_bodyslam_2022} \cite{qiu2022airdos};
    \item \textbf{Attribute estimates}, such as object state (open/close, on/off) \cite{olivastri_multi-modal_2024};
    \item \textbf{Relation hypotheses}, e.g., spatial (on, in, next-to) or functional (controls, supports) \cite{catalano_3d_2025} \cite{zhang_open-vocabulary_2025}.
\end{itemize}

In hybrid FG-SG systems like Kimera \cite{rosinol_kimera_2021} or Khronos \cite{schmid_khronos_2024}, these outputs are deliberately redundant: dense geometric observations coexist with higher-level semantic abstraction, resulting in robustness to perception failures. 

\subsubsection{Temporal Grouping and Tracking}
To cope with dynamics, front-ends often incorporate \textit{short-term temporal aggregation}, producing tracklets or object fragments prior to global optimization. 
Optical flow, frame-to-frame association, or local tracking modules are commonly used to stabilize detections before they enter the global world model \cite{morris_dynosam_2025} \cite{schmid_khronos_2024}.
Khronos \cite{schmid_khronos_2024}, for example, explicitly constructs temporally consistent object fragments ahead of the active-window integration described in Section \ref{sec:dyn_world_model_representations_scene_graphs}, postponing commitment to the global model until sufficient evidence is accumulated.

This early temporal coherence reduces spurious node creation and limits the combinatorial computational effort on later data association stages.

\subsection{Data Association}\label{sec:construction_pipeline_data_association}
Data association is the process of determining which observations correspond to which existing entities in the world model, or whether they should instantiate new ones. 
In dynamic environments, association errors are among the primary failure reasons, as incorrect correspondences can propagate across time, capturing both geometric estimation and semantic structure. 

\subsubsection{Deterministic and Probabilistic Association}
Classical SLAM systems rely on deterministic association mechanisms, such as nearest-neighbor matching or Hungarian assignment based on geometric or appearance similarity \cite{lingzhu_xiang_robust_2015} \cite{gou_visual_2024}. 
While efficient, these approaches degrade quickly under occlusion and fast motion.

To address this, many dynamic-world-model systems employ \textit{probabilistic multi-hypothesis association}, including:
\begin{itemize}
    \item \textbf{Joint Probabilistic Data Association (JPDA)}, which marginalizes over multiple correspondence hypotheses \cite{gulati_robust_2017};
    \item \textbf{Multiple Hypothesis Tracking (MHT)}, which explicitly maintains competing association trees over time \cite{zaech_learnable_2022};
    \item \textbf{Learned association}, where neural embeddings (e.g. CLIP features) define similarity metrics for object re-identification \cite{gu_conceptgraphs_2023} \cite{yan_dynamic_2025}.
\end{itemize}
In factor-graph formulations, association decisions are often represented implicitly via latent correspondence variables or explicitly through switchable or weighted constraints \cite{giberna_dynemo-slam_2025} \cite{lingzhu_xiang_robust_2015}.

\subsubsection{Association under Dynamics}
Dynamic environments needs distinguishing between moving agents, movable but stationary objects, and ephemeral observations, like transient clutter or false positives.
Several works incorporate \textit{movability or existence variables} to mediate association decisions \cite{brink_using_2016} \cite{gomez_object-based_2020}.  
From a probabilistic perspective, Random Finite Set (RFS)-based formulations (Section \ref{sec:construction_pipeline_graph_maintenance}) provide a mechanism for handling association, birth and death jointly \cite{ristic2013tutorial} \cite{garcia2018poisson}, although full integration with scene graph construction remains limited.

\subsection{Back-End Inference/Optimization}\label{sec:construction_pipeline_backend}
The back-end integrates associated observations into a globally consistent world model, resolving the geometric, semantic and temporal constraints through probabilistic inference.

\subsubsection{Incremental Factor Graph Optimization}
Most modern systems rely on \textit{incremental smoothing} over factor graphs, using solvers such as iSAM2 \cite{kaess2011isam2} or g2o \cite{kummerle_g2o_2011} to maintain real-time performance under continuous updates \cite{dellaert_factor_2017} \cite{morris_dynosam_2025} \cite{bavle_situational_2022}.
Variables typically include robot poses, static landmarks, dynamic objects states, and sometimes higher-level structural entities (walls, rooms) \cite{bavle_s-graphs_2025}.

Dynamic environments additionally introduce new challenges:
\begin{itemize}
    \item \textbf{Spatio-temporal factors} couple robot motion and object motion \cite{morris_dynosam_2025} \cite{zhang_semantic-integrated_2026};
    \item \textbf{Switchable or robust factors} downweight inconsistent measurements caused by dynamic entities \cite{song_rts-slam_2025} \cite{lingzhu_xiang_robust_2015};
    \item \textbf{Motion priors} (e.g. constant velocity) regularize underconstrained trajectories \cite{morris_dynosam_2025} \cite{giberna_dynemo-slam_2025} \cite{bescos_dynaslam_2021}.
\end{itemize}

\subsubsection{Continuous Motion Modeling}
Continuous dynamics are primarily modeled as trajectories of robot poses ($\mathit{X}_k$) or dynamic object poses ($\mathit{L}_k$) and motions ($\mathit{H}_k$).
Morris et al. \cite{morris_importance_2024} highlights a fundamental dichotomy in coordinate frame representation, presenting World-Centric and Object-Centric Formulations.
Traditional object-centric models express dynamic points in a local body frame, where they appear static.
However, this necessitates estimating the pose of each object at every timestep, often leading to instability because these frames are not directly observable. 
On the other side, world-centric representation describe rigid-body motion directly using variables in the world frame. 
Recent ``hybrid'' formulations \cite{morris_online_2025} anchor motion to an embedded frame ($\mathit{L}_e$) initialized upon the first observation, which preserves factor graph sparsity and enhances convergence for incremental solvers like iSAM2. 

When it comes to the design of kinematic motion models, most frameworks adopt a constant velocity model\cite{bescos_dynaslam_2021} \cite{morris_dynosam_2025}, to regularize the optimization problem. 
Some higher-fidelity models include constant acceleration \cite{brink_using_2016} or specific zero velocity priors to handle humans or robots that may remain stationary for intervals of time \cite{rosinol_3d_2020}.

These motion priors are realized in the back-end through spatio-temporal factors that couple camera poses with object motion vertices, introducing cross-terms in the Hessian matrix that improve accuracy and efficiency compared to independent spatial factors \cite{zhang_semantic-integrated_2026}.

\subsubsection{Loop Closure and Multi-layer Consistency}
Loop closure remains essential for long-term consistency but is particularly risky under dynamics. 
Incorrect loop closures induced by moved objects or agents can catastrophically corrupt the map \cite{giberna_dynemo-slam_2025}.
Systems like Khronos \cite{schmid_khronos_2024} and DPG-SLAM \cite{walcott-bryant_dynamic_2012} delay or gate loop closure decisions using geometric verification and temporal consistency checks.

In hybrid FG-SG systems, back-end inference must also ensure \textit{cross-layer consistency}. 
First, geometric optimization updates metric nodes poses. 
Furthermore, scene graph abstractions (e.g. room belonging and containment relations) must be updated accordingly. 
Unified optimizable representations, like S-Graphs \cite{bavle_s-graphs_2025}, handle this naturally, while layered architectures require explicit synchronization policies \cite{hughes_hydra_2022} \cite{rosinol_kimera_2021}.

\subsection{Graph Maintenance}\label{sec:construction_pipeline_graph_maintenance}
Beyond inference, dynamic world models must remain computationally tractable, stable and meaningful over long time horizons. 
Graph maintenance governs how nodes and edges are added, updated, kept, or removed.

\subsubsection{Node Lifecycle: Birth, Persistence, Death}
Entity lifecycle management is central in dynamic environments. 
Common strategies include:
\begin{itemize}
    \item \textbf{Evidence accumulation thresholds} for node birth \cite{schmid_khronos_2024};
    \item \textbf{Temporal decay or confidence scores} governing persistence \cite{gomez_object-based_2020};
    \item \textbf{Geometric verification} (e.g. ray tracing) to distinguish occlusion from true removal \cite{schmid_khronos_2024}.
\end{itemize}

These heuristics can be grounded in a more principled probabilistic treatment.
The Random Finite Set (RFS) framework elevates birth and death from heuristic thresholds to first-class elements of the Bayesian filtering recursion \cite{ristic2013tutorial}. 
In this formulation, the set of tracked entities is itself a random variable, and existence is modeled explicitly through a binary latent variable $\epsilon_k \in \{0,1\}$ evolving as a first-order Markov chain with transition probabilities $p_B$ (birth) and $p_S$ (survival). 
The Bernoulli filter propagates an existence probability $r_k \in [0,1]$ alongside the spatial density, decomposing additively into a survival term $p_S \cdot r_{k-1}$ and a birth term $(1-r_{k-1})\cdot p_B$ \cite{ristic2013tutorial}. 
This formulation has a direct and underexplored connection to dynamic world models: each scene graph node representing a potentially transient entity, e.g. a movable chair, a walking human, a momentarily occluded object, can in principle be equipped with such an existence probability, making node lifecycle a Bayesian inference problem rather than a threshold-based engineering decision.

The Probability Hypothesis Density (PHD) filter extends this to the multi-entity case by propagating the first-order moment of the multi-target posterior, avoiding explicit data association while jointly estimating the number and locations of dynamic entities \cite{mullane2011random}. 
Its application to SLAM \cite{mullane2011random} showed that modeling the landmark map as an RFS rather than a random vector provides a principled treatment of feature appearance, disappearance, missed detections and clutter within a single Bayesian recursion, and has since been applied directly to SLAMMOT \cite{siew2018simultaneous}. 
This separation into a static factor graph layer and a dynamic PHD layer mirrors the architectural splits between the factor graph backbone and the dynamic node layer of hybrid FG-SG systems.

For the full multi-entity case with explicit track management, the Poisson Multi-Bernoulli Mixture (PMBM) filter provides a Bayes-optimal conjugate prior \cite{garcia2018poisson}. 
A Poisson Point Process describes the existence of yet undetected targets, while a Multi-Bernoulli Mixture describes the distribution of already-detected targets \cite{granstrom2019poisson}. 
This maps naturally onto robotic world modeling: undetected entities correspond to the PPP intensity, while confirmed tracks carry an explicit existence probability $r_k^i$ updated at each timestep \cite{deusch2015labeled}, with track pruning by thresholding \cite{reuter2014labeled}. 
The PMBM formulation has been extended directly to SLAM \cite{kim2022pmbm}, and more recently combined with a graph-based SLAM back-end \cite{ge2025batch} to bridge RFS-based entity management with factor graph optimization.

The appearance rate of new entities can itself be modeled with a Poisson process of intensity $\lambda_B(\cdot)$ over the state space \cite{garcia2018poisson}; non-homogeneous variants let this birth intensity adapt over time, capturing phenomena such as increased agent density during peak hours \cite{li2023adaptive}. 
This connects naturally to Aion's MoD (Section \ref{sec:dyn_world_model_representations_scene_graphs}), readable as a time-varying birth intensity learned empirically over navigational nodes rather than specified a priori. 
Semantic extensions such as the Semantic PHD filter \cite{chen2019multi} further augment existence estimation with class labels, bridging the RFS tracking layer and the semantic node types of scene graphs.

Ray-tracing offers a geometric instantiation of this Bayesian update: if a ray from the robot's current pose to a background vertex passes through a previously mapped object's volume and returns a longer-than-expected depth, the system confirms removal rather than occlusion \cite{schmid_khronos_2024}. 
Belief Scene Graphs' blind nodes (Section \ref{sec:dyn_world_model_representations_scene_graphs}) correspond conceptually to the PPP component of the PMBM framework, a distribution over undetected but plausible entities, though implemented via Graph Convolutional Networks rather than closed-form Bayesian recursion \cite{saucedo_estimating_2025}. 
In systems like Aion \cite{catalano_aion_2025}, motion dynamics are initially stored in a spatial hash cell and only bound to a navigational node once the local topology has stabilized (after a loop closure); if a node is removed, its temporal history is retained in the hash space for potential re-association, providing a mechanism for soft deletion analogous to track dormancy in PMBM-based trackers.

\subsubsection{Discrete State Transition Modeling}
Beyond continuous motion and existence, semi-static entities exhibit discrete state changes that must be captured mechanistically rather than only taxonomically. 
A common approach assigns latent indicator variables to each feature: $\mathit{x}_{i,t}$, whether it is currently moving, and $\mathit{m}_{i,t}$, whether it is capable of movement \cite{brink_using_2016}. 
This allows the system to distinguish between an inherently short-term dynamic entity and a movable object (semi-static) that happens to be static.

Some models avoid binary classification in favor of a continuous ``object probability'' or ``movability'' score \cite{gomez_object-based_2020}, capturing the likelihood of finding an object in its last known location based on long-term experience, allowing robots to learn the ``rhythm'' of a dynamic household.

\subsubsection{Structural Stability and Forgetting}
Lifelong systems must prevent unbounded graph growth.
Approaches such as DPG-SLAM \cite{walcott-bryant_dynamic_2012} and Lost \& Found \cite{behrens_lost_2025} introduce \textbf{selective forgetting}, pruning nodes only when they no longer contribute topological or semantic information.
Hierarchical scene graphs further localize updates, allowing changes to be absorbed at appropriate abstraction levels (object vs. room vs. floor).

\subsubsection{Handling Errors and Recovery}
Incorrect associations or transient perception failures are inevitable. 
Robust systems therefore incorporate:
\begin{itemize}
    \item \textbf{Local graph repair}, updating only affected subgraphs \cite{yan_dynamic_2025};
    \item \textbf{Fragment-based representations}, isolating inconsistent observations \cite{schmid_khronos_2024};
    \item \textbf{Delayed commitment}, postponing semantic or structural decisions until sufficient confidence is reached \cite{schmid_khronos_2024}, mainly to overcome data association issues.
\end{itemize}

Graph maintenance thus acts as the final safeguard against error accumulation, ensuring that the world model remains usable for downstream tasks even under sustained dynamics.

\section{Interaction: Querying, Reasoning, and Action}\label{sec:querying_reasoning_action}
The shift from static environmental representations to \textit{actionable spatio-temporal perception} marks a fundamental transition where the world model is no longer a passive record of geometry but an active substrate for robot reasoning and decision-making \cite{rosinol_3d_2020} \cite{bavle_slam_2023}.
The convergence of factor graphs, providing metric-semantic consistency, and (3D) scene graphs, offering hierarchical relational logic, enables a rich set of interaction modalities, encompassing multi-modal querying, hierarchical planning, and closed-loop maintenance.

\subsection{Multi-Modal Query Paradigms}\label{sec:querying_reasoning_action_query}
The literature identifies three primary query modalities that leverage the structured nature of hybrid world models:

\noindent\textbf{Geometric-Metric Queries:} leveraging the low-level metric semantic mesh or factor-graph-backed pose and places layers, agents execute queries for fast collision checking via Bounding Volume Hierarchies\cite{rosinol_kimera_2021} and metric path planning\cite{catalano_3d_2025}, e.g. $A^*$ or Dijkstra.

\noindent\textbf{Symbolic-Semantic Queries:} hierarchical 3D Scene Graphs support high level symbolic commands, such as ``get the medication bottle from the nightstand''\cite{catalano_3d_2025}, often assisted by LLMs to achieve long-term trajectory prediction\cite{gorlo_long-term_2024}. 
By grounding abstract terms in specific room or object nodes, the model resolves spatial relationships, such as \textit{on}, \textit{inside}, to isolate target entities.
Open-vocabulary frameworks further extend this to previously unseen categories using CLIP-based embeddings. 

\noindent\textbf{Probabilistic-Expectation Queries:} 
systems like Belief Scene Graphs \cite{saucedo_estimating_2025} repurpose blind nodes (Section \ref{sec:dyn_world_model_representations_scene_graphs}) to answer ``what is likely present, even if unobserved'', using the same GCN-based co-occurence priors to predict object or, potentially, agent locations.

While these three paradigms differ in representation and reasoning mechanism, Table~\ref{tab:query_paradigms} reveals a shared limitation: each is fundamentally a query over a \textit{static} world snapshot. 
Geometric-metric queries assume the underlying mesh remains valid until the next re-planning trigger; symbolic-semantic queries assume the scene graph reflects the current arrangement of entities, requiring re-detection or change-point tracking to remain accurate; and probabilistic-expectation queries model uncertainty over \textit{unobserved} space rather than uncertainty over \textit{time}, leaving moving or transient objects largely unaddressed. 
Bridging this gap, recent work on dynamic scene graphs and continual re-localization (e.g.,~\cite{ejaz2025situationallyawarepathplanningexploiting}) begins to treat the world model itself as a temporally evolving belief, suggesting that future query paradigms may need to fuse the spatial reasoning surveyed above with explicit temporal consistency mechanisms.

\begin{table*}[t]
\centering
\scriptsize
\setlength{\tabcolsep}{3pt}
\renewcommand{\arraystretch}{0.0}
\caption{Comparison of the three query paradigms over hybrid world models.}
\label{tab:query_paradigms}
\begin{tabularx}{\textwidth}{@{}l X X X@{}}
\toprule
\textbf{Dimension} & \textbf{Geometric-Metric} & \textbf{Symbolic-Semantic} & \textbf{Probabilistic-Expectation} \\
\midrule
Representation &
Metric-semantic mesh, factor-graph pose layers, occupancy/TSDF volumes &
Hierarchical 3D scene graph (room/object/relation nodes) &
Scene graph augmented with latent \textit{blind nodes} and learned priors \\
\addlinespace
Core question &
Is this space free? What is the shortest path from A to B? &
Where is object X? What satisfies relation $R$ to $Y$? &
What is \textit{likely} present, even if unobserved? \\
\addlinespace
Typical methods &
BVH collision checks~\cite{rosinol_kimera_2021}, $A^*$, Dijkstra, RRT*, ESDF planners &
Graph traversal/matching, LLM-grounded language-to-node resolution~\cite{gorlo_long-term_2024}, CLIP-based open-vocabulary retrieval &
GCNs, learned co-occurrence priors, Bayesian belief updates~\cite{saucedo_estimating_2025} \\
\addlinespace
Granularity &
Fine-grained, continuous/metric &
Discrete, categorical, relational &
Discrete, but probability-weighted over unobserved nodes \\
\addlinespace
Failure modes &
Degrades under sensor noise/incomplete reconstruction; costly at scale without hierarchical pruning &
Sensitive to grounding errors; brittle to novel categories without open-vocab support &
Priors can encode dataset bias; blind-node confidence is hard to calibrate \\
\addlinespace
Role of dynamics &
Assumes quasi-static mesh; requires re-planning triggers on change &
Assumes static graph; needs incremental updates from re-detection &
Models unobserved-space uncertainty, not temporal/moving-object uncertainty \\
\bottomrule
\end{tabularx}
\end{table*}

\subsection{Action Coupling and Hierarchical Planning}\label{sec:querying_reasoning_action_action}
Interaction with the environment is increasingly handled through abstractions that trade off model fidelity for computational efficiency \cite{ejaz2025situationallyawarepathplanningexploiting}.

\noindent\textbf{Hierarchical and Multi-Resolution Planning}: 3D Scene Graphs enable robots to decompose complex tasks across abstraction layers.
For instance, path planning can occur first at building or room level before refining to traversable place nodes, accelerating query times by several orders of magnitude compared to dense volumetric approaches \cite{viswanathan2024xflieleveragingactionablehierarchical} \cite{viswanathan2025spadescalablepathplanning} \cite{ejaz2025situationallyawarepathplanningexploiting} \cite{ray2024taskmotionplanninghierarchical}, and allowing for easier storage of pre-computed path when replanning is necessary \cite{ejaz2025situationallyawarepathplanningexploiting}.

\noindent\textbf{Task-Conditioned Subgraphs:} To manage the scale of massive environmental graphs, frameworks like \textit{SayPlan}\cite{rana2023sayplangroundinglargelanguage} and \textit{Clio}\cite{maggio_clio_2024} use ``task-driven'' abstraction. 
These systems interactively shrink the 3D Scene Graph by clustering and pruning nodes irrelevant to the current natural language mission, creating a minimal, actionable subgraph ready to use for the planner.

\noindent\textbf{Control-Centric Hierarchy:} Some architectures incorporates an \textit{Action Layer} directly into the scene graph \cite{ravichandran_hierarchical_2021}. 
This layer represents candidate robot actions, such as \textit{forward}, \textit{turn}, as graph nodes connected to semantic elements, grounding affordances, and next-step reasoning within the spatial model.

\subsection{Dynamic Updates and Closed-Loop Consistency}\label{sec:querying_reasoning_action_updates}
In non-stationary environments, maintaining the accuracy of the world model requires the continuous correction of nodes and edges as the scene evolves. 

\noindent\textbf{Re-Perception and Credibility Loops:} Systems like SG-Nav \cite{yin2024sgnav} implement a re-perception mechanism that evaluates the credibility of goal candidates. 
If a classification is found to be ambiguous or incorrect during closer inspection, the agent dynamically adjusts the graph structure in real time.

\noindent\textbf{Interaction-Aware Updates:} DovSG \cite{yan_dynamic_2025} and Lost\&Found\cite{behrens_lost_2025} detect hand-object contact events to update object poses and parent-child relations locally, without requiring a full-dense scene reconstruction. 

\noindent\textbf{Geometric Verification:} 
Khronos \cite{schmid_khronos_2024} applies the removal-vs-occlusion check at the interaction layer, back-propagating confirmed removals to prune stale nodes and maintain temporal consistency.

\subsection{Functional Interaction and Action-Effect Reasoning}\label{sec:querying_reasoning_action_interaction}
Beyond spatial relations, recent world models explicitly encode \textit{functional} structure that captures how robot actions induce state changes in the environment.
Functional Scene Graphs model interactive elements such as handles, switches, and control surfaces as first-class nodes, and represent causal dependencies through directed functional relations (e.g. \textit{controls}, \textit{opens}, \textit{enables}) \cite{zhang_open-vocabulary_2025} \cite{rotondi_fungraph_2025}. 
This representation allows robots to reason about which actions are required to achieve a desired state change, effectively grounding manipulation and task planning directly in the world model.

Crucially, functional relations are activated through interaction: they are updated when actions are executed or when human–object interactions are observed. 
This enables localized, semantically meaningful graph updates without requiring global re-optimization or dense scene reconstruction, as demonstrated in interaction-aware systems such as DovSG \cite{yan_dynamic_2025} and Lost \& Found \cite{behrens_lost_2025}. 
By encoding action–effect couplings explicitly, functional scene graphs bridge symbolic task reasoning and geometric execution, providing a structured model for multi-step manipulation and interaction in dynamic environments.




\section{Applications}\label{sec:applications}
The dynamic world models presented throughout this paper are not limited in their generation, but their value is realized through the downstream tasks they can enable.
The shift from static, geometric maps to hybrid probabilistic graph representations that jointly handle geometry, semantics and temporal changes unlocks a richer set of robotic capabilities.
This section organizes these applications around four themes that progress from self-localization to high-level reasoning: robust SLAM and consistent world modeling under dynamics (Section \ref{sec:applications_slam_dynamics}), multi-object tracking and interaction understanding (Section \ref{sec:applications_mot}), navigation and hierarchical planning (Section \ref{sec:applications_navigation}), scene understanding, memory and long-term retrieval (Section \ref{sec:applications_scene_understanding}).

\subsection{Robust SLAM and Consistent World Modeling under Dynamics}\label{sec:applications_slam_dynamics}
The most immediate application of dynamic world models is improving the robustness of self-localization and map building and updating when the environment contains moving or movable entities. 
In static-world SLAM, dynamic elements degrade trajectory estimation through incorrect data associations and inconsistent landmark observations, leading to the standard approach of outlier rejection but ending up discarding information that a richer model could exploit.
Tightly-coupled systems such as DynaSLAM II \cite{bescos_dynaslam_2021}, DynoSAM \cite{morris_dynosam_2025}, BodySLAM++\cite{henning_bodyslam_2023} and AirDOS \cite{qiu2022airdos} demonstrate that the joint-optimization formulation introduced in Section \ref{sec:dynamics_representations_agents} improves ego-trajectory accuracy compared to outlier-rejection baselines, since the cross-terms make self-localization and dynamic-entity tracking mutually reinforcing.
DynEMo-SLAM \cite{giberna_dynemo-slam_2025} applies this formulation in practice, allowing the SLAM back-end to benefit from semantic priors on entity motion while producing a richer joint world model.

The world model also needs to remain consistent across sessions and under long-term changes. 
Khronos \cite{schmid_khronos_2024} applies the spatio-temporal mechanisms described in Section \ref{sec:dyn_world_model_representations_scene_graphs} and \ref{sec:construction_pipeline_graph_maintenance} at system scale, demonstrating that the same active-window/global-process split supports lifelong operation across extended deployments.
Systems like DPG-SLAM \cite{walcott-bryant_dynamic_2012} implement explicit forgetting policies, pruning nodes only when they no longer contribute topological information, thereby managing graph complexity without losing historical data.
These mechanisms are essential for robots operating across extended deployment horizons, such as warehouses, offices, and hospitals, where the entity population changes between sessions, and represent a direct application of the probabilistic birth-death modeling discussed in Section ~\ref{sec:construction_pipeline_graph_maintenance}.

\subsection{Multi-Object Tracking and Interaction Understanding}\label{sec:applications_mot}
Beyond robust self-localization, dynamic world models serve as the substrate for understanding the agents and objects that populate the environment. 
In tightly-coupled SLAMMOT paradigm, the factor graph back-end estimates object trajectories jointly with the robot pose, providing smoothed, geometrically consistent tracks that outperform pure MOT systems which cannot benefit their estimates from a broader scene understanding. 
DynoSAM \cite{morris_dynosam_2025}, BodySLAM \cite{henning_bodyslam_2022} and AirDOS \cite{qiu2022airdos} generate 4D spatio-temporal maps of dynamic objects and agents as an output of their unified optimization, enabling downstream tasks such as trajectory prediction and collision forecasting.

For human agents specifically, BodySLAM \cite{henning_bodyslam_2022} and BodySLAM++ \cite{henning_bodyslam_2023} recover full SMPL body shape and pose jointly with camera trajectory (Section \ref{sec:dynamics_representations_agents}), producing a world model that captures postural state directly relevant to human-robot interaction and intention prediction. 
The structured prior provided by the parametric body model acts as a strong regularizer that improves both the human pose estimate and the camera trajectory, particularly in scenes where humans occlude most of the field of view and purely geometric features tracking fails. 
At the level of intentions and trajectories, Gorlo et al. \cite{gorlo_long-term_2024} use the temporal structure of 3D dynamic scene graph, e.g. recognizing that a person moving toward a doorway is likely to exit, rather than relying solely on kinematic history.
Aion \cite{catalano_aion_2025} extends this predictive capability to activity forecasting, building a model of human rhythms that lets the robot anticipate congestion before it occurs. 

Scene graphs enrich this tracking with semantic understanding of interactions, like Lost \& Found \cite{behrens_lost_2025} and its human-object interactions triggering updates at graph edge level, effectively building a running causal system for tracking how the scene has changed due to human activity without requiring a full scene reconstruction after each interaction.

\subsection{Navigation and Hierarchical Planning}\label{sec:applications_navigation}
Dynamic world models directly inform navigation and planning by providing structured, multi-resolution representations that support both metric path execution and high-level task decomposition.
The hierarchical layers of 3D scene graphs allows planners to operate appropriate level of abstraction for the current task, consistently reducing computation compared to planning over dense volumetric maps \cite{ray2024taskmotionplanninghierarchical} \cite{viswanathan2025spadescalablepathplanning}. Ejaz et al. \cite{ejaz2025situationallyawarepathplanningexploiting} demonstrate this computation efficiency through situationally-aware path planning over 3D scene graphs, using the relational structure of the graph to select traversal sequences that account for both metric cost and semantic context. 

The key advantage of advantage of factor graph-backed scene graph in this context is geometric consistency: because the scene graph nodes are grounded in a factor graph optimization, the metric positions used by the path planner are globally consistent and uncertainty-aware, avoiding the drift that would otherwise accumulate during long navigation tasks. 
SayPlan \cite{rana2023sayplangroundinglargelanguage} and Clio \cite{maggio_clio_2024} apply the task-driven pruning described in Section \ref{sec:querying_reasoning_action_action} to reduce planning complexity by orders of magnitude in large environments. 

In dynamic environments, however, static planning assumptions break down: a path computed over the current scene graph may become infeasible as agents move, objects are displaced, or passageways are shut. 
The connection between the factor graph dynamics layer and the planning layer remains largely open. 
Current systems either replan from scratch on graph updates or maintain pre-computed paths that are invalidated when dynamic nodes change \cite{ejaz2025situationallyawarepathplanningexploiting}, a limitation that Aion's predictive flow statistics (Section \ref{sec:dyn_world_model_representations_scene_graphs}) begin to address by routing around congestion proactively\cite{catalano_aion_2025}.

\subsection{Scene Understanding, Memory, and Long-Term Retrieval}\label{sec:applications_scene_understanding}
The fourth identified application area treats the dynamic world model as a persistent, queryable memory that supports object retrieval, semantic localization and long-term scene understanding.
DovSG \cite{yan_dynamic_2025} demonstrates this for mobile manipulation, where the open-vocabulary 3D scene graph serves as the robot's persistent object memory, supporting natural language queries for object retrieval even after objects have been displaced by human interaction. 

Belief Scene Graphs \cite{saucedo_estimating_2025} extend this memory to unobserved regions, transforming the world model from reactive record into a predictive spatial memory that can guide active exploration under partial observability.

\bigskip

Across all four application areas, the common thread is that the dynamic world model serves an active layer, not a passive geometric record but a probabilistically maintained, semantically structured representation that simultaneously supports estimation, prediction, planning and interaction. 
These works demonstrate that hybrid factor graph-scene graph representations are uniquely positioned to provide this layer, suggesting that hybrid representations can enable downstream applications that are difficult to achieve with either paradigm in isolation. 

\section{Evaluation Protocols, Datasets and Metrics}\label{sec:evaluation_metrics}

\begin{table*}[!htbp]
  \caption{Public Datasets and Benchmarks for Dynamic SLAM and Object Tracking}
  \label{tab:datasets_benchmarks}
  \centering
  \scriptsize
  \setlength{\tabcolsep}{3pt}
  \renewcommand{\arraystretch}{1.15}
  \begin{tabularx}{\textwidth}{@{}X X X X X X X@{}}
    \toprule
    \textbf{Dataset} & \textbf{Key Features} & \textbf{Environment} & \textbf{Dynamics} & \textbf{Use Case} & \textbf{Sensors} & \textbf{Used In} \\
    \midrule
    \midrule
    \multicolumn{7}{@{}c}{\textbf{I. Public Outdoor and Urban Driving Benchmarks}} \\
    \midrule
    KITTI Tracking/Odometry \cite{geiger_are_2012} & Grayscale and RGB stereo, GPS, IMU, 3D LiDAR (Velodyne HDL-64E) & Urban/Highway & Vehicles, pedestrians & High-speed dynamics, ego-motion ambiguity & GPS, IMU, 3D LiDAR, Camera & \cite{bescos_dynaslam_2021}, \cite{brink_using_2016}, \cite{morris_importance_2024}, \cite{tian_dl-slot_2024}, \cite{tian_dynaquadric_2024} \\
    nuScenes \cite{caesar_nuscenes_2019} & 3D MOT (AMOTA), urban LiDAR detections & Urban & High traffic density & Multi-object tracking, accuracy evaluation & LiDAR, Camera & \cite{zaech_learnable_2022} \\
    Virtual KITTI \cite{geiger_vision_2013} & Synthetic, photo-realistic, ground truth camera/object poses & Urban/Highway & Vehicles, pedestrians & Depth error isolation, object segmentation & Camera,  & \cite{henein_dynamic_2020} \\
    Victoria Park \cite{guivant_simultaneous_2002} & 2D odometry, landmark observations (trees), simulated northward movement & Urban & Corrupted landmarks & Solver divergence, robustness testing & Laser, GPS & \cite{lingzhu_xiang_robust_2015} \\
    Alcazar of Seville \cite{iros14-frog} & Visual-stereo, dynamic pedestrians & Indoor/Outdoor & Pedestrian dynamics & Re-localization, landmark stability & Laser, RGB & \cite{lingzhu_xiang_robust_2015} \\
    \midrule
    \midrule
    \multicolumn{7}{@{}c}{\textbf{II. Public Indoor and Semantic Mapping Datasets}} \\
    \midrule
    OpenLORIS-Scene \cite{shi_are_2019} & Office and supermarket environments, lifelong SLAM & Indoor & Pedestrian occlusion, scene changes & Long-term SLAM, furniture relocation & RGB-D, Semantic Labels & \cite{wang_semantic_2023}, \cite{ding_dynamic_2025} \\
    ScanNet \cite{dai_scannet_2017} & 1500+ RGB-D sequences, indoor spaces & Indoor & Static and dynamic objects & 3D scene graph construction, object relationships \& reconstruction & RGB-D, Semantic Labels & \cite{kim_3-d_2020} \\
    Matterport3D \cite{chang_matterport3d_2017} & 35 semantic classes, residential/commercial spaces & Indoor & Room layouts, objects & Learning co-occurrence, spatial distribution & RGB-D, 3D Models & \cite{saucedo_estimating_2025} \\
    TUM RGB-D \cite{sturm_benchmark_2012} & Office sequences, Kinect sensors, Vicon motion capture & Indoor & Walking, desk movements & High/Low dynamics, object tracking & RGB-D, Motion Capture & \cite{wang_semantic_2023} \\
    Bonn RGB-D Dynamic\cite{palazzolo2019iros} & Moving people interacting with moving objects & Indoor & Humans, Objects & Multi-object tracking Scene reconstruction & RGB-D  & \cite{cheng2022sg}\\
    Oxford Multimotion Dataset (OMD) \cite{judd_oxford_2019} & Swinging cubes, unconstrained motion & Indoor & Non-planar, fast-moving objects & Challenging dynamic tracking & Stereo, RGB-D, IMU, Vicon motion capture & \cite{gou_visual_2024}, \cite{morris_importance_2024}, \cite{morris_dynosam_2025} \\
    \midrule
    \midrule
    \multicolumn{7}{@{}c}{\textbf{III. Synthetic and Semi-Synthetic Environment Simulators}} \\
    \midrule
    TESSE Simulator \cite{ravichandran_bridging_2020} & Short-term dynamics, long-term changes, precise ground truth & Indoor & Walking, bouncing balls, furniture changes & Controlled ablation studies & RGB, Depth & \cite{schmid_khronos_2024} \\
    VirtualHome \cite{puig_virtualhome_2018} & Object movements in household routines, 50+ days & Indoor & Object movement in routine activities & Object tracking, daily activities & RGB & \cite{patel_proactive_2022} \\
    CARLA Simulator \cite{dosovitskiy_carla_2017} & Urban scene graph partitioning, weather dynamics & Urban & Traffic, pedestrian dynamics & Scene graph partitioning, dynamic environments & Camera, LiDAR & \cite{greve_collaborative_2024} \\
    Dynamic House Simulator / iGridson \cite{kurenkov_modeling_2023} & Household dynamic objects, movement between furniture & Indoor & Object movement, probabilistic dynamics & Link prediction, dynamic environments & RGB & \cite{kurenkov_modeling_2023} \\
    \midrule
    \midrule
    \multicolumn{7}{@{}c}{\textbf{IV. Specialized Benchmarks and Self-Collected Data}} \\
    \midrule
    VIODE \cite{minodaRAL2021} & UAV-based visual-inertial odometry, extreme motions, occlusion & Outdoor & UAV in parking lot, dynamic agents & Real-time update, occlusion handling & Synthetic ground truth & \cite{morris_dynosam_2025} \\
    TartanAir Shibuya \cite{tartanair2020iros}  & 30+ humans, dense urban crossing & Urban & Non-rigid, crowded scenes & Robustness in crowded environments & Ground truth poses & \cite{morris_dynosam_2025} \cite{qiu2022airdos} \\
    uHumans/uHumans2 \cite{rosinol_kimera_2021} & Unity-based simulator, 12-60 human agents, SMPL meshes & Indoor/Outdoor & Dense human tracking & Human interaction, dynamic tracking & Ground truth poses & \cite{rosinol_kimera_2021} \\
    FunGraph3D \cite{zhang_open-vocabulary_2025} & iPad Pro video, high-fidelity laser scans, Apple Vision Pro & Indoor/Outdoor & Functional relationships (e.g., switches) & Functional annotation, relationship tracking & Functional annotations & \cite{zhang_open-vocabulary_2025} \\
    OC-NaVQA \cite{gorlo_describe_2025} & Object-centric NaVQA, spatio-temporal Q\&A & Indoor/Outdoor & Static and dynamic environments & Question answering, memory sequences & Ground truth Q\&A labels & \cite{gorlo_describe_2025} \\
    Husky/Spot Field Tests & Real university halls/offices, LiDAR and RGB-D sensors & Indoor & Dynamic agents (robots, pedestrians), dynamic objects & Real-time performance validation & LiDAR, RGB-D & \cite{schmid_khronos_2024}, \cite{giberna_dynemo-slam_2025} \\
    \bottomrule
  \end{tabularx}
\end{table*}

\begin{table*}[!htbp]
\caption{Metric catalog for dynamic SLAM (FG) and scene graph (SG) evaluation. Use as a checklist and to standardize naming across encountered works.}
\label{tab:metrics_table}
\centering
\scriptsize
\setlength{\tabcolsep}{3pt}
\renewcommand{\arraystretch}{1.15}
\begin{tabularx}{\textwidth}{@{}l l X@{} r}
\toprule

\textbf{Metric} & \textbf{Type} & \textbf{What it measures / notes} & \textbf{Used in} \\

\midrule

\textbf{ATE (RMSE/median)} & Trajectory & 
Absolute trajectory error vs ground truth; report RMSE + median, and whether aligned (SE(3)/Sim(3)). 
& [...]\\
\textbf{RPE ($\Delta t$ / $\Delta s$)} & Trajectory & Relative pose error; clarifies short-term drift; specify interval. 
& [...]\\
\textbf{Drift (m/100m, deg/100m)} & Trajectory & 
Aggregate drift over distance; common in automotive datasets. 
& \cite{rosinol_kimera_2021}\\
\textbf{Relocalization success rate} & Trajectory & 
Fraction of trials where tracking is recovered after dropout/occlusion. 
& \cite{vega_torres_bim-slam_2023} \cite{wang_semantic_2023}\\
\textbf{CS-R (Correctness Score of Re-Localization)} & Trajectory &
Reliability of re-localization events in dynamic scenes; typically validates that a declared relocalization is correct (not a false positive). 
& \cite{wang_semantic_2023}\\
\textbf{Failure rate / time-to-failure} & System & 
Robustness under dynamics; define failure criterion (e.g. lost tracking $>$ N frames). 
& \cite{scherer_ogm2pgbm_2023}\\

\midrule

\textbf{Runtime (Hz/FPS), latency (ms)} & System & 
Throughput and per-module latency; note hardware and whether real-time. 
& \cite{behrens_lost_2025} \cite{bescos_dynaslam_2021} \cite{schmid_khronos_2024}\\
\textbf{CPU/GPU, memory footprint} & System & 
Resource profile for deployability comparisons. 
& \cite{ferraina_lost-3dsg_2026}\\

\midrule

\textbf{Point cloud Accuracy and Completeness / ICP residual} & Map & 
Geometric map quality vs reference (dense scan/mesh). 
& \cite{rosinol_kimera_2021} \cite{yan_dynamic_2025}\\
Map IoU (occupancy) & Map & Overlap between predicted and GT occupancy grids; specify resolution and thresholding. 
& \cite{rosinol_kimera_2021}\\
\textbf{TP rate (2D / BEV / 3D)} & Map &
True positive rate for occupancy estimation across viewpoints/representations; report the corresponding FP/FN definitions. 
& \cite{zhang_semantic-integrated_2026} \cite{bescos_dynaslam_2021}\\
\textbf{SCDA (Scene Change Detection Accuracy)} & Change &
Accuracy of detecting voxel-level or object-level changes over time; clarify temporal window and whether class-conditioned. 
& \cite{yan_dynamic_2025}\\
\textbf{Completeness / accuracy} & Map & 
Surface coverage or distance-to-surface statistics (often pairs). 
& \cite{rosinol_kimera_2021}\\

\midrule

\textbf{Mask IoU / mIoU} & Semantics & 
Segmentation quality for dynamic masking or semantic labels. 
& \cite{rosinol_kimera_2021}\\
\textbf{Precision/Recall/F1 (objects)} & Semantics & 
Detection/instance metrics; specify IoU threshold (e.g. 0.5). 
& \cite{schmid_khronos_2024}\\
\textbf{Motion Error (ME)} & Motion &
Frame-agnostic object motion error evaluated in the ground-truth object frame; helps compare methods using different reference frame conventions. 
& \cite{morris_dynosam_2025} \cite{morris_online_2025}\\

\midrule

\textbf{MOTA / AMOTA / MOTP} & Tracking &
Classic multi-object tracking metrics (association + localization); sensitive to thresholds and missed detections. 
& \makecell{\cite{gou_visual_2024} \cite{tian_dl-slot_2024} \cite{zaech_learnable_2022} \\ \cite{zhang_semantic-integrated_2026} \cite{gou_visual_2024} \cite{bescos_dynaslam_2021}} \\
\textbf{ID Switches} & Tracking &
Counts identity swaps in tracking; lower is better. 
& \cite{zhang_semantic-integrated_2026} \cite{zaech_learnable_2022}\\
\textbf{Fragments (track fragmentation)} & Tracking &
Counts interruptions/splits in a trajectory; measures continuity of tracks. 
& \cite{zhang_semantic-integrated_2026}\\
\textbf{IDF1} & Tracking &
Identity-based F1; balances correct associations over time. 
& \cite{zhang_semantic-integrated_2026}\\

\midrule

\textbf{Node Precision/Recall/F1} & Graph &
Correctness of entity discovery (rooms/walls/objects) in scene graphs; depends on matching policy between predicted and GT nodes. 
& \cite{zhang_open-vocabulary_2025} \cite{ferraina_lost-3dsg_2026}\\
\textbf{Edge Precision/Recall/F1 / edge prediction/accuracy} & Graph &
Correctness of predicted relations (adjacent-to, in-room, supports, etc.); report relation set and matching rules. 
& \cite{zhang_open-vocabulary_2025}\\
\textbf{Spurious \& Missing Entities} & Graph &
Human-judgment metric counting incorrect (spurious) or omitted (missing) nodes/edges in constructed SGs. 
& \cite{kim_3-d_2020}\\
\textbf{Precision/Recall of scene parsing} & Parsing &
Accuracy of partitioning places into rooms or higher-level structures (often computed from segmentation or region adjacency). 
& \cite{rosinol_3d_2020}\cite{rosinol_kimera_2020}\\
\textbf{Graph Edit Distance (GED)} & Graph &
Structural similarity to GT graph; sensitive to node/edge labels and matching. 
& \cite{gorlo_long-term_2024}\\
\textbf{Temporal stability / churn} & Graph-time &
Persistence of nodes/edges over time (e.g., additions/deletions per minute, or lifetime distributions). 
& \cite{catalano_aion_2025} \\

\midrule

\textbf{NDCG (Normalized Discounted Cumulative Gain)} & Ranking &
Quality of predicted rankings (e.g., likelihood-ranked object locations); requires relevance labels and a cutoff $K$. 
& \cite{kurenkov_modeling_2023}\\

\midrule

\textbf{NEES / $\chi^2$ consistency} & Prob. (FG) &
Consistency of estimated uncertainty; important when works claim probabilistic reliability. 
& \cite{morris_importance_2024}\\

\midrule
\midrule

\textbf{Descriptive question accuracy (SQA)} & Reasoning &
Accuracy on spatio-temporal question answering using 4D/temporal scene graphs. 
& \cite{gorlo_describe_2025}\\
\textbf{ADE (Avg. Displacement Error)} & Prediction &
Average distance between predicted and GT trajectories (e.g., human motion). 
& \cite{gorlo_long-term_2024}\\
\textbf{BoN ADE (Best-of-N ADE)} & Prediction &
Minimum ADE over $N$ sampled predictions; measures multi-modal forecasting quality. 
& \cite{gorlo_long-term_2024}\\
\textbf{Task success rate} & Task &
Completion rate of robot tasks informed by the world model (e.g., pick-and-place); specify task definition and failure modes. 
& \cite{yan_dynamic_2025}\\
\textbf{BLEU / METEOR / BERTScore} & Language &
Similarity of generated textual SG annotations to references; report whether multiple references are used and tokenization details. 
& \cite{linok_dygenc_2025}\\

\bottomrule
\end{tabularx}
\end{table*}

\subsection{Protocols}\label{sec:evaluation_metrics_protocols}
A thorough analysis of the evaluation protocols reveals a rigorous methodological framework designed to validate the robustness of world models against non-stationary environments.
These protocols go beyond traditional SLAM metrics, including stochastic analysis, longitudinal studies and modular ablations to isolate the effect of various architectural components. 
Moreover, research in dynamic SLAM has shifted towards validating the efficacy of the system in and for specific use cases.

\subsubsection{Per-Sequence and Multi-Run Evaluation}
To address the inherent non-determinism in multi-threaded systems and the randomness introduced by dynamic entities, researchers employ a protocol of repeated trials:

\begin{itemize}
    \item \textbf{Per-Sequence Analysis}: Results are almost universally reported on a per-sequence basis to account for varying environmental conditions as traffic density and feature availability \cite{tian_dl-slot_2024} \cite{giberna_dynemo-slam_2025} \cite{zhang_semantic-integrated_2026}.
    \item \textbf{Stochasticity Management}: To mitigate the impact of random initialization in particle filters or the non-deterministic nature of multi-threaded optimization, experiments are executed multiple times.
    Standard protocols include running each sequence multiple times and reporting the median value \cite{zhang_semantic-integrated_2026} or average \cite{scherer_ogm2pgbm_2023}.
    Other collaborative frameworks report mean and standard deviation over several randomly initialized runs to quantify stability \cite{greve_collaborative_2024}.
\end{itemize}

\subsubsection{Dynamic Severity Sweeps}
Evaluation protocols frequently involve sweeping through levels of environmental complexity to test the limits of dynamic robustness.

\begin{itemize}
    \item \textbf{Entity Density Scalability}: The uHumans and uHumans2 datasets are used to evaluate how spatial perception engines handle environments with increasing numbers of dynamic entities, ranging from 12 to 60 human agents \cite{rosinol_3d_2020}. Similar approaches are partaken when the dynamics in the scene are displaced objects rather than continuously moving humans \cite{ferraina_lost-3dsg_2026}. Also self-collected datasets are generated accounting for different progressive levels, densities and complexities of dynamic entities populating the scene \cite{giberna_dynemo-slam_2025}.
    \item \textbf{Dynamic Entity Types}: Taking into account that different kinds of dynamics can occur in the scene, e.g. persistently moving agents and occasionally displaced objects, the systems' generalization capabilities are often evaluated testing for subsets or different configurations of possible dynamics occurring in the scene \cite{giberna_dynemo-slam_2025}.
    \item \textbf{Clutter and Deviation Levels}: In BIM-based localization, scenarios are designed to progressively increase the level of clutter, thereby decreasing the overlap with permanent building objects \cite{vega_torres_bim-slam_2023} \cite{shaheer_tightly_2025}.
    \item \textbf{Controlled Configuration Sweeps}: Navigation models are tested against varying obstacles configurations to ensure that learned policies generalize across different collision trajectories \cite{maguire_lidar-only_2025}. 
\end{itemize}

\subsubsection{Relocalization Dropouts and Success Rate}
The robustness of the world model is tested through its ability to recover from tracking loss, particularly when dynamic occlusion causes feature dropouts. 

\begin{itemize}
    \item \textbf{Sequential Retrieval Tests}: Systems are evaluated by sequentially feeding data from different session and counting the number of successful against unsuccessful relocalization events \cite{wang_semantic_2023}.
    \item \textbf{Convergence Metrics}: Success is often defined by a threshold-based protocol, where a method is deemed to have converged if the pose estimate, either of the exploring robot's trajectories or detected entities, stabilizes within a specific distance of the ground truth within a set percentage of the sequence length\cite{vega_torres_bim-slam_2023} \cite{morris_importance_2024}.
    \item \textbf{Initialization and Hand-off}: Some protocols evaluate the efficiency of switching from stochastic global localizer to a precise graph-based localization algorithm once a covariance threshold is met \cite{scherer_ogm2pgbm_2023}.
\end{itemize}

\subsubsection{Multi-Session and Long-Term Change Protocols}
Lifelong autonomy is validated though longitudinal data collection and multi-pass traversal.

\begin{itemize}
    \item \textbf{Long-Term Monitoring}: Systems are evaluated on their ability to update object probabilities across extensive periods \cite{gomez_object-based_2020} \cite{patel_proactive_2022}.
    \item \textbf{Incremental Pass Validation}: Protocols include repeated traversals of the same environment to determine if the model can successfully ``forget'' ephemeral data and identify static static structural components \cite{walcott-bryant_dynamic_2012}.
    \item \textbf{Consecutive Task Execution}: Robots are tested on their ability to complete consecutive manipulation tasks without manual resets, ensuring the updated 3D scene graph remains actionable after manual modification of the environment \cite{yan_dynamic_2025}.
\end{itemize}

\subsubsection{Systematic Ablation Studies}
Ablation protocols are essential for isolating the contribution of proposed factors, modules and policies.

\begin{itemize}
    \item \textbf{Dynamic Filtering Modules}: Protocols isolate the dynamic entities or features removal module to identify the best trade-off for loop closure quality \cite{giberna_dynemo-slam_2025}.
    \item \textbf{Scene Graph Update Policies}: Researchers ablate the logic used for graph maintenance, such as the effectiveness of \textit{sequential reasoning} in functional graphs \cite{zhang_open-vocabulary_2025} or the utility of \textit{region clustering} and \textit{frame selection heuristics} \cite{gorlo_describe_2025} or \textit{keyframe selection policies} \cite{giberna_dynemo-slam_2025} in 4D scene graph generation.
    \item \textbf{Factor-Level Ablations}: In tightly-coupled systems, models are evaluated with and without specific constraints, such as \textit{non-rigid factors} for humans or \textit{rigid factors} for objects, to verify their individual contributions to trajectory accuracy and world model generation \cite{giberna_dynemo-slam_2025} \cite{ding_dynamic_2025}.
\end{itemize}

\subsection{Datasets}\label{sec:evaluation_metrics_datasets}
The evaluation of hybrid world models, integrating scene graphs and/or factor graphs, and spatio-temporal dynamics, relies on a hierarchical selection of commonly employed datasets ranging from established autonomous driving benchmarks to specialized indoor semantic simulator.
Additionally, to assess the system behavior on specific situations, self-collected datasets presenting peculiar kinds of environments and dynamics are created, either in real world or simulation. 
These datasets can be categorized by their environmental focus, sensor suites, and the specific nature of the dynamics they encapsulate. 
Table \ref{tab:datasets_benchmarks} reports the most commonly used publicly available datasets, benchmarks and simulators. 

\subsection{Metrics}\label{sec:evaluation_metrics_metrics}
The metrics employed to assess any proposed method depend on the downstream use of it and its tackled problem.
When it comes to the evaluation of GMs, either factor graphs or scene graphs, specifically developed for dynamic environments, the attention is focused on assessing the systems performance on different levels: the robot/self trajectory, the map geometry, the semantics detection and segmentation, the graph structure, the system performance in terms of runtime and memory usage, and the probabilistic consistency of the graph. 
Additionally, when SGs are exploited or augmented with the employment of LLM methods for task accomplishment, capabilities like language semantic reasoning and downstream task success are evaluated. 
Table \ref{tab:metrics_table} reports the most commonly used metric in these scenarios.

\section{Comparative Analysis of Representative Systems}\label{sec:taxonomy}
The literature on world modeling in dynamic environments spans a wide range of representational choices, inference strategies and application targets.
While individual systems are often described in isolation, a comparative perspective is necessary to understand how these design decisions relate to one another and how different combinations of graphical models, dynamic handling, and uncertainty treatments emerge in practice. 
This section provides a structured comparison of the surveyed \emph{systems} themselves, classifying graph-based dynamic world models for robotics along a set of common axes.
Rather than ranking methods by performance, the comparison aims to map the design space, highlighting recurring architectural patterns and characteristic trade-offs across factor-graph-based, scene-graph-based and hybrid representations.

\subsection{Axes of Comparison}\label{sec:taxonomy_axes}
The analysis organizes existing approaches along several dimensions that recur throughout the surveyed literature:
\begin{itemize}
    \item \textbf{Graph Stack}: the primary representational backbone employed by the system, distinguishing between factor graphs (FG), scene graph (SG), and hybrid architectures in which the two are explicitly coupled;
    \item \textbf{Dynamics Type}: the class of environmental change modeled, using the same three-tier Temporal Horizon distinction introduced in Section \ref{sec:taxonomy_dynamics} (short-term dynamics, semi-static changes, and long-term structural evolution);
    \item \textbf{Represented Scene}: the operational context, such as indoor, outdoor, or any environment;
    \item \textbf{Data Association Strategy}: the mechanism used to link observations over time, ranging from deterministic matching to probabilistic multi-hypothesis methods or learned similarity models;
    \item \textbf{Inference Policy}: the underlying inference method, including batch maximum a posteriori (MAP) optimization, incremental smoothing, filtering-based approaches, or variational inference;
    \item \textbf{Uncertainty Treatment}: whether uncertainty is modeled explicitly in a probabilistic sense or handled through heuristic confidence measures;
    \item \textbf{Sensor}: the employed sensing inputs used to build the world model;
    \item \textbf{System Properties}: practical characteristics such as real-time capability, open-vocabulary support, and multi-robot operation.
    \item \textbf{Downstream Usage}: the primary tasks specifically enabled or targeted by the representation, such as robust SLAM, navigation, manipulation, interaction understanding, or prediction. 
\end{itemize}
These axes stem from the three research questions posed in Section \ref{sec:introduction}: Graph Stack, Dynamics Type, and Represented Scene characterize \emph{what is represented} (RQ1); Data Association Strategy, Inference Policy, Uncertainty Treatment, and Sensor  characterize \emph{how the representation is constructed and maintained} (RQ2); and System Properties, and Downstream Usage characterize \emph{how the representation is exploited} (RQ3). 
Together, these axes provide a unified lens through which heterogeneous systems can be compared across all three questions simultaneously, without assuming a notion of optimality.

\subsection{Comparison of Existing Approaches}\label{sec:taxonomy_taxonomy}
Table \ref{tab:taxonomy_overview} summarizes representative works in the literature according to the comparison dimensions defined above.
The listed systems are grouped by their underlying graph stack, enabling a direct comparison between factor-graph-based methods, scene-graph-centric approaches, and hybrid representations that integrate probabilistic estimation with semantic abstraction.

\begin{table*}[!ht]
\caption{Comparative analysis of representative probabilistic graph-based
\emph{systems} for robotics in dynamic environments.}
\label{tab:taxonomy_overview}
\centering

\begingroup
\fontsize{6.5}{7.4}\selectfont
\setlength{\tabcolsep}{1.2pt}
\renewcommand{\arraystretch}{1.08}
\setlength{\extrarowheight}{0pt}
\setlength{\emergencystretch}{1em}

\begin{adjustbox}{max width=0.8\textwidth}
\begin{tabular}{@{}
  L{1.80cm}  
  L{0.70cm}  
  L{1.55cm}  
  L{1.50cm}  
  L{1.50cm}  
  L{0.85cm}  
  L{1.45cm}  
  L{1.35cm}  
  L{1.35cm}  
  L{0.95cm}  
  C{0.35cm}  
  C{0.35cm}  
  C{0.35cm}  
  C{1.80cm}  
@{}}
    \toprule
    & \multicolumn{5}{c}{\textbf{Representation}} &
    \multicolumn{4}{c}{\textbf{Construction \& Maintenance}} &
    \multicolumn{4}{c}{\textbf{Exploitation}} \\
    \cmidrule(lr){2-6} \cmidrule(lr){7-10} \cmidrule(lr){11-14}
    \textbf{Work} & \textbf{Graph stack} & \textbf{Nodes} & \textbf{Edges} & \textbf{Dynamics type} &
    \textbf{Represented scene} &
    \textbf{Association} & \textbf{Inference} & \textbf{Uncertainty} & \textbf{Sensors} &
    \textbf{RT} & \textbf{OV} & \textbf{MR} & \textbf{Downstream use} \\
    \midrule

    \multicolumn{14}{@{}c}{\textit{Factor graph (FG) only}} \\
    \midrule

    \wcell{DynaSLAM II\\Bescos et al.\\ (2021)\cite{bescos_dynaslam_2021}} &
    \wcell{FG} &
    \wcell{Camera pose,\\object pose, \\velocity,\\points} &
    \wcell{reprojection \\error,\\motion error} &
    \wcell{short-term \cmark\\semi-static \xmark\\lifelong \xmark} &
    Outdoor &
    \wcell{deterministic\\(ORB + IoU)} &
    \wcell{batch\\MAP} &
    \wcell{explicit\\covariances} &
    \wcell{Stereo or\\RGB-D} &
    \cmark & \xmark & \xmark &
    \wcell{Collision \\avoidance\\+ SLAMMOT} \\
    \midrule

    \wcell{Brink\\(2016)\cite{brink_using_2016}} &
    \wcell{FG} &
    \wcell{Image features,\\velocity} &
    \wcell{same-object\\probabilistic\\ dependency} &
    \wcell{short-term \cmark\\semi-static \cmark\\lifelong \xmark} &
    Any &
    \wcell{deterministic\\(Delaunay\\ triangulation)} &
    \wcell{incremental\\loopy BP} &
    \wcell{explicit\\Gaussian} &
    \wcell{Stereo} &
    \xmark & \xmark & \xmark &
    \wcell{Dynamic\\ object\\detection} \\
    \midrule

    \wcell{DynoSAM\\Morris et al.\\(2025)\cite{morris_dynosam_2025}} &
    \wcell{FG} &
    \wcell{Camera poses,\\object \\motions/poses,\\map points} &
    \wcell{point meas.,\\ odometry,\\object motion,\\ obj. smoothing,\\prior factors} &
    \wcell{short-term \cmark\\semi-static \xmark\\lifelong \xmark} &
    Any &
    \wcell{optical flow\\+ instance seg.} &
    \wcell{incremental\\smoothing} &
    \wcell{explicit\\covariances} &
    \wcell{RGB-D} &
    \cmark & \xmark & \xmark &
    \wcell{4D dynamic\\reconstruction} \\
    \midrule

    \wcell{Gou et al.\\(2024)\cite{gou_visual_2024}} &
    \wcell{FG} &
    \wcell{Camera poses,\\landmarks,\\ objects} &
    \wcell{static reproj.,\\ camera meas.,\\dynamic reproj.,\\ object motion} &
    \wcell{short-term \cmark\\semi-static \xmark\\lifelong \xmark} &
    Any &
    \wcell{Hungarian\\algorithm} &
    \wcell{local/global\\BA} &
    \wcell{explicit\\Gaussian} &
    \wcell{RGB-D} &
    \cmark & \xmark & \xmark &
    \wcell{Industrial\\inspection} \\
    \midrule

    \wcell{Gomez et al.\\(2020)\cite{gomez_object-based_2020}} &
    \wcell{FG} &
    \wcell{Poses,\\ objects\\(cuboid\\ centroids)} &
    \wcell{Geometric\\ spatial\\ relations} &
    \wcell{short-term \xmark\\semi-static \cmark\\lifelong \xmark} &
    Any &
    \wcell{area of\\influence} &
    \wcell{multi-session\\adaptation} &
    \wcell{explicit\\prob. presence} &
    \wcell{RGB-D} &
    \cmark & \xmark & \xmark &
    \wcell{Autonomous\\ indoor\\service robots} \\
    \midrule

    \wcell{Zaech et al.\\(2022)\cite{zaech_learnable_2022}} &
    \wcell{FG} &
    \wcell{Object\\ observations,\\object tracks} &
    \wcell{data\\association} &
    \wcell{short-term \cmark\\semi-static \xmark\\lifelong \xmark} &
    Any &
    \wcell{learned\\ matching\\(NMP)} &
    \wcell{filtering\\(Kalman)} &
    \wcell{explicit\\(prob. edges)} &
    \wcell{LiDAR} &
    \cmark & \xmark & \xmark &
    \wcell{3D multi-object\\tracking} \\
    \midrule

    \wcell{BodySLAM\\Henning et al.\\(2022)\cite{henning_bodyslam_2022}} &
    \wcell{FG} &
    \wcell{Camera poses,\\human centre\\poses, SMPL\\body params} &
    \wcell{reprojection\\error, human\\motion model\\(temporal)} &
    \wcell{short-term \cmark\\semi-static \xmark\\lifelong \xmark} &
    Any &
    \wcell{OpenPose\\keypoints} &
    \wcell{batch\\MAP} &
    \wcell{explicit\\covariances} &
    \wcell{Monocular} &
    \xmark & \xmark & \xmark &
    \wcell{Human pose\\+ camera\\trajectory\\estimation} \\
    \midrule

    \wcell{BodySLAM++\\Henning et al.\\(2023)\cite{henning_bodyslam_2023}} &
    \wcell{FG} &
    \wcell{Camera poses,\\IMU states,\\human pose\\+shape(SMPL)} &
    \wcell{repr. error,\\ IMU factors,\\human motion\\constraints} &
    \wcell{short-term \cmark\\semi-static \xmark\\lifelong \xmark} &
    Indoor &
    \wcell{learned\\(HMR)} &
    \wcell{incremental\\smoothing\\(OKVIS2)} &
    \wcell{explicit\\covariances} &
    \wcell{Monocular\\+ IMU} &
    \cmark & \xmark & \xmark &
    \wcell{Real-time\\human +\\camera state\\estimation} \\
    \midrule

    \wcell{AirDOS\\Qiu et al.\\(2022)\cite{qiu2022airdos}} &
    \wcell{FG} &
    \wcell{Camera poses,\\object motion,\\rigid part\\structure\\(3D points)} &
    \wcell{rigidity and\\motion\\constraints,\\reprojection\\error} &
    \wcell{short-term \cmark\\semi-static \xmark\\lifelong \xmark} &
    Outdoor &
    \wcell{ORB features\\+ instance\\segmentation} &
    \wcell{batch\\MAP} &
    \wcell{explicit\\covariances} &
    \wcell{Stereo} &
    \xmark & \xmark & \xmark &
    \wcell{4D spatio-\\temporal maps,\\crowded\\urban SLAM} \\
    \midrule

    \multicolumn{14}{@{}c}{\textit{Hybrid (FG + SG)}} \\
    \midrule

    \wcell{DynEMo-SLAM\\Giberna et al.\\(2025)\cite{giberna_dynemo-slam_2025}} &
    \wcell{Hybrid} &
    \wcell{Keyframes,\\ entity obs.,\\planes, floor} &
    \wcell{KF-entity,\\intra-entity,\\floor-entity} &
    \wcell{short-term \cmark\\semi-static \cmark\\lifelong \xmark} &
    Indoor &
    \wcell{deterministic\\(AprilTags)} &
    \wcell{--} &
    \wcell{explicit\\information matrix} &
    \wcell{LiDAR +\\camera} &
    \cmark & \xmark & \xmark &
    \wcell{Metric-semantic\\reconstruction in\\crowded envs.} \\
    \midrule

    \wcell{Kimera\\Rosinol et al.\\(2021)\cite{rosinol_kimera_2021}} &
    \wcell{Hybrid} &
    \wcell{Free space,\\agents/objects,\\rooms,\\ building} &
    \wcell{Semantic\\ relations} &
    \wcell{short-term \cmark\\semi-static \xmark\\lifelong \xmark} &
    Indoor &
    \wcell{TEASER++\\+ seg.} &
    \wcell{iSAM2 +\\batch parsing} &
    \wcell{explicit\\(Bayesian)} &
    \wcell{Vis.-inertial\\+ RGB-D} &
    \cmark & \xmark & \cmark &
    \wcell{Hierarchical\\ metric-semantic\\reconstruction \\+ planning} \\
    \midrule

    \wcell{Khronos\\Schmid et al.\\(2024)\cite{schmid_khronos_2024}} &
    \wcell{Hybrid} &
    \wcell{Poses, objects,\\object\\ fragments} &
    \wcell{associations,\\odometry} &
    \wcell{short-term \cmark\\semi-static \cmark\\lifelong \xmark} &
    Indoor &
    \wcell{volumetric IoU\\(multi-hyp.)} &
    \wcell{async global\\optimization} &
    \wcell{explicit\\4D metric} &
    \wcell{RGB-D} &
    \cmark & \cmark & \xmark &
    \wcell{Lifelong\\ spatio-temporal\\maps} \\
    \midrule

    \multicolumn{14}{@{}c}{\textit{Scene graph (SG) only}} \\
    \midrule

    \wcell{Lost \& Found\\Behrens et al.\\ (2025)\cite{behrens_lost_2025}} &
    \wcell{SG} &
    \wcell{Objects,\\ object parts,\\points} &
    \wcell{Semantic\\ spatial\\ relations} &
    \wcell{short-term \cmark\\semi-static \cmark\\lifelong \xmark} &
    Indoor &
    \wcell{image-based\\detection} &
    \wcell{visual point\\tracking} &
    \wcell{heuristic\\confidence} &
    \wcell{RGB-D} &
    \cmark & \cmark & \xmark &
    \wcell{Object\\ tracking +\\manipulation} \\
    \midrule

    \wcell{DovSG\\Yan et al.\\(2025)\cite{yan_dynamic_2025}} &
    \wcell{SG} &
    \wcell{Objects\\ (labels,\\color,\\ material)} &
    \wcell{Semantic\\ spatial\\ relations} &
    \wcell{short-term \cmark\\semi-static \cmark\\lifelong \xmark} &
    Indoor &
    \wcell{similarity-based\\(CLIP features)} &
    \wcell{online logic\\updates} &
    \wcell{heuristic\\confidence} &
    \wcell{RGB-D} &
    \cmark & \cmark & \xmark &
    \wcell{Mobile\\ manipulation\\+ task\\ planning} \\
    \midrule

    \wcell{CURB-SG\\Greve et al.\\(2024)\cite{greve_collaborative_2024}} &
    \wcell{SG} &
    \wcell{Environment,\\roads and\\ intersections,\\landmarks} &
    \wcell{spatial\\ relations} &
    \wcell{short-term \cmark\\semi-static \xmark\\lifelong \xmark} &
    Outdoor &
    \wcell{scan match.\\+ local search} &
    \wcell{pose graph\\opt. (PGO)} &
    \wcell{explicit\\(fitness score)} &
    \wcell{LiDAR} &
    \cmark & \xmark & \cmark &
    \wcell{Automated\\driving} \\
    \midrule

    \wcell{Aion\\Catalano et al.\\(2025)\cite{catalano_aion_2025}} &
    \wcell{SG} &
    \wcell{Entity obs.,\\navigational\\ flows, semantic\\ abstraction} &
    \wcell{spatial/semantic\\ rels., planning\\ connectivity,\\traversal costs} &
    \wcell{short-term \cmark\\semi-static \xmark\\lifelong \xmark} &
    Indoor &
    \wcell{pos.-invariant\\indexing} &
    \wcell{FreMEn\\(frequency)} &
    \wcell{explicit\\probabilistic} &
    \wcell{RGB-D /\\Stereo} &
    \cmark & \xmark & \xmark &
    \wcell{Forecasting\\ human\\activity for\\ navigation} \\
    \midrule

    \wcell{Functional SG\\Zhang et al.\\(2025)\cite{zhang_open-vocabulary_2025}} &
    \wcell{SG} &
    \wcell{Objects and\\interactive\\ elements} &
    \wcell{functional\\relationships} &
    \wcell{short-term \xmark\\semi-static \cmark\\lifelong \xmark} &
    Indoor &
    \wcell{learned\\matching} &
    \wcell{sequential\\LLM reasoning} &
    \wcell{learned\\confidence} &
    \wcell{RGB-D} &
    \xmark & \cmark & \xmark &
    \wcell{3D VQA\\+ manipulation} \\
    \midrule

    \wcell{Saucedo et al.\\(2025)\cite{saucedo_estimating_2025}} &
    \wcell{SG (BSG)} &
    \wcell{Building,\\ rooms,\\objects} &
    \wcell{Semantic and\\ geometric\\ spatial rel.} &
    \wcell{short-term \xmark\\semi-static \cmark\\lifelong \xmark} &
    Indoor &
    \wcell{learned\\matching} &
    \wcell{variational\\(GCN pred.)} &
    \wcell{explicit\\probabilistic} &
    \wcell{RGB-D} &
    \xmark & \cmark & \xmark &
    \wcell{Unvisited\\ scene\\completion}\\
    \bottomrule
    \end{tabular}
\end{adjustbox}

\vspace{2pt}

{\scriptsize
\begin{tabularx}{\textwidth}{@{}l >{\raggedright\arraybackslash}X@{}}
  \textbf{Legend:} &
  \textbf{Graph stack}: FG (factor graph), SG (scene graph),
  Hybrid (FG-backed SG / SG-constrained FG).
  \textbf{Dynamics type}: (i) short-term, (ii) semi-static,
  (iii) lifelong/structural change.
  \textbf{Represented Scene}: Indoor, Outdoor, Any.
  \textbf{Association}: data association
  (deterministic, MHT, JPDA, learned matching).
  \textbf{Inference}: batch MAP, incremental smoothing,
  filtering, variational, particle/mixture.
  \textbf{Uncertainty}: explicit probabilistic treatment
  versus heuristic confidence only.
  \textbf{RT/OV/MR}: real-time / open-vocabulary / multi-robot.
\end{tabularx}
}

\endgroup
\end{table*}

The table illustrates the variety of design choices explored, from tightly-coupled SLAMMOT systems that extend classical factor graphs with dynamic entity modeling, to scene-graph-based frameworks emphasizing semantic structure and interaction, finally to hybrid systems that combine geometric consistency with relational reasoning.

\subsection{Cross-Comparison}\label{sec:taxonomy_crosscomparison}
\begin{figure}
    \centering
    \includegraphics[width=0.9\columnwidth]{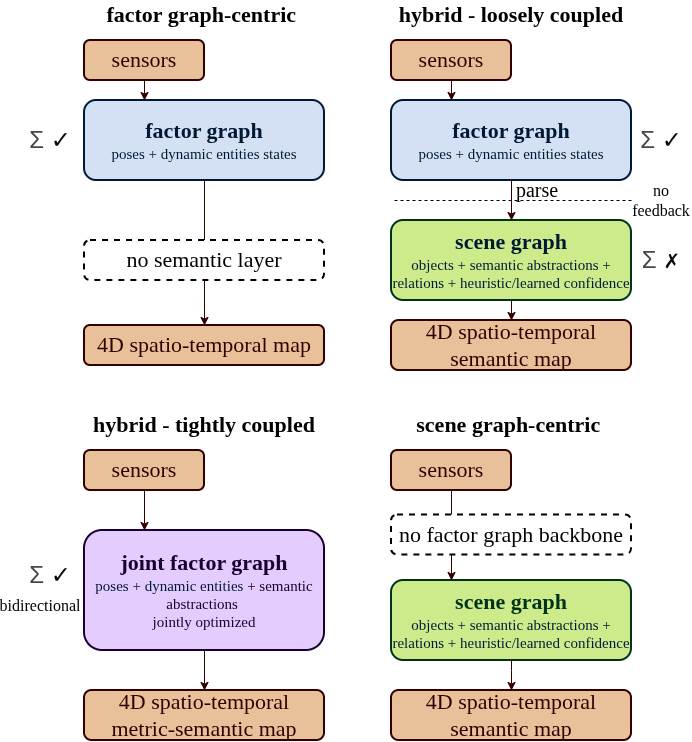}
    \caption{Four distinct architectures have emerged throughout the literature for representing dynamic environment, specifically \textit{factor graph-centric} (top left), \textit{scene graph-centric} (bottom right), \textit{hybrid - loosely coupled} (top right) and \textit{hybrid - tightly-coupled} (bottom left).}
    \label{fig:architectural_patterns}
\end{figure}

Across the surveyed works, four recurring architectural patterns emerge, showed in Figure \ref{fig:architectural_patterns}, distinguished primarily by which graph stack a system commits to and, for hybrid systems, how tightly the two stacks are coupled. 
Two patterns rely on a single representational backbone, factor graphs alone or scene graphs alone, while two combine both, differing in whether information flows in one direction or both.

\paragraph{Tightly-Coupled SLAMMOT (FG-centric)}
Systems in this pattern, like DynaSLAM II \cite{bescos_dynaslam_2021}, DynoSAM \cite{morris_dynosam_2025}, AirDOS \cite{qiu2022airdos}, BodySLAM \cite{henning_bodyslam_2022}, BodySLAM++ \cite{henning_bodyslam_2023}, Gou et al. \cite{gou_visual_2024}, adopt the joint-optimization approach detailed in Section \ref{sec:dynamics_representations_agents}, treating dynamic entities as first-class factor graph variables rather than outliers.
The trade-off is that this template primarily addresses continuously moving agents, while semi-static displacement and lifelong structural changes are typically outside its scope, and no explicit semantic abstraction layer is provided for higher-level reasoning tasks.

\paragraph{Layered Stack with FG Backbone (Hybrid, loosely coupled)}
Kimera \cite{rosinol_kimera_2021} and Hydra \cite{hughes_hydra_2022} exemplify a layered architecture in which a factor graph optimizer provides a globally consistent geometric foundation, while a separate scene graph parser constructs the semantic-relational superstructure. 
The advantage is modularity: the scene graph logic can be developed independently of the SLAM back-end. 
The disadvantage is one-directional information flow: the scene graph is informed by the factor graph, but semantic information rarely propagates back to improve localization, so geometric uncertainty never reaches the semantic layer.

\paragraph{Unified Optimizable Graph (Hybrid, tightly-coupled)}
S-Graphs \cite{bavle_s-graphs_2025} and DynEMo-SLAM \cite{giberna_dynemo-slam_2025} incorporate semantic entities directly as variables and factors within the same factor graph that optimizes robot poses. 
This enables bidirectional information flow: semantic structural priors constrain ego-trajectory estimation, while geometric observations refine semantic entity states. 
DynEMo-SLAM extends this template to dynamic entities by constraining entity pose variables with semantic motion priors and structural relations. 
The pattern is particularly expressive in environments where structural semantics are stable and informative, while its applicability to large-scale, dynamic outdoor settings, where the number and identity of semantic entities grow rapidly, has not yet been demonstrated.

\paragraph{Belief-Augmented and Dynamic Scene Graph (SG-centric)}
DovSG \cite{yan_dynamic_2025}, Lost \& Found \cite{behrens_lost_2025}, BSG \cite{saucedo_estimating_2025}, Aion \cite{catalano_aion_2025}, and Functional Scene Graphs \cite{zhang_open-vocabulary_2025} emphasize the scene graph as a persistent and explicitly dynamic world model, maintained primarily through semantic reasoning and graph-level updates rather than joint probabilistic optimization.
Persistence, forgetting, and state transition are central to this pattern: Lost \& Found distinguishes object removal from occlusion through the containment mechanism described in Section \ref{sec:dyn_world_model_representations_scene_graphs}.
Aion further extends this paradigm by encoding temporal activity patterns directly into the scene graph structure. 
These approaches excel at long-term object permanence, natural language grounding, and interaction aware updates. 
Their reliance on heuristic or learned confidence measures, rather than trajectory level covariances, reflects a different trade-off between semantic expressiveness and probabilistic state estimation.

\subsection{Trends in Graph-based World Modeling}\label{sec:taxonomy_trends}
Three directional trends cut across the surveyed approaches and collectively reshape the design space of probabilistic world modeling in dynamic environments.

\paragraph{From Closed-Vocabulary to Open-Vocabulary Association} 
Early dynamic SLAM systems defined a fixed ontology of entities types at design type, like pedestrians, vehicles, or furniture, and relied on classifiers trained on these predefined categories.
More recent systems increasingly leverage vision-language embeddings, including CLIP-based representations and vision-language models, to support open-vocabulary perception and association. 
Frameworks such as DovSG \cite{yan_dynamic_2025} and ConceptGraphs \cite{gu_conceptgraphs_2023} demonstrate that open-set object discovery and association can be integrated into real-time robotic systems, enabling world models to accommodate previously unseen categories without retraining.

\paragraph{From Reactive Maps to Predictive Spatial Memory}
Early dynamic world models were predominantly reactive, updating internal representations only in response to incoming observations and answering queries about the current state of the environment.
A clear trend in more recent work is the incorporation of predictive components that encode expectations about unobserved entities and future activity, of which Aion's flow statistics (discussed below) are a representative case\cite{catalano_aion_2025}.
Belief Scene Graphs' blind nodes (Section \ref{sec:dyn_world_model_representations_scene_graphs}) similarly estimate likely object locations from learned co-occurrence patterns \cite{saucedo_estimating_2025}, while Gorlo et al. \cite{gorlo_long-term_2024} leverage relational scene context to forecast long-term human trajectories.
Altogether, these approaches extend world models from reactive state estimates toward representations that encode spatio-temporal patterns. 

\paragraph{From Static Hierarchy to 4D Spatio-Temporal Representations}
Conventional 3D scene graph formulations organize the environment into a spatial hierarchies, such as objects, places, rooms, and buildings, that are largely static in their structural definition.
Recent work has shifted from this assumption by explicitly incorporating temporal information into the representation\cite{schmid_khronos_2024}.
Aion\cite{catalano_aion_2025} is the clearest instance of this shift: by attaching time-dependent flow statistics directly to navigational nodes, it extends the scene graph into a four-dimensional representation that captures not only spatial structure but temporal patterns of activity.
This shift reflects a broader trend toward world models that encode when and how frequently interactions occur, rather than treating temporal variation as a sequence of independent snapshots.

\section{Open Challenges and Future Directions}\label{sec:challenges_future_works}

The works surveyed throughout this paper collectively demonstrate that probabilistic graph-based representations, i.e. combining factor graphs for geometric consistency with scene graphs for semantic abstraction, form a powerful paradigm for dynamic world modeling.
Yet the synthesis presented in Sections \ref{sec:dyn_world_model_representations}–\ref{sec:taxonomy} shows that this paradigm is still evolving. 
Several fundamental tensions remain unresolved: between geometric uncertainty and semantic expressiveness, between short-term agent tracking and lifelong changes detections, between representational richness and computational tractability, and between single-robot operation and distributed autonomy.
This section organizes open challenges around the three research questions presented in Section \ref{sec:introduction}, clarifying where the field has progressed and where substantial work remains.

\subsection{Representation: What to Model and How}\label{sec:open_challenges_representation}
\paragraph{Uncertainty Propagation}
This survey has identified the FG-SG uncertainty boundary as a recurring structural gap, with an illustrative example in Figure \ref{fig:FG_SG_uncertainty_boundary}.
Within the factor graph layer, uncertainty is principled: variables carry covariance (or information) matrices, factors are weighted according to measurement quality, and the Hessian of the non-linear least-squares problem propagates uncertainty globally through the graph \cite{dellaert_factor_2017}. 
In many hybrid systems, this stops at the interface between layers.
In layered-stack architectures, the most prevalent pattern, scene graph nodes are often instantiated from the geometric layer without inheriting pose uncertainty, such that a node extracted from a poorly constrained region of the map may be treated with the same confidence as one from a well-constrained area.
The unified optimizable graph pattern (e.g. S-Graphs \cite{bavle_s-graphs_2025}, DynEMo-SLAM \cite{giberna_dynemo-slam_2025}) partially mitigates this by representing semantic entities as factor graph variables. 
However, propagation of geometric uncertainty into \textit{relational} edges remains largely unaddressed (e.g. how certain is an \textit{on} relation given the covariance of both object pose nodes?).
The work of Millan-Romera et al. \cite{millanromera2026generationuncertaintyawarehighlevelspatial} is a step toward learned uncertainty estimates for semantic factors, but a general framework for uncertainty-aware scene-graph relations in dynamic environments does not yet exist.
Lost \& Found \cite{behrens_lost_2025} explicitly acknowledges that principled uncertainty quantification for tracked object states under interaction is a critical missing piece. 

A symmetric and largely unexamined problem exists at the opposite boundary too, between the perception front-ends and the factor graph itself.
The probabilistic machinery described in Section \ref{sec:background} rests on the assumption of zero-mean, temporally independent Gaussian measurement noise. 
As front-ends increasingly replace geometric primitives with learned geometric priors, this assumption is systematically violated: monocular depth foundation models, designed for single images, exhibit temporally inconsistent, flickering predictions across consecutive frames of a video \cite{video_depth_anything}, producing errors that are correlated over time and biased rather than zero-mean. 
Such errors are particularly damaging for dynamic entities, where the multi-view temporal fusion used to stabilize depth on static structure \cite{video_depth_anything} is unavailable, since independent object motion breaks the underlying temporal consistency constraints.
While learned depth and its uncertainty have been integrated into factor graph estimation for the static case \cite{bloesch2018codeslam} \cite{yang2020d3vo}, no equivalent framework calibrates the error statistics of learned geometric priors before their injection into dynamic-entity factors; the learned semantic covariances of \cite{millan-romera_learning_2024} address the semantic factor class only.

Closing these gaps, at both perception-factor graph and factor graph-scene graph boundaries, appears increasingly important for safety-critical deployment, where a robot must know not only \textit{what} its world model asserts, but also \textit{how much} to trust each assertion.

\paragraph{Non-Rigid Deformation Modeling}
The surveyed works are largely restricted to rigid-body dynamics: agents are modeled as rigid objects with 6-DoF pose trajectories, and structures are assumed piecewise rigid.
Non-rigid deformation, such as soft objects, cloth, liquids, deformable tools, remains almost entirely absent from probabilistic graph-based representations.
The body-model comparison in Section \ref{sec:dynamics_representations_agents} and Table \ref{tab:body_models} shows the spectrum: SMPL-family parametric models handle articulated human motion through skeletal kinematics, but treat the body surface as a predefined template with limited expressiveness for atypical morphologies or extreme deformation, while neural representation such as InstantAvatar \cite{jiang2022instantavatar} achieve higher fidelity at the cost of losing differentiability with respect to classical factor-graph optimization. 
Extending factor graphs to non-rigid entities requires defining suitable state spaces, motion priors, and measurement models, a challenge connected to the broader problem of deformable SLAM and still an open research frontier \cite{luca_carlone_slam_2025}.

\paragraph{Completeness of Dynamics Coverage}
As summarized by Table~\ref{tab:taxonomy_overview} (Section~\ref{sec:taxonomy}), no current system in the surveyed literature handles all three levels of dynamics (continuously moving agents, semi-static changes, and long-term structural evolution) simultaneously within a single unified probabilistic framework.
Systems that handle agents well (DynoSAM \cite{morris_dynosam_2025}, AirDOS \cite{qiu2022airdos}) typically do not model semi-static change; systems that handle semi-static change well (Khronos \cite{schmid_khronos_2024}, DynEMo-SLAM \cite{giberna_dynemo-slam_2025}) generally lack principled lifelong structural maintenance.
This gap reflects not only engineering constraints but also a deeper modeling difficulty: these dynamics occur at timescales that differ by order of magnitude, and the most effective probabilistic treatments are not trivially composable (e.g.  tightly-coupled factor-graph optimization for agent motion, persistence/decay mechanism for semi-static change, and inter session change detection with loop closure for structural evolution).
Unifying these regimes without sacrifying tractability remains an open problem.

\paragraph{Metric Scale Observability of Dynamic Entities under Minimal Sensing}
A structural pattern visible in Table \ref{tab:taxonomy_overview} but rarely made explicit in the literature is that the coverage of dynamics reported above is conditioned on metrically informative sensing: nearly all surveyed factor-graph formulations rely on stereo, RGB-D, or LiDAR input \cite{bescos_dynaslam_2021} \cite{morris_dynosam_2025} \cite{qiu2022airdos} \cite{gou_visual_2024}. 
This is not incidental. For an independently moving entity observed monocularly, the entity's structure scale, its motion, and the ego-motion scale are mutually entangled: unlike static structure, whose metric scale can be recovered from inertial measurements or transferred from the ego solution, the observed image displacements of a dynamic entity hinder ego-motion and object motion, making ratio-based scale transfer invalid.
The few monocular systems that do estimate dynamic entities resolve this only through strong metric priors: learned object dimensions \cite{yang_cubeslam_2019}, parametric human body models \cite{henning_bodyslam_2022} \cite{henning_bodyslam_2023}, or rigid-body kinematic structure \cite{qiu2022airdos}.
Existing analysis of dynamic SLAM formulations address coordinate frame choices \cite{morris_importance_2024} and estimation efficiency \cite{henein_dynamic_2020}, but to the best of our knowledge no principled observability analysis exists that characterizes which components of a dynamic entity's $SE(3)$ trajectory and scale are recoverable given a specific sensor suite, nor which shared static structure can serve as a metric anchor in the absence of direct depth. 
Closing this gap is a precondition for deploying the dynamic world models surveyed here on low-cost, weight-constrained platforms such as micro aerial vehicles and AR devices, where monocular-inertial sensing is often the only option.

\paragraph{Gaussian Splatting as a Geometric Substrate}
3D Gaussian Splatting (3DGS) \cite{kerbl_3d_2023}, a differentiable, high-fidelity alternative to point clouds and meshes, is an emerging geometric substrate not yet integrated with the probabilistic graph-based frameworks surveyed here.
Dynamic extensions \cite{matsuki_gaussian_2024} \cite{zheng_wildgs-slam_2025} \cite{li_dynagslam_2025}  \cite{xu_dg-slam_2024} separate static and dynamic Gaussian primitives, but largely follow an outlier-rejection paradigm rather than a probabilistic SLAMMOT formulation. 
A natuaral next step is a tripartite representation in which a factor graph provides the probabilistic trajectory backbone, a Gaussian-splatting layer the geometric-photometric substrate, and a scene graph the relational semantic abstraction, raising open questions on propagating factor-graph uncertainty into Gaussian primitives, extracting scene-graph entities from Gaussian fields, and jointly optimizing all three layers in real time.

\subsection{Construction and Maintenance: How to Build and Maintain}\label{sec:open_challenges_construction}

\paragraph{Scalable Lifelong Graph Maintenance}
Long-term autonomy requires world models that remain tractable and accurate across sessions separated by hours, days, or months.
DPG-SLAM \cite{walcott-bryant_dynamic_2012} established a core principle: a graph cannot grow indefinitely, and an explicit forgetting policy, such as pruning nodes when they no longer contribute topological information, is essential for lifelong operation.
LT-mapper \cite{kim2022lt} demonstrated this principle at large scale for outdoor LiDAR SLAM. 
However, as the dynamics and structural complexity of the environment increase, the decision between what to keep and what to discard becomes harder to resolve reliably.
Many existing forgetting policies are primarily geometric or heuristic, and do not explicitly reason about the downstream utility of information for future tasks.
A promising direction is task-informed graph maintenance \cite{maggio_clio_2024}, where the decision to retain, update or prune nodes is conditioned on anticipated task distributions, eventually linking memory management to the planning and interactions layers described in Sections \ref{sec:applications} and \ref{sec:querying_reasoning_action}.

\paragraph{Open-Vocabulary Association under Severe Dynamics}
The shift toward open-vocabulary perception (Section \ref{sec:taxonomy_trends}) is a significant advance, but current open-vocabulary systems are often evaluated under relatively controlled conditions.
Under severe dynamics, like dense crowd occlusion, rapid viewpoint change, and simultaneous motion of many agents, the vision-language embeddings used for association (e.g. CLIP features in \cite{yan_dynamic_2025}, \cite{gu_conceptgraphs_2023}) may become unstable, and the lack of closed-form data-association uncertainty increases vulnerability to catastrophic misassociations. 
Integrating open-vocabulary perception into principled multi-hypothesis association frameworks (Section \ref{sec:construction_pipeline_data_association}) within this domain remains open, as does the question of how to bound false-association rates in scenes with many visually similar entities in motion.

\paragraph{Multi-Robot Shared Dynamic World Models}
As shown in Table~\ref{tab:taxonomy_overview}, explicit multi-robot capability (MR) is rare across the surveyed works: only Kimera \cite{rosinol_kimera_2021} and CURB-SG \cite{greve_collaborative_2024} explicitly target it, and neither addresses the full complexity of shared dynamic world modeling in which multiple robots simultaneously observe, track, and update a common representation of moving agents and changing structures.
Multi S-Graphs \cite{fernandez-cortizas_multi_2024} demonstrates that semantic-relational information can be exchanged efficiently to improve map merging, but extending this to dynamic entity tracking, where an agent observed by robot \textit{A} must be correctly associated with the track maintained by robot \textit{B} despite occlusion, viewpoint and time differences, introduces cross-agent data-association challenges that remain largely unaddressed. 
PMBM-based formulations \cite{ge2025batch} offer a theoretically grounded starting point towards a distributed extension, but have not yet been coupled to scene-graph-level semantic reasoning and maintenance policies.

\subsection{Exploitation}
The applications surveyed in Section \ref{sec:applications} demonstrate that hybrid probabilistic graph representations can support a qualitatively richer set of downstream capabilities than either factor graphs or scene graphs alone. 
Yet the exploitation layer reveals its own set of open challenges, which are distinct from the representation and maintenance gaps above: even a perfectly maintained world model cannot be fully exploited by current task-level systems.

\paragraph{Uncertainty-Aware Downstream Tasks and Proactive Planning under Dynamics}
The probabilistic machinery underlying factor graphs produces rich uncertainty estimates, such as pose covariances and existence probabilities for dynamic entities, that are largely discarded at task execution time. 
In practice, planners exploit scene graph nodes as point estimates, manipulation systems act on deterministic object poses, and task reasoners treat scene state as binary truth, regardless of how well constrained the underlying map is. 
This disconnection is particularly problematic near the boundaries of map reliability, where sparse loop closures or recent dynamics induce high uncertainty but do not influence decision-making. 
Planning over scene graphs has begun to address graph updates \cite{ejaz2025situationallyawarepathplanningexploiting}, yet broader questions remain open: how map uncertainty should affect task selection, action sequencing, or the decision to actively seek additional observations. 
A closely related limitation appears in planning under dynamics. 
Current systems either replan reactively when scene graph nodes change or invalidate precomputed paths \cite{ejaz2025situationallyawarepathplanningexploiting}, while predictive approaches, such as Aion’s temporal flow statistics \cite{catalano_aion_2025} or trajectory forecasting from scene graph context \cite{gorlo_long-term_2024}, remain decoupled from probabilistic task planning. 
Bridging factor-graph uncertainty and predictive temporal models into task-level planning objectives, so that actions are conditioned on the expected future state of the world rather than its instantaneous estimate, remains a largely open challenge.

\paragraph{Closing the Exploitation-to-Representation Loop: Active Perception}
All surveyed systems are effectively open-loop between exploitation and representation: task execution consumes the world model but does not feed information back to guide its refinement. 
A robot that fails to grasp an object because its pose estimate was uncertain has acquired evidence about which part of the map is unreliable, but no current system uses task-execution outcomes to trigger targeted re-observation or prioritize re-mapping of specific regions. 
The active perception literature addresses related problems, but largely in isolation from the structured world models surveyed here. 
Coupling the task layer to the maintenance layer, so that downstream failures or high-uncertainty decisions trigger directed exploration to reduce map uncertainty in task-relevant regions, would close the loop between the three research questions. 
This direction is conceptually previewed by task-informed graph pruning in Clio \cite{maggio_clio_2024}, which uses task context to decide what to retain, but the inverse direction, using task failures to decide what to re-observe, remains entirely open.

\paragraph{Temporal Consistency of Language-Grounded Task Plans}
Systems like SayPlan \cite{rana2023sayplangroundinglargelanguage} and SG-Nav \cite{yin2024sgnav} ground natural language instructions in scene graph state to produce task plans. 
This grounding is computed at a single moment in time and assumed stable for the duration of task execution. 
In dynamic environments, however, the scene graph changes continuously: objects are displaced by human interaction, agents move, doors open and close. 
A task plan grounded in a scene graph that was valid at planning time may refer to entities or relations that no longer hold at execution time, leading to silent failures that are difficult to detect and recover from. 
No current system monitors the temporal consistency between an active task plan and the evolving world model, or triggers re-planning when the graph changes in ways that invalidate the current plan's preconditions. 
This is particularly acute for multi-step manipulation tasks in human-populated environments, where the interval between planning and execution may span multiple human interactions. 
The connection between dynamic graph update policies (Section~\ref{sec:construction_pipeline_graph_maintenance}) and plan validity monitoring is entirely unaddressed in the surveyed literature.

\paragraph{Evaluation Gap: Task Performance as a Function of World Model Quality}
A methodological gap cuts across all exploitation challenges: there is currently no benchmark or evaluation protocol that measures downstream task performance as a function of world model accuracy under dynamics. 
This limitation stems primarily from the lack of challenging datasets in which the tight integration of dynamic handling and reasoning is required for the system to succeed on them.
Most work relies on commonly used and accepted datasets (Section ~\ref{sec:evaluation_metrics_datasets}) which are not built to create a realistic, challenging scenario regarding the dynamicity level of the environment. 
This results on under-validated works or the requirement for them to generate self-collected datasets to showcase the effective benefit of the proposed contributions \cite{giberna_dynemo-slam_2025}.
This limitation is compounded by a more basic one: accurate 6-DoF ground truth for the dynamic entities themselves is essentially only available inside motion-capture volumes \cite{sturm_benchmark_2012}\cite{palazzolo2019iros}\cite{judd_oxford_2019}, restricting validation to room-scale scenes, or as human-annotated 3D bounding boxes in driving benchmarks\cite{geiger_are_2012}\cite{caesar_nuscenes_2019}, whose accuracy and domain do not transfer to indoor and mixed indoor-outdoor robotics. 
No existing dataset provides metrically accurate object trajectories for handheld or robot-mounted sequences at building scale, which is the regime targeted by most systems in Table \ref{tab:taxonomy_overview}, and a further reason why simulation\cite{rosinol_kimera_2021}\cite{tartanair2020iros} and self-collected data \cite{giberna_dynemo-slam_2025} dominate current evaluations.
Existing evaluation protocols (Section~\ref{sec:evaluation_metrics_protocols}) assess trajectory accuracy, map geometry, graph structure, and tracking metrics largely in isolation from task outcomes. 
The connection between, for instance, a 10\% increase in object pose RMSE under crowded dynamics and the resulting manipulation success rate degradation is not measured by any of the surveyed evaluation frameworks. 
This gap makes it difficult to quantify the practical benefit of more accurate dynamic world models for robotic autonomy, and to motivate the computational cost of the more expressive representations described in this survey. 
Developing task-conditioned evaluation protocols, where world model quality is assessed through its effect on downstream task success under varying levels of environmental dynamics, would provide a more grounded benchmark for future progress. 
This connects naturally to the consecutive task execution protocols already adopted by works like DovSG \cite{yan_dynamic_2025}, which could be extended to include controlled degradation of world model quality as an independent variable.

\bigskip
The open challenges above allow the three research questions presented in Section \ref{sec:introduction} to be summarized. 
Regarding \textit{which representations} are suited to dynamic environments, the taxonomy shows that no single graph stack dominates; instead, key limitations arise at the interfaces between paradigms (notably uncertainty transfer across the FG-SG boundary and incomplete coverage across dynamics timescales). 
Regarding \textit{how representations can be constructed and maintained}, the pipeline is increasingly well understood for single sessions with moderate dynamics, but scalable lifelong maintenance, robust open-vocabulary association under severe dynamics, and multi-robot shared construction remain unresolved. 
Regarding \textit{how representations can be exploited}, downstream capabilities are already broad, yet their reliability depends on representations that remain consistent under change and can quantify confidence across geometric and semantic layers. 
In this context, tighter integration between probabilistic factor graphs, semantically rich scene graphs, and emerging dense geometric substrates, in our view, one of the central challenges, and a promising direction, for long-term robotic situational awareness in dynamic environments.

\bibliographystyle{IEEEtran}
\bibliography{bibliography, references}

@article{tourani2026vs,
    title={vs-graphs: Tightly coupling visual slam and 3d scene graphs exploiting hierarchical scene understanding},
    author={Tourani, Ali and Ejaz, Saad and Bavle, Hriday and Fernandez-Cortizas, Miguel and Morilla-Cabello, David and Sanchez-Lopez, Jose Luis and Voos, Holger},
    journal={IEEE Robotics and Automation Letters},
    year={2026},
    publisher={IEEE}
}

@inproceedings{adaimi2023composite,
  title={Composite relationship fields with transformers for scene graph generation},
  author={Adaimi, George and Mizrahi, David and Alahi, Alexandre},
  booktitle={Proceedings of the IEEE/CVF Winter Conference on Applications of Computer Vision},
  pages={52--64},
  year={2023}
}

@article{li2024scene,
  title={Scene graph generation: A comprehensive survey},
  author={Li, Hongsheng and Zhu, Guangming and Zhang, Liang and Jiang, Youliang and Dang, Yixuan and Hou, Haoran and Shen, Peiyi and Zhao, Xia and Shah, Syed Afaq Ali and Bennamoun, Mohammed},
  journal={Neurocomputing},
  volume={566},
  pages={127052},
  year={2024},
  publisher={Elsevier}
}

@InProceedings{Johnson_2015_CVPR,
author = {Johnson, Justin and Krishna, Ranjay and Stark, Michael and Li, Li-Jia and Shamma, David and Bernstein, Michael and Fei-Fei, Li},
title = {Image Retrieval Using Scene Graphs},
booktitle = {Proceedings of the IEEE Conference on Computer Vision and Pattern Recognition (CVPR)},
month = {June},
year = {2015}
}

@article{mukadam2018continuous,
  title={Continuous-time Gaussian process motion planning via probabilistic inference},
  author={Mukadam, Mustafa and Dong, Jing and Yan, Xinyan and Dellaert, Frank and Boots, Byron},
  journal={The International Journal of Robotics Research},
  volume={37},
  number={11},
  pages={1319--1340},
  year={2018},
  publisher={SAGE Publications Sage UK: London, England}
}

@inproceedings{hahnel2003map,
  title={Map building with mobile robots in dynamic environments},
  author={Hahnel, Dirk and Triebel, Rudolph and Burgard, Wolfram and Thrun, Sebastian},
  booktitle={2003 IEEE International Conference on Robotics and Automation (Cat. No. 03CH37422)},
  volume={2},
  pages={1557--1563},
  year={2003},
  organization={IEEE}
}

@article{video_depth_anything,
  title={Video Depth Anything: Consistent Depth Estimation for Super-Long Videos},
  author={Chen, Sili and Guo, Hengkai and Zhu, Shengnan and Zhang, Feihu and Huang, Zilong and Feng, Jiashi and Kang, Bingyi},
  journal={arXiv:2501.12375},
  year={2025}
}

@inproceedings{bloesch2018codeslam,
  title={Codeslam—learning a compact, optimisable representation for dense visual slam},
  author={Bloesch, Michael and Czarnowski, Jan and Clark, Ronald and Leutenegger, Stefan and Davison, Andrew J},
  booktitle={Proceedings of the IEEE conference on computer vision and pattern recognition},
  pages={2560--2568},
  year={2018}
}

@inproceedings{yang2020d3vo,
  title={D3vo: Deep depth, deep pose and deep uncertainty for monocular visual odometry},
  author={Yang, Nan and Stumberg, Lukas von and Wang, Rui and Cremers, Daniel},
  booktitle={Proceedings of the IEEE/CVF conference on computer vision and pattern recognition},
  pages={1281--1292},
  year={2020}
}

@article{kucner_survey_2023,
author = {Tomasz Piotr Kucner and Martin Magnusson and Sariah Mghames and Luigi Palmieri and Francesco Verdoja and Chittaranjan Srinivas Swaminathan and Tomáš Krajník and Erik Schaffernicht and Nicola Bellotto and Marc Hanheide and Achim J Lilienthal},
title ={Survey of maps of dynamics for mobile robots},

journal = {The International Journal of Robotics Research},
volume = {42},
number = {11},
pages = {977-1006},
year = {2023},
doi = {10.1177/02783649231190428},

URL = { 
    
        https://doi.org/10.1177/02783649231190428
    
    

},
eprint = { 
    
        https://doi.org/10.1177/02783649231190428
    
    

}
}

@article{cheng2022sg,
  title={SG-SLAM: A real-time RGB-D visual SLAM toward dynamic scenes with semantic and geometric information},
  author={Cheng, Shuhong and Sun, Changhe and Zhang, Shijun and Zhang, Dianfan},
  journal={IEEE Transactions on Instrumentation and Measurement},
  volume={72},
  pages={1--12},
  year={2022},
  publisher={IEEE}
}

@article{wang2025prompthmr,
  title={PromptHMR: Promptable Human Mesh Recovery},
  author={Wang, Yufu and Sun, Yu and Patel, Priyanka and Daniilidis, Kostas and Black, Michael J and Kocabas, Muhammed},
  journal={Proceedings of the IEEE/CVF Conference on Computer Vision and Pattern Recognition},
  year={2025}
}

@article{xu_dg-slam_2024,
  title={Dg-slam: Robust dynamic gaussian splatting slam with hybrid pose optimization},
  author={Xu, Yueming and Jiang, Haochen and Xiao, Zhongyang and Feng, Jianfeng and Zhang, Li},
  journal={Advances in Neural Information Processing Systems},
  volume={37},
  pages={51577--51596},
  year={2024}
}

@inproceedings{li_dynagslam_2025,
  title={Dynagslam: Real-time gaussian-splatting slam for online rendering, tracking, motion predictions of moving objects in dynamic scenes},
  author={Li, Runfa Blark and Shaghaghi, Mahdi and Suzuki, Keito and Liu, Xinshuang and Moparthi, Varun and Du, Bang and Curtis, Walker and Renschler, Martin and Lee, Ki Myung Brian and Atanasov, Nikolay and others},
  booktitle={Proceedings of the IEEE/CVF Winter Conference on Applications of Computer Vision},
  pages={2434--2444},
  year={2026}
}

@inproceedings{zheng_wildgs-slam_2025,
  title={Wildgs-slam: Monocular gaussian splatting slam in dynamic environments},
  author={Zheng, Jianhao and Zhu, Zihan and Bieri, Valentin and Pollefeys, Marc and Peng, Songyou and Armeni, Iro},
  booktitle={Proceedings of the IEEE/CVF Conference on Computer Vision and Pattern Recognition},
  pages={11461--11471},
  year={2025}
}

@InProceedings{palazzolo2019iros,
author = {E. Palazzolo and J. Behley and P. Lottes and P. Gigu\`ere and C. Stachniss},
title = {{ReFusion: 3D Reconstruction in Dynamic Environments for RGB-D Cameras Exploiting Residuals}},
booktitle = iros,
year = {2019},
url = {https://www.ipb.uni-bonn.de/pdfs/palazzolo2019iros.pdf},
codeurl = {https://github.com/PRBonn/refusion},
videourl = {https://youtu.be/1P9ZfIS5-p4},
}

@misc{millanromera2026generationuncertaintyawarehighlevelspatial,
      title={Generation of Uncertainty-Aware High-Level Spatial Concepts in Factorized 3D Scene Graphs via Graph Neural Networks}, 
      author={Jose Andres Millan-Romera and Muhammad Shaheer and Miguel Fernandez-Cortizas and Martin R. Oswald and Holger Voos and Jose Luis Sanchez-Lopez},
      year={2026},
      eprint={2409.11972},
      archivePrefix={arXiv},
      primaryClass={cs.RO},
      url={https://arxiv.org/abs/2409.11972}, 
}

@inproceedings{kim2022lt,
  title={Lt-mapper: A modular framework for lidar-based lifelong mapping},
  author={Kim, Giseop and Kim, Ayoung},
  booktitle={2022 International Conference on Robotics and Automation (ICRA)},
  pages={7995--8002},
  year={2022},
  organization={IEEE}
}

@inproceedings{labbe2014online,
  title={Online global loop closure detection for large-scale multi-session graph-based SLAM},
  author={Labbe, Mathieu and Michaud, Fran{\c{c}}ois},
  booktitle={2014 IEEE/RSJ International Conference on Intelligent Robots and Systems},
  pages={2661--2666},
  year={2014},
  organization={IEEE}
}

@inproceedings{kaess2011isam2,
  title={iSAM2: Incremental smoothing and mapping with fluid relinearization and incremental variable reordering},
  author={Kaess, Michael and Johannsson, Hordur and Roberts, Richard and Ila, Viorela and Leonard, John and Dellaert, Frank},
  booktitle={2011 IEEE international conference on robotics and automation},
  pages={3281--3288},
  year={2011},
  organization={IEEE}
}

@ARTICLE{zhang_outlier_2023,
  author={Zhang, Jianbo and Yuan, Liang and Ran, Teng and Tao, Qing and Wu, Zhizhou},
  journal={IEEE Sensors Journal}, 
  title={Outlier Elimination for Monocular Object SLAM Based on Spatiotemporal Consistency Constraints}, 
  year={2023},
  volume={23},
  number={8},
  pages={8887-8898},
  doi={10.1109/JSEN.2023.3252050}}

@article{ristic2013tutorial,
  title={A tutorial on Bernoulli filters: theory, implementation and applications},
  author={Ristic, Branko and Vo, Ba-Tuong and Vo, Ba-Ngu and Farina, Alfonso},
  journal={IEEE Transactions on Signal Processing},
  volume={61},
  number={13},
  pages={3406--3430},
  year={2013},
  publisher={IEEE}
}

@article{garcia2018poisson,
  title={Poisson multi-Bernoulli mixture filter: Direct derivation and implementation},
  author={Garc{\'\i}a-Fern{\'a}ndez, {\'A}ngel F and Williams, Jason L and Granstr{\"o}m, Karl and Svensson, Lennart},
  journal={IEEE Transactions on Aerospace and Electronic Systems},
  volume={54},
  number={4},
  pages={1883--1901},
  year={2018},
  publisher={IEEE}
}

@article{reuter2014labeled,
  title={The labeled multi-Bernoulli filter},
  author={Reuter, Stephan and Vo, Ba-Tuong and Vo, Ba-Ngu and Dietmayer, Klaus},
  journal={IEEE Transactions on Signal Processing},
  volume={62},
  number={12},
  pages={3246--3260},
  year={2014},
  publisher={IEEE}
}

@article{li2023adaptive,
  title={An adaptive and scalable multi-object tracker based on the non-homogeneous Poisson process},
  author={Li, Qing and Gan, Runze and Liang, Jiaming and Godsill, Simon J},
  journal={IEEE Transactions on Signal Processing},
  volume={71},
  pages={105--120},
  year={2023},
  publisher={IEEE}
}

@article{mullane2011random,
  title={A random-finite-set approach to Bayesian SLAM},
  author={Mullane, John and Vo, Ba-Ngu and Adams, Martin D and Vo, Ba-Tuong},
  journal={IEEE transactions on robotics},
  volume={27},
  number={2},
  pages={268--282},
  year={2011},
  publisher={IEEE}
}

@article{kim2022pmbm,
  title={PMBM-based SLAM filters in 5G mmWave vehicular networks},
  author={Kim, Hyowon and Granstr{\"o}m, Karl and Svensson, Lennart and Kim, Sunwoo and Wymeersch, Henk},
  journal={IEEE Transactions on Vehicular Technology},
  volume={71},
  number={8},
  pages={8646--8661},
  year={2022},
  publisher={IEEE}
}

@article{ge2025batch,
  title={Batch SLAM with PMBM data association sampling and graph-based optimization},
  author={Ge, Yu and Kaltiokallio, Ossi and Xia, Yuxuan and Garc{\'\i}a-Fern{\'a}ndez, {\'A}ngel F and Kim, Hyowon and Talvitie, Jukka and Valkama, Mikko and Wymeersch, Henk and Svensson, Lennart},
  journal={IEEE Transactions on Signal Processing},
  year={2025},
  publisher={IEEE}
}

@incollection{siew2018simultaneous,
  title={Simultaneous localization and mapping with moving object tracking in 3d range data using probability hypothesis density (phd) filter},
  author={Siew, Peng Mun and Linares, Richard and Bageshwar, Vibhor},
  booktitle={2018 AIAA Information Systems-AIAA Infotech@ Aerospace},
  pages={0507},
  year={2018}
}

@inproceedings{chen2019multi,
  title={Multi-class target tracking using the semantic phd filter},
  author={Chen, Jun and Dames, Philip},
  booktitle={The International Symposium of Robotics Research},
  pages={526--541},
  year={2019},
  organization={Springer}
}

@article{deusch2015labeled,
  title={The labeled multi-Bernoulli SLAM filter},
  author={Deusch, Hendrik and Reuter, Stephan and Dietmayer, Klaus},
  journal={IEEE Signal Processing Letters},
  volume={22},
  number={10},
  pages={1561--1565},
  year={2015},
  publisher={IEEE}
}

@article{granstrom2019poisson,
  title={Poisson multi-Bernoulli mixture conjugate prior for multiple extended target filtering},
  author={Granstr{\"o}m, Karl and Fatemi, Maryam and Svensson, Lennart},
  journal={IEEE Transactions on Aerospace and Electronic Systems},
  volume={56},
  number={1},
  pages={208--225},
  year={2019},
  publisher={IEEE}
}

@inproceedings{SMPL-X:2019,
  title = {Expressive Body Capture: {3D} Hands, Face, and Body from a Single Image},
  author = {Pavlakos, Georgios and Choutas, Vasileios and Ghorbani, Nima and Bolkart, Timo and Osman, Ahmed A. A. and Tzionas, Dimitrios and Black, Michael J.},
  booktitle = {Proceedings IEEE Conf. on Computer Vision and Pattern Recognition (CVPR)},
  pages     = {10975--10985},
  year = {2019}
}

@inproceedings{STAR:2020,
      author = {Osman, Ahmed A A and Bolkart, Timo and Black, Michael J.},
      title = {{STAR}: A Sparse Trained Articulated Human Body Regressor},
      booktitle = {European Conference on Computer Vision (ECCV)},
      pages = {598--613},
      year = {2020},
      url = {https://star.is.tue.mpg.de}
}

@article{tartanair2020iros,
  title =   {TartanAir: A Dataset to Push the Limits of Visual SLAM},
  author =  {Wang, Wenshan and Zhu, Delong and Wang, Xiangwei and Hu, Yaoyu and Qiu, Yuheng and Wang, Chen and Hu, Yafei and Kapoor, Ashish and Scherer, Sebastian},
  booktitle = {2020 IEEE/RSJ International Conference on Intelligent Robots and Systems (IROS)},
  year =    {2020}
}

@inproceedings{kocabas2019vibe,
  title={VIBE: Video Inference for Human Body Pose and Shape Estimation},
  author={Kocabas, Muhammed and Athanasiou, Nikos and Black, Michael J.},
  booktitle = {The IEEE Conference on Computer Vision and Pattern Recognition (CVPR)},
  month = {June},
  year = {2020}
}

@InProceedings{rong2021frankmocap,
  title={FrankMocap: A Monocular 3D Whole-Body Pose Estimation System via Regression and Integration},
  author={Rong, Yu and Shiratori, Takaaki and Joo, Hanbyul},
  booktitle={IEEE International Conference on Computer Vision Workshops},
  year={2021}
}

@article{joo2020eft,
  title={Exemplar Fine-Tuning for 3D Human Pose Fitting Towards In-the-Wild 3D Human Pose Estimation},
  author={Joo, Hanbyul and Neverova, Natalia and Vedaldi, Andrea},
  journal={3DV},
  year={2021}
}

@inproceedings{lin2021end-to-end,
author = {Lin, Kevin and Wang, Lijuan and Liu, Zicheng},
title = {End-to-End Human Pose and Mesh Reconstruction with Transformers},
booktitle = {CVPR},
year = {2021},
}

@article{jiang2022instantavatar,
  author    = {Jiang, Tianjian and Chen, Xu and Song, Jie and Hilliges, Otmar},
  title     = {InstantAvatar: Learning Avatars from Monocular Video in 60 Seconds},
  booktitle = {Proceedings of the IEEE/CVF Conference on Computer Vision and Pattern Recognition (CVPR)},
  month     = {June},
  year      = {2023},
}

@misc{bregier2025humanmeshmodelinganny,
      title={Human Mesh Modeling for Anny Body}, 
      author={Romain Bregier and Guenole Fiche and Laura Bravo-Sanchez and Thomas Lucas and Matthieu Armando and Philippe Weinzaepfel and Gr\’egory Rogez and Fabien Baradel},
      year={2025},
      eprint={2511.03589},
      archivePrefix={arXiv},
      primaryClass={cs.CV},
      url={https://arxiv.org/abs/2511.03589}, 
}

@article{minodaRAL2021,
  title={{VIODE}: A Simulated Dataset to Address the Challenges of Visual-Inertial Odometry in Dynamic Environments},
  author={Minoda, Koji and Schilling, Fabian and W\"{u}est, Valentin and Floreano, Dario and Yairi, Takehisa},
  journal={IEEE Robotics and Automation Letters},
  year={2021},
  volume={6},
  number={2},
  pages={1343-1350},
  doi={10.1109/LRA.2021.3058073},
 }

@misc{viswanathan2025spadescalablepathplanning,
      title={SPADE: Towards Scalable Path Planning Architecture on Actionable Multi-Domain 3D Scene Graphs}, 
      author={Vignesh Kottayam Viswanathan and Akash Patel and Mario Alberto Valdes Saucedo and Sumeet Satpute and Christoforos Kanellakis and George Nikolakopoulos},
      year={2025},
      eprint={2505.19098},
      archivePrefix={arXiv},
      primaryClass={cs.RO},
      url={https://arxiv.org/abs/2505.19098}, 
}

@misc{ejaz2025situationallyawarepathplanningexploiting,
      title={Situationally-aware Path Planning Exploiting 3D Scene Graphs}, 
      author={Saad Ejaz and Marco Giberna and Muhammad Shaheer and Jose Andres Millan-Romera and Ali Tourani and Paul Kremer and Holger Voos and Jose Luis Sanchez-Lopez},
      year={2025},
      eprint={2508.06283},
      archivePrefix={arXiv},
      primaryClass={cs.RO},
      url={https://arxiv.org/abs/2508.06283}, 
}

@misc{viswanathan2024xflieleveragingactionablehierarchical,
      title={xFLIE: Leveraging Actionable Hierarchical Scene Representations for Autonomous Semantic-Aware Inspection Missions}, 
      author={Vignesh Kottayam Viswanathan and Mario A. V. Saucedo and Sumeet Gajanan Satpute and Christoforos Kanellakis and George Nikolakopoulos},
      year={2024},
      eprint={2412.19571},
      archivePrefix={arXiv},
      primaryClass={cs.RO},
      url={https://arxiv.org/abs/2412.19571}, 
}

@misc{ray2024taskmotionplanninghierarchical,
      title={Task and Motion Planning in Hierarchical 3D Scene Graphs}, 
      author={Aaron Ray and Christopher Bradley and Luca Carlone and Nicholas Roy},
      year={2024},
      eprint={2403.08094},
      archivePrefix={arXiv},
      primaryClass={cs.RO},
      url={https://arxiv.org/abs/2403.08094}, 
}

@misc{rana2023sayplangroundinglargelanguage,
      title={SayPlan: Grounding Large Language Models using 3D Scene Graphs for Scalable Robot Task Planning}, 
      author={Krishan Rana and Jesse Haviland and Sourav Garg and Jad Abou-Chakra and Ian Reid and Niko Suenderhauf},
      year={2023},
      eprint={2307.06135},
      archivePrefix={arXiv},
      primaryClass={cs.RO},
      url={https://arxiv.org/abs/2307.06135}, 
}

@article{yin2024sgnav, 
                  title={SG-Nav: Online 3D Scene Graph Prompting for LLM-based Zero-shot Object Navigation}, 
                  author={Hang Yin and Xiuwei Xu and Zhenyu Wu and Jie Zhou and Jiwen Lu},
                  journal={arXiv preprint arXiv:2410.08189},
                  year={2024}
            }

@INPROCEEDINGS{iros14-frog,
  author = {Rafael Ram\'{o}n-Vigo and Javier P\'{e}rez  and Fernando Caballero and Luis Merino},
  title = {Navigating among people in crowded environment: Datasets for localization and human robot interaction},
  booktitle = {Proceedings of the  Workshop on Robots in Clutter: Perception and Interaction in Clutter, IEEE/RSJ International Conference on Intelligent Robots and Systems (IROS)},
  year = {2014},
}

@inproceedings{qiu2022airdos,
  title={AirDOS: Dynamic SLAM benefits from articulated objects},
  author={Qiu, Yuheng and Wang, Chen and Wang, Wenshan and Henein, Mina and Scherer, Sebastian},
  booktitle={2022 International Conference on Robotics and Automation (ICRA)},
  year={2022},
  organization={IEEE}
}

@article{loper_smpl_2015,
	title = {{SMPL}: a skinned multi-person linear model},
	volume = {34},
	issn = {0730-0301, 1557-7368},
	shorttitle = {{SMPL}},
	url = {https://dl.acm.org/doi/10.1145/2816795.2818013},
	doi = {10.1145/2816795.2818013},
	language = {en},
	number = {6},
	urldate = {2026-04-01},
	journal = {ACM Transactions on Graphics},
	author = {Loper, Matthew and Mahmood, Naureen and Romero, Javier and Pons-Moll, Gerard and Black, Michael J.},
	month = nov,
	year = {2015},
	pages = {1--16},
}

@misc{henning_bodyslam_2022,
	title = {{BodySLAM}: {Joint} {Camera} {Localisation}, {Mapping}, and {Human} {Motion} {Tracking}},
	copyright = {arXiv.org perpetual, non-exclusive license},
	shorttitle = {{BodySLAM}},
	url = {https://arxiv.org/abs/2205.02301},
	doi = {10.48550/ARXIV.2205.02301},
	urldate = {2026-04-01},
	publisher = {arXiv},
	author = {Henning, Dorian F. and Laidlow, Tristan and Leutenegger, Stefan},
	year = {2022},
	note = {Version Number: 3},
}

@book{luca_carlone_slam_2025,
	title = {{SLAM} {Handbook}. {From} {Localization} and {Mapping} to {Spatial} {Intelligence}},
	publisher = {Cambridge University Press},
	author = {{Luca Carlone} and {Ayoung Kim} and {Timothy Barfoot} and {Daniel Cremers} and {Frank Dellaert}},
	year = {2025},
}

@article{wang_survey_2024,
	title = {A {Survey} of {Visual} {SLAM} in {Dynamic} {Environment}: {The} {Evolution} {From} {Geometric} to {Semantic} {Approaches}},
	volume = {73},
	copyright = {https://ieeexplore.ieee.org/Xplorehelp/downloads/license-information/IEEE.html},
	issn = {0018-9456, 1557-9662},
	shorttitle = {A {Survey} of {Visual} {SLAM} in {Dynamic} {Environment}},
	url = {https://ieeexplore.ieee.org/document/10577209/},
	doi = {10.1109/TIM.2024.3420374},
	urldate = {2026-03-31},
	journal = {IEEE Transactions on Instrumentation and Measurement},
	author = {Wang, Yanan and Tian, Yaobin and Chen, Jiawei and Xu, Kun and Ding, Xilun},
	year = {2024},
	pages = {1--21},
}

@article{wang_-depth_2025,
	title = {An {In}-{Depth} {Examination} of {SLAM} {Methods}: {Challenges}, {Advancements}, and {Applications} in {Complex} {Scenes} for {Autonomous} {Driving}},
	volume = {26},
	copyright = {https://ieeexplore.ieee.org/Xplorehelp/downloads/license-information/IEEE.html},
	issn = {1524-9050, 1558-0016},
	shorttitle = {An {In}-{Depth} {Examination} of {SLAM} {Methods}},
	url = {https://ieeexplore.ieee.org/document/10908366/},
	doi = {10.1109/TITS.2025.3545479},
	number = {7},
	urldate = {2026-03-31},
	journal = {IEEE Transactions on Intelligent Transportation Systems},
	author = {Wang, Ke and Guo, Juwei and Chen, Kai and Lu, Jianbo},
	month = jul,
	year = {2025},
	pages = {11066--11087},
}

@article{chen_semantic_2025,
	title = {Semantic {Visual} {Simultaneous} {Localization} and {Mapping}: {A} {Survey}},
	volume = {26},
	copyright = {https://ieeexplore.ieee.org/Xplorehelp/downloads/license-information/IEEE.html},
	issn = {1524-9050, 1558-0016},
	shorttitle = {Semantic {Visual} {Simultaneous} {Localization} and {Mapping}},
	url = {https://ieeexplore.ieee.org/document/11005698/},
	doi = {10.1109/TITS.2025.3556928},
	number = {6},
	urldate = {2026-03-31},
	journal = {IEEE Transactions on Intelligent Transportation Systems},
	author = {Chen, Kaiqi and Xiao, Junhao and Liu, Jialing and Tong, Qiyi and Zhang, Heng and Liu, Ruyu and Zhang, Jianhua and Ajoudani, Arash and Chen, Shengyong},
	month = jun,
	year = {2025},
	pages = {7426--7449},
}

@article{xu_survey_2021,
	title = {A survey: which features are required for dynamic visual simultaneous localization and mapping?},
	volume = {4},
	issn = {2524-4442},
	shorttitle = {A survey},
	url = {https://vciba.springeropen.com/articles/10.1186/s42492-021-00086-w},
	doi = {10.1186/s42492-021-00086-w},
	language = {en},
	number = {1},
	urldate = {2026-03-31},
	journal = {Visual Computing for Industry, Biomedicine, and Art},
	author = {Xu, Zewen and Rong, Zheng and Wu, Yihong},
	month = dec,
	year = {2021},
	pages = {20},
}

@article{dellaert_factor_2017,
	title = {Factor {Graphs} for {Robot} {Perception}},
	volume = {6},
	issn = {1935-8253, 1935-8261},
	url = {http://www.nowpublishers.com/article/Details/ROB-043},
	doi = {10.1561/2300000043},
	language = {en},
	number = {1-2},
	urldate = {2026-03-31},
	journal = {Foundations and Trends in Robotics},
	author = {Dellaert, Frank and Kaess, Michael},
	year = {2017},
	pages = {1--139},
}

@article{mascaro_scene_2025,
	title = {Scene {Representations} for {Robotic} {Spatial} {Perception}},
	volume = {8},
	copyright = {http://creativecommons.org/licenses/by/4.0/},
	issn = {2573-5144},
	url = {https://www.annualreviews.org/content/journals/10.1146/annurev-control-040423-030709},
	doi = {10.1146/annurev-control-040423-030709},
	language = {en},
	number = {1},
	urldate = {2026-03-31},
	journal = {Annual Review of Control, Robotics, and Autonomous Systems},
	author = {Mascaro, Ruben and Chli, Margarita},
	month = may,
	year = {2025},
	pages = {351--377},
}

@article{dellaert_factor_2021,
	title = {Factor {Graphs}: {Exploiting} {Structure} in {Robotics}},
	volume = {4},
	issn = {2573-5144, 2573-5144},
	shorttitle = {Factor {Graphs}},
	url = {https://www.annualreviews.org/doi/10.1146/annurev-control-061520-010504},
	doi = {10.1146/annurev-control-061520-010504},
	language = {en},
	number = {1},
	urldate = {2026-03-31},
	journal = {Annual Review of Control, Robotics, and Autonomous Systems},
	author = {Dellaert, Frank},
	month = may,
	year = {2021},
	pages = {141--166},
}

@article{bavle_slam_2023,
	title = {From {SLAM} to {Situational} {Awareness}: {Challenges} and {Survey}},
	volume = {23},
	issn = {1424-8220},
	shorttitle = {From {SLAM} to {Situational} {Awareness}},
	url = {https://www.mdpi.com/1424-8220/23/10/4849},
	doi = {10.3390/s23104849},
	language = {en},
	number = {10},
	urldate = {2026-03-31},
	journal = {Sensors},
	author = {Bavle, Hriday and Sanchez-Lopez, Jose Luis and Cimarelli, Claudio and Tourani, Ali and Voos, Holger},
	month = may,
	year = {2023},
	pages = {4849},
}

@misc{khan_foundation_2025,
	title = {Foundation {Model} {Driven} {Robotics}: {A} {Comprehensive} {Review}},
	copyright = {Creative Commons Attribution 4.0 International},
	shorttitle = {Foundation {Model} {Driven} {Robotics}},
	url = {https://arxiv.org/abs/2507.10087},
	doi = {10.48550/ARXIV.2507.10087},
	urldate = {2026-03-31},
	publisher = {arXiv},
	author = {Khan, Muhammad Tayyab and Waheed, Ammar},
	year = {2025},
	note = {Version Number: 1},
}

@article{lai_review_2022,
	title = {A {Review} on {Visual}-{SLAM}: {Advancements} from {Geometric} {Modelling} to {Learning}-{Based} {Semantic} {Scene} {Understanding} {Using} {Multi}-{Modal} {Sensor} {Fusion}},
	volume = {22},
	issn = {1424-8220},
	shorttitle = {A {Review} on {Visual}-{SLAM}},
	url = {https://www.mdpi.com/1424-8220/22/19/7265},
	doi = {10.3390/s22197265},
	language = {en},
	number = {19},
	urldate = {2026-03-31},
	journal = {Sensors},
	author = {Lai, Tin},
	month = sep,
	year = {2022},
	pages = {7265},
}

@article{amuche_advancements_2025,
	title = {Advancements, {Challenges}, and {Future} {Directions} in {Scene}-{Graph}-{Based} {Image} {Generation}: {A} {Comprehensive} {Review}},
	volume = {14},
	issn = {2079-9292},
	shorttitle = {Advancements, {Challenges}, and {Future} {Directions} in {Scene}-{Graph}-{Based} {Image} {Generation}},
	url = {https://www.mdpi.com/2079-9292/14/6/1158},
	doi = {10.3390/electronics14061158},
	language = {en},
	number = {6},
	urldate = {2026-03-31},
	journal = {Electronics},
	author = {Amuche, Chikwendu Ijeoma and Zhang, Xiaoling and Monday, Happy Nkanta and Nneji, Grace Ugochi and Ukwuoma, Chiagoziem C. and Chikwendu, Okechukwu Chinedum and Hyeon Gu, Yeong and Al-antari, Mugahed A.},
	month = mar,
	year = {2025},
	pages = {1158},
}

@article{chang_comprehensive_2023,
	title = {A {Comprehensive} {Survey} of {Scene} {Graphs}: {Generation} and {Application}},
	volume = {45},
	copyright = {https://ieeexplore.ieee.org/Xplorehelp/downloads/license-information/IEEE.html},
	issn = {0162-8828, 2160-9292, 1939-3539},
	shorttitle = {A {Comprehensive} {Survey} of {Scene} {Graphs}},
	url = {https://ieeexplore.ieee.org/document/9661322/},
	doi = {10.1109/TPAMI.2021.3137605},
	number = {1},
	urldate = {2026-03-31},
	journal = {IEEE Transactions on Pattern Analysis and Machine Intelligence},
	author = {Chang, Xiaojun and Ren, Pengzhen and Xu, Pengfei and Li, Zhihui and Chen, Xiaojiang and Hauptmann, Alex},
	month = jan,
	year = {2023},
	pages = {1--26},
}

@misc{giberna_dynemo-slam_2025,
	title = {{DYNEMO}-{SLAM}: {Dynamic} {Entity} and {Motion}-{Aware} {3D} {Scene} {Graph} {SLAM}},
	shorttitle = {{DYNEMO}-{SLAM}},
	url = {http://arxiv.org/abs/2503.02050},
	doi = {10.48550/arXiv.2503.02050},
	urldate = {2026-01-30},
	publisher = {arXiv},
	author = {Giberna, Marco and Shaheer, Muhammad and Fernandez-Cortizas, Miguel and Millan-Romera, Jose Andres and Sanchez-Lopez, Jose Luis and Voos, Holger},
	month = dec,
	year = {2025},
	note = {arXiv:2503.02050 [cs]},
}

@misc{henning_bodyslam_2023,
	title = {{BodySLAM}++: {Fast} and {Tightly}-{Coupled} {Visual}-{Inertial} {Camera} and {Human} {Motion} {Tracking}},
	copyright = {arXiv.org perpetual, non-exclusive license},
	shorttitle = {{BodySLAM}++},
	url = {https://arxiv.org/abs/2309.01236},
	doi = {10.48550/ARXIV.2309.01236},
	urldate = {2026-01-29},
	publisher = {arXiv},
	author = {Henning, Dorian F. and Choi, Christopher and Schaefer, Simon and Leutenegger, Stefan},
	year = {2023},
	note = {Version Number: 1},
}

@article{guivant_simultaneous_2002,
	title = {Simultaneous localization and map building using natural features and absolute information},
	volume = {40},
	copyright = {https://www.elsevier.com/tdm/userlicense/1.0/},
	issn = {09218890},
	url = {https://linkinghub.elsevier.com/retrieve/pii/S0921889002002336},
	doi = {10.1016/S0921-8890(02)00233-6},
	language = {en},
	number = {2-3},
	urldate = {2026-01-13},
	journal = {Robotics and Autonomous Systems},
	author = {Guivant, José E. and Masson, Favio R. and Nebot, Eduardo M.},
	month = aug,
	year = {2002},
	pages = {79--90},
}

@misc{dosovitskiy_carla_2017,
	title = {{CARLA}: {An} {Open} {Urban} {Driving} {Simulator}},
	copyright = {arXiv.org perpetual, non-exclusive license},
	shorttitle = {{CARLA}},
	url = {https://arxiv.org/abs/1711.03938},
	doi = {10.48550/ARXIV.1711.03938},
	urldate = {2026-01-13},
	publisher = {arXiv},
	author = {Dosovitskiy, Alexey and Ros, German and Codevilla, Felipe and Lopez, Antonio and Koltun, Vladlen},
	year = {2017},
	note = {Version Number: 1},
}

@misc{puig_virtualhome_2018,
	title = {{VirtualHome}: {Simulating} {Household} {Activities} via {Programs}},
	copyright = {arXiv.org perpetual, non-exclusive license},
	shorttitle = {{VirtualHome}},
	url = {https://arxiv.org/abs/1806.07011},
	doi = {10.48550/ARXIV.1806.07011},
	urldate = {2026-01-13},
	publisher = {arXiv},
	author = {Puig, Xavier and Ra, Kevin and Boben, Marko and Li, Jiaman and Wang, Tingwu and Fidler, Sanja and Torralba, Antonio},
	year = {2018},
	note = {Version Number: 1},
}

@misc{ravichandran_bridging_2020,
	title = {Bridging {Scene} {Understanding} and {Task} {Execution} with {Flexible} {Simulation} {Environments}},
	copyright = {arXiv.org perpetual, non-exclusive license},
	url = {https://arxiv.org/abs/2011.10452},
	doi = {10.48550/ARXIV.2011.10452},
	urldate = {2026-01-13},
	publisher = {arXiv},
	author = {Ravichandran, Zachary and Griffith, J. Daniel and Smith, Benjamin and Frost, Costas},
	year = {2020},
	note = {Version Number: 1},
}

@article{judd_oxford_2019,
	title = {The {Oxford} {Multimotion} {Dataset}: {Multiple} {SE}(3) {Motions} {With} {Ground} {Truth}},
	volume = {4},
	copyright = {https://ieeexplore.ieee.org/Xplorehelp/downloads/license-information/IEEE.html},
	issn = {2377-3766, 2377-3774},
	shorttitle = {The {Oxford} {Multimotion} {Dataset}},
	url = {https://ieeexplore.ieee.org/document/8610209/},
	doi = {10.1109/LRA.2019.2892656},
	number = {2},
	urldate = {2026-01-13},
	journal = {IEEE Robotics and Automation Letters},
	author = {Judd, Kevin Michael and Gammell, Jonathan D.},
	month = apr,
	year = {2019},
	pages = {800--807},
}

@inproceedings{sturm_benchmark_2012,
	address = {Vilamoura-Algarve, Portugal},
	title = {A benchmark for the evaluation of {RGB}-{D} {SLAM} systems},
	isbn = {978-1-4673-1736-8 978-1-4673-1737-5 978-1-4673-1735-1},
	url = {http://ieeexplore.ieee.org/document/6385773/},
	doi = {10.1109/IROS.2012.6385773},
	urldate = {2026-01-13},
	booktitle = {2012 {IEEE}/{RSJ} {International} {Conference} on {Intelligent} {Robots} and {Systems}},
	publisher = {IEEE},
	author = {Sturm, Jrgen and Engelhard, Nikolas and Endres, Felix and Burgard, Wolfram and Cremers, Daniel},
	month = oct,
	year = {2012},
	pages = {573--580},
}

@misc{dai_scannet_2017,
	title = {{ScanNet}: {Richly}-annotated {3D} {Reconstructions} of {Indoor} {Scenes}},
	copyright = {arXiv.org perpetual, non-exclusive license},
	shorttitle = {{ScanNet}},
	url = {https://arxiv.org/abs/1702.04405},
	doi = {10.48550/ARXIV.1702.04405},
	urldate = {2026-01-13},
	publisher = {arXiv},
	author = {Dai, Angela and Chang, Angel X. and Savva, Manolis and Halber, Maciej and Funkhouser, Thomas and Nießner, Matthias},
	year = {2017},
	note = {Version Number: 2},
}

@misc{shi_are_2019,
	title = {Are {We} {Ready} for {Service} {Robots}? {The} {OpenLORIS}-{Scene} {Datasets} for {Lifelong} {SLAM}},
	copyright = {arXiv.org perpetual, non-exclusive license},
	shorttitle = {Are {We} {Ready} for {Service} {Robots}?},
	url = {https://arxiv.org/abs/1911.05603},
	doi = {10.48550/ARXIV.1911.05603},
	urldate = {2026-01-13},
	publisher = {arXiv},
	author = {Shi, Xuesong and Li, Dongjiang and Zhao, Pengpeng and Tian, Qinbin and Tian, Yuxin and Long, Qiwei and Zhu, Chunhao and Song, Jingwei and Qiao, Fei and Song, Le and Guo, Yangquan and Wang, Zhigang and Zhang, Yimin and Qin, Baoxing and Yang, Wei and Wang, Fangshi and Chan, Rosa H. M. and She, Qi},
	year = {2019},
	note = {Version Number: 2},
}

@article{geiger_vision_2013,
	title = {Vision meets robotics: {The} {KITTI} dataset},
	volume = {32},
	issn = {0278-3649, 1741-3176},
	shorttitle = {Vision meets robotics},
	url = {https://journals.sagepub.com/doi/10.1177/0278364913491297},
	doi = {10.1177/0278364913491297},
	language = {en},
	number = {11},
	urldate = {2026-01-13},
	journal = {The International Journal of Robotics Research},
	author = {Geiger, A and Lenz, P and Stiller, C and Urtasun, R},
	month = sep,
	year = {2013},
	pages = {1231--1237},
}

@inproceedings{geiger_are_2012,
	address = {Providence, RI},
	title = {Are we ready for autonomous driving? {The} {KITTI} vision benchmark suite},
	isbn = {978-1-4673-1228-8 978-1-4673-1226-4 978-1-4673-1227-1},
	shorttitle = {Are we ready for autonomous driving?},
	url = {http://ieeexplore.ieee.org/document/6248074/},
	doi = {10.1109/CVPR.2012.6248074},
	urldate = {2026-01-13},
	booktitle = {2012 {IEEE} {Conference} on {Computer} {Vision} and {Pattern} {Recognition}},
	publisher = {IEEE},
	author = {Geiger, A. and Lenz, P. and Urtasun, R.},
	month = jun,
	year = {2012},
	pages = {3354--3361},
}

@misc{caesar_nuscenes_2019,
	title = {{nuScenes}: {A} multimodal dataset for autonomous driving},
	copyright = {arXiv.org perpetual, non-exclusive license},
	shorttitle = {{nuScenes}},
	url = {https://arxiv.org/abs/1903.11027},
	doi = {10.48550/ARXIV.1903.11027},
	urldate = {2026-01-13},
	publisher = {arXiv},
	author = {Caesar, Holger and Bankiti, Varun and Lang, Alex H. and Vora, Sourabh and Liong, Venice Erin and Xu, Qiang and Krishnan, Anush and Pan, Yu and Baldan, Giancarlo and Beijbom, Oscar},
	year = {2019},
	note = {Version Number: 5},
}

@misc{ferraina_lost-3dsg_2026,
	title = {{LOST}-{3DSG}: {Lightweight} {Open}-{Vocabulary} {3D} {Scene} {Graphs} with {Semantic} {Tracking} in {Dynamic} {Environments}},
	copyright = {Creative Commons Attribution 4.0 International},
	shorttitle = {{LOST}-{3DSG}},
	url = {https://arxiv.org/abs/2601.02905},
	doi = {10.48550/ARXIV.2601.02905},
	urldate = {2026-01-12},
	publisher = {arXiv},
	author = {Ferraina, Sara Micol and Brienza, Michele and Argenziano, Francesco and Musumeci, Emanuele and Suriani, Vincenzo and Bloisi, Domenico D. and Nardi, Daniele},
	year = {2026},
	note = {Version Number: 1},
}

@article{ding_dynamic_2025,
	title = {Dynamic {SLAM} by {Combining} {Rigid} {Feature} {Point} {Set} {Modeling} and {YOLO}},
	volume = {26},
	issn = {1424-8220},
	url = {https://www.mdpi.com/1424-8220/26/1/235},
	doi = {10.3390/s26010235},
	language = {en},
	number = {1},
	urldate = {2026-01-02},
	journal = {Sensors},
	author = {Ding, Pengchao and Wang, Weidong and Wu, Xian and Xu, Kangle and Wu, Dongmei and Du, Zhijiang},
	month = dec,
	year = {2025},
	pages = {235},
}

@article{maggio_clio_2024,
	title = {\textit{{Clio}:} {Real}-{Time} {Task}-{Driven} {Open}-{Set} {3D} {Scene} {Graphs}},
	volume = {9},
	copyright = {https://ieeexplore.ieee.org/Xplorehelp/downloads/license-information/IEEE.html},
	issn = {2377-3766, 2377-3774},
	shorttitle = {{\textless}i{\textgreater}{Clio}},
	url = {https://ieeexplore.ieee.org/document/10659066/},
	doi = {10.1109/LRA.2024.3451395},
	number = {10},
	urldate = {2026-01-02},
	journal = {IEEE Robotics and Automation Letters},
	author = {Maggio, Dominic and Chang, Yun and Hughes, Nathan and Trang, Matthew and Griffith, Dan and Dougherty, Carlyn and Cristofalo, Eric and Schmid, Lukas and Carlone, Luca},
	month = oct,
	year = {2024},
	pages = {8921--8928},
}

@misc{rotondi_fungraph_2025,
	title = {{FunGraph}: {Functionality} {Aware} {3D} {Scene} {Graphs} for {Language}-{Prompted} {Scene} {Interaction}},
	shorttitle = {{FunGraph}},
	url = {http://arxiv.org/abs/2503.07909},
	doi = {10.48550/arXiv.2503.07909},
	urldate = {2026-01-02},
	publisher = {arXiv},
	author = {Rotondi, Dennis and Scaparro, Fabio and Blum, Hermann and Arras, Kai O.},
	month = aug,
	year = {2025},
	note = {arXiv:2503.07909 [cs]},
}

@misc{zhang_open-vocabulary_2025,
	title = {Open-{Vocabulary} {Functional} {3D} {Scene} {Graphs} for {Real}-{World} {Indoor} {Spaces}},
	copyright = {Creative Commons Attribution 4.0 International},
	url = {https://arxiv.org/abs/2503.19199},
	doi = {10.48550/ARXIV.2503.19199},
	urldate = {2026-01-02},
	publisher = {arXiv},
	author = {Zhang, Chenyangguang and Delitzas, Alexandros and Wang, Fangjinhua and Zhang, Ruida and Ji, Xiangyang and Pollefeys, Marc and Engelmann, Francis},
	year = {2025},
	note = {Version Number: 1},
}

@misc{shao_large_2025,
	title = {Large {VLM}-based {Vision}-{Language}-{Action} {Models} for {Robotic} {Manipulation}: {A} {Survey}},
	copyright = {arXiv.org perpetual, non-exclusive license},
	shorttitle = {Large {VLM}-based {Vision}-{Language}-{Action} {Models} for {Robotic} {Manipulation}},
	url = {https://arxiv.org/abs/2508.13073},
	doi = {10.48550/ARXIV.2508.13073},
	urldate = {2026-01-02},
	publisher = {arXiv},
	author = {Shao, Rui and Li, Wei and Zhang, Lingsen and Zhang, Renshan and Liu, Zhiyang and Chen, Ran and Nie, Liqiang},
	year = {2025},
	note = {Version Number: 2},
}

@misc{gu_conceptgraphs_2023,
	title = {{ConceptGraphs}: {Open}-{Vocabulary} {3D} {Scene} {Graphs} for {Perception} and {Planning}},
	copyright = {arXiv.org perpetual, non-exclusive license},
	shorttitle = {{ConceptGraphs}},
	url = {https://arxiv.org/abs/2309.16650},
	doi = {10.48550/ARXIV.2309.16650},
	urldate = {2025-12-30},
	publisher = {arXiv},
	author = {Gu, Qiao and Kuwajerwala, Alihusein and Morin, Sacha and Jatavallabhula, Krishna Murthy and Sen, Bipasha and Agarwal, Aditya and Rivera, Corban and Paul, William and Ellis, Kirsty and Chellappa, Rama and Gan, Chuang and de Melo, Celso Miguel and Tenenbaum, Joshua B. and Torralba, Antonio and Shkurti, Florian and Paull, Liam},
	year = {2023},
	note = {Version Number: 1},
}

@misc{gorlo_describe_2025,
	title = {Describe {Anything} {Anywhere} {At} {Any} {Moment}},
	copyright = {arXiv.org perpetual, non-exclusive license},
	url = {https://arxiv.org/abs/2512.00565},
	doi = {10.48550/ARXIV.2512.00565},
	urldate = {2025-12-30},
	publisher = {arXiv},
	author = {Gorlo, Nicolas and Schmid, Lukas and Carlone, Luca},
	year = {2025},
	note = {Version Number: 1},
}

@article{zhang_semantic-integrated_2026,
	title = {Semantic-integrated multi-model fitting for real-time {VSLAM} in highly dynamic environments},
	volume = {42},
	issn = {0178-2789, 1432-2315},
	url = {https://link.springer.com/10.1007/s00371-025-04235-7},
	doi = {10.1007/s00371-025-04235-7},
	language = {en},
	number = {1},
	urldate = {2025-12-23},
	journal = {The Visual Computer},
	author = {Zhang, Tiantian and Li, Ni and Gong, Guanghong and Lin, Xin},
	month = jan,
	year = {2026},
	pages = {56},
}

@misc{catalano_aion_2025,
	title = {Aion: {Towards} {Hierarchical} {4D} {Scene} {Graphs} with {Temporal} {Flow} {Dynamics}},
	copyright = {Creative Commons Attribution Non Commercial No Derivatives 4.0 International},
	shorttitle = {Aion},
	url = {https://arxiv.org/abs/2512.11903},
	doi = {10.48550/ARXIV.2512.11903},
	urldate = {2025-12-19},
	publisher = {arXiv},
	author = {Catalano, Iacopo and Montijano, Eduardo and Civera, Javier and Placed, Julio A. and Pena-Queralta, Jorge},
	year = {2025},
	note = {Version Number: 1},
}

@article{morris_dynosam_2025,
	title = {{DynoSAM}: {Open}-{Source} {Smoothing} and {Mapping} {Framework} for {Dynamic} {SLAM}},
	copyright = {https://ieeexplore.ieee.org/Xplorehelp/downloads/license-information/IEEE.html},
	issn = {1552-3098, 1941-0468},
	shorttitle = {{DynoSAM}},
	url = {https://ieeexplore.ieee.org/document/11288097/},
	doi = {10.1109/TRO.2025.3641813},
	urldate = {2025-12-18},
	journal = {IEEE Transactions on Robotics},
	author = {Morris, Jesse and Wang, Yiduo and Kliniewski, Mikolaj and Ila, Viorela},
	year = {2025},
	pages = {1--19},
}

@article{tian_dl-slot_2024,
	title = {{DL}-{SLOT}: {Tightly}-{Coupled} {Dynamic} {LiDAR} {SLAM} and {3D} {Object} {Tracking} {Based} on {Collaborative} {Graph} {Optimization}},
	volume = {9},
	copyright = {https://ieeexplore.ieee.org/Xplorehelp/downloads/license-information/IEEE.html},
	issn = {2379-8904, 2379-8858},
	shorttitle = {{DL}-{SLOT}},
	url = {https://ieeexplore.ieee.org/document/10255288/},
	doi = {10.1109/TIV.2023.3317308},
	number = {1},
	urldate = {2025-11-21},
	journal = {IEEE Transactions on Intelligent Vehicles},
	author = {Tian, Xuebo and Zhu, Zhongyang and Zhao, Junqiao and Tian, Gengxuan and Ye, Chen},
	month = jan,
	year = {2024},
	pages = {1017--1027},
}

@article{song_rts-slam_2025,
	title = {{RTS}-{SLAM}: {Robust}, {Tightly}-{Coupled}, and {Semantic}-{Enhanced} {LiDAR}-{Inertial} {SLAM}},
	copyright = {https://ieeexplore.ieee.org/Xplorehelp/downloads/license-information/IEEE.html},
	issn = {1530-437X, 1558-1748, 2379-9153},
	shorttitle = {{RTS}-{SLAM}},
	url = {https://ieeexplore.ieee.org/document/11174098/},
	doi = {10.1109/JSEN.2025.3609911},
	urldate = {2025-09-25},
	journal = {IEEE Sensors Journal},
	author = {Song, Shuang and Zhu, Qingyuan and Hu, Huosheng},
	year = {2025},
	pages = {1--1},
}

@misc{morris_online_2025,
	title = {Online {Dynamic} {SLAM} with {Incremental} {Smoothing} and {Mapping}},
	copyright = {Creative Commons Attribution 4.0 International},
	url = {https://arxiv.org/abs/2509.08197},
	doi = {10.48550/ARXIV.2509.08197},
	urldate = {2025-09-17},
	publisher = {arXiv},
	author = {Morris, Jesse and Wang, Yiduo and Ila, Viorela},
	year = {2025},
	note = {Version Number: 1},
}

@misc{catalano_3d_2025,
	title = {{3D} {Scene} {Graphs} in {Robotics}: {A} {Unified} {Representation} {Bridging} {Geometry}, {Semantics}, and {Action}},
	copyright = {https://creativecommons.org/licenses/by-nc-sa/4.0/},
	shorttitle = {{3D} {Scene} {Graphs} in {Robotics}},
	url = {https://www.techrxiv.org/doi/full/10.36227/techrxiv.175561766.62405781},
	doi = {10.36227/techrxiv.175561766.62405781/v1},
	urldate = {2025-09-01},
	author = {Catalano, Iacopo and Zumaya, Carlos Cueto and Placed, Julio A and Civera, Javier and Bessa, Wallace Moreira and Peña-Queralta, Jorge},
	month = aug,
	year = {2025},
}

@inproceedings{maguire_lidar-only_2025,
	address = {Wollongong, Australia},
	title = {{LiDAR}-{Only} {Navigation} of {UGVs} in {Dynamic} {Environments} via {Graph} {Attention} {Networks} and {Deep} {Reinforcement} {Learning}},
	copyright = {https://doi.org/10.15223/policy-029},
	isbn = {9798331533892},
	url = {https://ieeexplore.ieee.org/document/10934926/},
	doi = {10.1109/ICM62621.2025.10934926},
	urldate = {2025-07-29},
	booktitle = {2025 {IEEE} {International} {Conference} on {Mechatronics} ({ICM})},
	publisher = {IEEE},
	author = {Maguire, Luke and Van, Mien and Olayemi, Kabirat and McLoone, Sean},
	month = feb,
	year = {2025},
	pages = {1--6},
}

@inproceedings{gulati_robust_2017,
	address = {Xi'an, China},
	title = {Robust cooperative localization in a dynamic environment using factor graphs and probability data association filter},
	isbn = {978-0-9964527-0-0},
	url = {http://ieeexplore.ieee.org/document/8009654/},
	doi = {10.23919/ICIF.2017.8009654},
	urldate = {2025-07-29},
	booktitle = {2017 20th {International} {Conference} on {Information} {Fusion} ({Fusion})},
	publisher = {IEEE},
	author = {Gulati, Dhiraj and Zhang, Feihu and Malovetz, Daniel and Clarke, Daniel and Knoll, Alois},
	month = jul,
	year = {2017},
	pages = {1--6},
}

@incollection{scherer_ogm2pgbm_2023,
	address = {London},
	edition = {1},
	title = {{OGM2PGBM}: {Robust} {BIM}-based {2D}-{LiDAR} localization for lifelong indoor navigation},
	isbn = {978-1-003-35422-2},
	shorttitle = {{OGM2PGBM}},
	url = {https://www.taylorfrancis.com/books/9781003354222/chapters/10.1201/9781003354222-72},
	doi = {10.1201/9781003354222-72},
	language = {en},
	urldate = {2025-07-29},
	booktitle = {{ECPPM} 2022 - {eWork} and {eBusiness} in {Architecture}, {Engineering} and {Construction} 2022},
	publisher = {CRC Press},
	author = {Vega Torres, M.A. and Braun, A. and Borrmann, A.},
	collaborator = {Scherer, Raimar J and Sujan, Sujesh F. and Hjelseth, Eilif},
	month = mar,
	year = {2023},
	pages = {567--574},
}

@article{tian_dynaquadric_2024,
	title = {{DynaQuadric}: {Dynamic} {Quadric} {SLAM} for {Quadric} {Initialization}, {Mapping}, and {Tracking}},
	volume = {25},
	copyright = {https://ieeexplore.ieee.org/Xplorehelp/downloads/license-information/IEEE.html},
	issn = {1524-9050, 1558-0016},
	shorttitle = {{DynaQuadric}},
	url = {https://ieeexplore.ieee.org/document/10591477/},
	doi = {10.1109/tits.2024.3418949},
	number = {11},
	urldate = {2025-07-10},
	journal = {IEEE Transactions on Intelligent Transportation Systems},
	publisher = {Institute of Electrical and Electronics Engineers (IEEE)},
	author = {Tian, Rui and Zhang, Yunzhou and Yang, Linghao and Zhang, Jinpeng and Coleman, Sonya and Kerr, Dermot},
	month = nov,
	year = {2024},
	pages = {17234--17246},
}

@inproceedings{morris_importance_2024,
	address = {Yokohama, Japan},
	title = {The {Importance} of {Coordinate} {Frames} in {Dynamic} {SLAM}},
	copyright = {https://doi.org/10.15223/policy-029},
	url = {https://ieeexplore.ieee.org/document/10610840/},
	doi = {10.1109/icra57147.2024.10610840},
	urldate = {2025-07-09},
	booktitle = {2024 {IEEE} {International} {Conference} on {Robotics} and {Automation} ({ICRA})},
	publisher = {IEEE},
	author = {Morris, Jesse and Wang, Yiduo and Ila, Viorela},
	month = may,
	year = {2024},
	pages = {13755--13761},
}

@article{gou_visual_2024,
	title = {A {Visual} {SLAM} {With} {Tightly} {Coupled} {Integration} of {Multiobject} {Tracking} for {Production} {Workshop}},
	volume = {11},
	copyright = {https://ieeexplore.ieee.org/Xplorehelp/downloads/license-information/IEEE.html},
	issn = {2327-4662, 2372-2541},
	url = {https://ieeexplore.ieee.org/document/10443596/},
	doi = {10.1109/jiot.2024.3368417},
	language = {en},
	number = {11},
	urldate = {2025-07-07},
	journal = {IEEE Internet of Things Journal},
	publisher = {Institute of Electrical and Electronics Engineers (IEEE)},
	author = {Gou, Rongsong and Chen, Guangzhu and Pu, Xin and Liao, Xiaojuan and Chen, Runji},
	month = jun,
	year = {2024},
	pages = {19949--19962},
}

@inproceedings{greve_collaborative_2024,
	address = {Yokohama, Japan},
	title = {Collaborative {Dynamic} {3D} {Scene} {Graphs} for {Automated} {Driving}},
	copyright = {https://doi.org/10.15223/policy-029},
	url = {https://ieeexplore.ieee.org/document/10610112/},
	doi = {10.1109/icra57147.2024.10610112},
	urldate = {2025-07-07},
	booktitle = {2024 {IEEE} {International} {Conference} on {Robotics} and {Automation} ({ICRA})},
	publisher = {IEEE},
	author = {Greve, Elias and Büchner, Martin and Vödisch, Niclas and Burgard, Wolfram and Valada, Abhinav},
	month = may,
	year = {2024},
	pages = {11118--11124},
}

@article{behrens_lost_2025,
	title = {Lost \& {Found}: {Tracking} {Changes} {From} {Egocentric} {Observations} in {3D} {Dynamic} {Scene} {Graphs}},
	volume = {10},
	copyright = {https://ieeexplore.ieee.org/Xplorehelp/downloads/license-information/IEEE.html},
	issn = {2377-3766, 2377-3774},
	shorttitle = {Lost \& {Found}},
	url = {https://ieeexplore.ieee.org/document/10897827/},
	doi = {10.1109/lra.2025.3544518},
	number = {4},
	urldate = {2025-07-07},
	journal = {IEEE Robotics and Automation Letters},
	publisher = {Institute of Electrical and Electronics Engineers (IEEE)},
	author = {Behrens, Tjark and Zurbrügg, René and Pollefeys, Marc and Bauer, Zuria and Blum, Hermann},
	month = apr,
	year = {2025},
	pages = {3739--3746},
}

@misc{henein_dynamic_2020,
	title = {Dynamic {SLAM}: {The} {Need} {For} {Speed}},
	copyright = {arXiv.org perpetual, non-exclusive license},
	shorttitle = {Dynamic {SLAM}},
	url = {https://arxiv.org/abs/2002.08584},
	doi = {10.48550/ARXIV.2002.08584},
	urldate = {2025-05-28},
	publisher = {arXiv},
	author = {Henein, Mina and Zhang, Jun and Mahony, Robert and Ila, Viorela},
	year = {2020},
	note = {Version Number: 2},
}

@article{patel_proactive_2022,
	title = {Proactive {Robot} {Assistance} via {Spatio}-{Temporal} {Object} {Modeling}},
	copyright = {arXiv.org perpetual, non-exclusive license},
	url = {https://arxiv.org/abs/2211.15501},
	doi = {10.48550/ARXIV.2211.15501},
	urldate = {2025-05-27},
	publisher = {arXiv},
	author = {Patel, Maithili and Chernova, Sonia},
	year = {2022},
	note = {Version Number: 1},
}

@article{zaech_learnable_2022,
	title = {Learnable {Online} {Graph} {Representations} for {3D} {Multi}-{Object} {Tracking}},
	volume = {7},
	copyright = {https://ieeexplore.ieee.org/Xplorehelp/downloads/license-information/IEEE.html},
	issn = {2377-3766, 2377-3774},
	url = {https://ieeexplore.ieee.org/document/9691806/},
	doi = {10.1109/LRA.2022.3145952},
	number = {2},
	urldate = {2025-05-27},
	journal = {IEEE Robotics and Automation Letters},
	author = {Zaech, Jan-Nico and Liniger, Alexander and Dai, Dengxin and Danelljan, Martin and Van Gool, Luc},
	month = apr,
	year = {2022},
	pages = {5103--5110},
}

@inproceedings{deeb_model-based_2019,
	address = {Belo Horizonte, Brazil},
	title = {Model-based {Dynamic} {Pose} {Graph} {SLAM} in {Unstructured} {Dynamic} {Environments}},
	copyright = {https://ieeexplore.ieee.org/Xplorehelp/downloads/license-information/IEEE.html},
	isbn = {978-1-72812-467-4},
	url = {https://ieeexplore.ieee.org/document/8981632/},
	doi = {10.1109/ICAR46387.2019.8981632},
	urldate = {2025-05-27},
	booktitle = {2019 19th {International} {Conference} on {Advanced} {Robotics} ({ICAR})},
	publisher = {IEEE},
	author = {Deeb, A. and Seto, M. L. and Pan, Y.J.},
	month = dec,
	year = {2019},
	pages = {123--128},
}

@article{yan_dynamic_2025,
	title = {Dynamic {Open}-{Vocabulary} {3D} {Scene} {Graphs} for {Long}-{Term} {Language}-{Guided} {Mobile} {Manipulation}},
	volume = {10},
	copyright = {https://ieeexplore.ieee.org/Xplorehelp/downloads/license-information/IEEE.html},
	issn = {2377-3766, 2377-3774},
	url = {https://ieeexplore.ieee.org/document/10909193/},
	doi = {10.1109/LRA.2025.3547643},
	number = {5},
	urldate = {2025-05-27},
	journal = {IEEE Robotics and Automation Letters},
	author = {Yan, Zhijie and Li, Shufei and Wang, Zuoxu and Wu, Lixiu and Wang, Han and Zhu, Jun and Chen, Lijiang and Liu, Jihong},
	month = may,
	year = {2025},
	pages = {4252--4259},
}

@article{deeb_piecewise-deterministic_2022,
	title = {Piecewise-deterministic {Quasi}-static {Pose} {Graph} {SLAM} in {Unstructured} {Dynamic} {Environments}},
	volume = {106},
	issn = {0921-0296, 1573-0409},
	url = {https://link.springer.com/10.1007/s10846-022-01739-5},
	doi = {10.1007/s10846-022-01739-5},
	language = {en},
	number = {2},
	urldate = {2025-05-27},
	journal = {Journal of Intelligent \& Robotic Systems},
	author = {Deeb, Amy and Seto, Mae and Pan, Ya-Jun},
	month = oct,
	year = {2022},
	pages = {34},
}

@misc{olivastri_multi-modal_2024,
	title = {Multi-{Modal} {3D} {Scene} {Graph} {Updater} for {Shared} and {Dynamic} {Environments}},
	copyright = {arXiv.org perpetual, non-exclusive license},
	url = {https://arxiv.org/abs/2411.02938},
	doi = {10.48550/ARXIV.2411.02938},
	urldate = {2025-05-27},
	publisher = {arXiv},
	author = {Olivastri, Emilio and Francis, Jonathan and Pretto, Alberto and Sünderhauf, Niko and Rana, Krishan},
	year = {2024},
	note = {Version Number: 1},
}

@article{wang_semantic_2023,
	title = {A {Semantic} {Topology} {Graph} to {Detect} {Re}-{Localization} and {Loop} {Closure} of the {Visual} {Simultaneous} {Localization} and {Mapping} {System} in a {Dynamic} {Environment}},
	volume = {23},
	copyright = {https://creativecommons.org/licenses/by/4.0/},
	issn = {1424-8220},
	url = {https://www.mdpi.com/1424-8220/23/20/8445},
	doi = {10.3390/s23208445},
	language = {en},
	number = {20},
	urldate = {2025-05-26},
	journal = {Sensors},
	author = {Wang, Yang and Zhang, Yi and Hu, Lihe and Wang, Wei and Ge, Gengyu and Tan, Shuyi},
	month = oct,
	year = {2023},
	pages = {8445},
}

@inproceedings{lingzhu_xiang_robust_2015,
	address = {Hamburg, Germany},
	title = {Robust graph {SLAM} in dynamic environments with moving landmarks},
	isbn = {978-1-4799-9994-1},
	url = {http://ieeexplore.ieee.org/document/7353723/},
	doi = {10.1109/IROS.2015.7353723},
	urldate = {2025-05-26},
	booktitle = {2015 {IEEE}/{RSJ} {International} {Conference} on {Intelligent} {Robots} and {Systems} ({IROS})},
	publisher = {IEEE},
	author = {{Lingzhu Xiang} and {Zhile Ren} and {Mengrui Ni} and Jenkins, Odest Chadwicke},
	month = sep,
	year = {2015},
	pages = {2543--2549},
}

@article{gorlo_long-term_2024,
	title = {Long-{Term} {Human} {Trajectory} {Prediction} {Using} {3D} {Dynamic} {Scene} {Graphs}},
	volume = {9},
	copyright = {https://ieeexplore.ieee.org/Xplorehelp/downloads/license-information/IEEE.html},
	issn = {2377-3766, 2377-3774},
	url = {https://ieeexplore.ieee.org/document/10720207/},
	doi = {10.1109/LRA.2024.3482169},
	number = {12},
	urldate = {2025-05-21},
	journal = {IEEE Robotics and Automation Letters},
	author = {Gorlo, Nicolas and Schmid, Lukas and Carlone, Luca},
	month = dec,
	year = {2024},
	pages = {10978--10985},
}

@inproceedings{kurenkov_modeling_2023,
	series = {Proceedings of {Machine} {Learning} {Research}},
	title = {Modeling {Dynamic} {Environments} with {Scene} {Graph} {Memory}},
	volume = {202},
	url = {https://proceedings.mlr.press/v202/kurenkov23a.html},
	booktitle = {Proceedings of the 40th {International} {Conference} on {Machine} {Learning}},
	publisher = {PMLR},
	author = {Kurenkov, Andrey and Lingelbach, Michael and Agarwal, Tanmay and Jin, Emily and Li, Chengshu and Zhang, Ruohan and Fei-Fei, Li and Wu, Jiajun and Savarese, Silvio and Martín-Martín, Roberto},
	editor = {Krause, Andreas and Brunskill, Emma and Cho, Kyunghyun and Engelhardt, Barbara and Sabato, Sivan and Scarlett, Jonathan},
	month = jul,
	year = {2023},
	pages = {17976--17993},
}

@inproceedings{walcott-bryant_dynamic_2012,
	address = {Vilamoura-Algarve, Portugal},
	title = {Dynamic pose graph {SLAM}: {Long}-term mapping in low dynamic environments},
	isbn = {978-1-4673-1736-8 978-1-4673-1737-5 978-1-4673-1735-1},
	shorttitle = {Dynamic pose graph {SLAM}},
	url = {http://ieeexplore.ieee.org/document/6385561/},
	doi = {10.1109/IROS.2012.6385561},
	urldate = {2025-05-19},
	booktitle = {2012 {IEEE}/{RSJ} {International} {Conference} on {Intelligent} {Robots} and {Systems}},
	publisher = {IEEE},
	author = {Walcott-Bryant, Aisha and Kaess, Michael and Johannsson, Hordur and Leonard, John J.},
	month = oct,
	year = {2012},
	pages = {1871--1878},
}

@misc{linok_dygenc_2025,
	title = {{DyGEnc}: {Encoding} a {Sequence} of {Textual} {Scene} {Graphs} to {Reason} and {Answer} {Questions} in {Dynamic} {Scenes}},
	shorttitle = {{DyGEnc}},
	url = {http://arxiv.org/abs/2505.03581},
	doi = {10.48550/arXiv.2505.03581},
	urldate = {2025-05-19},
	publisher = {arXiv},
	author = {Linok, Sergey and Semenov, Vadim and Trunova, Anastasia and Bulichev, Oleg and Yudin, Dmitry},
	month = may,
	year = {2025},
	note = {arXiv:2505.03581 [cs]},
}

@phdthesis{brink_using_2016,
	title = {Using probabilistic graphical models to detect dynamic objects for mobile robots},
	url = {http://hdl.handle.net/10019.1/100224},
	school = {Stellenbosch University},
	author = {Brink, Daniek},
	year = {2016},
}

@misc{saucedo_estimating_2025,
	title = {Estimating {Commonsense} {Scene} {Composition} on {Belief} {Scene} {Graphs}},
	url = {http://arxiv.org/abs/2505.02405},
	doi = {10.48550/arXiv.2505.02405},
	urldate = {2025-05-16},
	publisher = {arXiv},
	author = {Saucedo, Mario A. V. and Viswanathan, Vignesh Kottayam and Kanellakis, Christoforos and Nikolakopoulos, George},
	month = may,
	year = {2025},
	note = {arXiv:2505.02405 [cs]},
}

@inproceedings{vega_torres_bim-slam_2023,
	title = {{BIM}-{SLAM}: {Integrating} {BIM} {Models} in {Multi}-session {SLAM} for {Lifelong} {Mapping} using {3D} {LiDAR}},
	isbn = {978-0-6458322-0-4},
	issn = {2413-5844},
	doi = {10.22260/ISARC2023/0070},
	booktitle = {Proceedings of the 40th {International} {Symposium} on {Automation} and {Robotics} in {Construction} ({IAARC})},
	author = {Vega Torres, Miguel Arturo and Braun, Alex and Borrmann, André},
	month = jul,
	year = {2023},
	pages = {521--528},
}

@inproceedings{millan-romera_learning_2024,
	title = {Learning {High}-level {Semantic}-{Relational} {Concepts} for {SLAM}},
	booktitle = {2024 {IEEE}/{RSJ} {International} {Conference} on {Intelligent} {Robots} and {Systems} ({IROS})},
	publisher = {IEEE},
	author = {Millan-Romera, Jose Andres and Bavle, Hriday and Shaheer, Muhammad and Oswald, Martin R and Voos, Holger and Sanchez-Lopez, Jose Luis},
	year = {2024},
	pages = {9803--9810},
}

@inproceedings{schmid_khronos_2024,
	address = {Delft, Netherlands},
	title = {Khronos: {A} {Unified} {Approach} for {Spatio}-{Temporal} {Metric}-{Semantic} {SLAM} in {Dynamic} {Environments}},
	doi = {10.15607/RSS.2024.XX.081},
	booktitle = {Proc. of {Robotics}: {Science} and {Systems} ({RSS})},
	author = {Schmid, Lukas and Abate, Marcus and Chang, Yun and Carlone, Luca},
	month = jul,
	year = {2024},
}

@article{shaheer_tightly_2025,
	title = {Tightly {Coupled} {SLAM} with {Imprecise} {Architectural} {Plans}},
	journal = {arXiv preprint arXiv:2408.01737v2},
	author = {Shaheer, Muhammad and Millan-Romera, Jose Andres and Bavle, Hriday and Giberna, Marco and Sanchez-Lopez, Jose Luis and Civera, Javier and Voos, Holger},
	year = {2025},
}

@article{bavle_s-graphs_2025,
	title = {S-{Graphs} 2.0–{A} {Hierarchical}-{Semantic} {Optimization} and {Loop} {Closure} for {SLAM}},
	journal = {arXiv preprint arXiv:2502.18044},
	author = {Bavle, Hriday and Sanchez-Lopez, Jose Luis and Shaheer, Muhammad and Civera, Javier and Voos, Holger},
	year = {2025},
}

@article{fernandez-cortizas_multi_2024,
	title = {Multi {S}-{Graphs}: {An} {Efficient} {Distributed} {Semantic}-{Relational} {Collaborative} {SLAM}},
	volume = {9},
	issn = {2377-3766},
	shorttitle = {Multi {S}-{Graphs}},
	url = {https://ieeexplore.ieee.org/document/10529513/?arnumber=10529513},
	doi = {10.1109/LRA.2024.3399997},
	number = {6},
	urldate = {2025-01-14},
	journal = {IEEE Robotics and Automation Letters},
	author = {Fernandez-Cortizas, Miguel and Bavle, Hriday and Perez-Saura, David and Sanchez-Lopez, Jose Luis and Campoy, Pascual and Voos, Holger},
	month = jun,
	year = {2024},
	note = {Conference Name: IEEE Robotics and Automation Letters},
	pages = {6004--6011},
}

@inproceedings{rosinol_3d_2020,
	title = {{3D} {Dynamic} {Scene} {Graphs}: {Actionable} {Spatial} {Perception} with {Places}, {Objects}, and {Humans}},
	isbn = {978-0-9923747-6-1},
	shorttitle = {{3D} {Dynamic} {Scene} {Graphs}},
	url = {http://www.roboticsproceedings.org/rss16/p079.pdf},
	doi = {10.15607/RSS.2020.XVI.079},
	language = {en},
	urldate = {2024-08-01},
	booktitle = {Robotics: {Science} and {Systems} {XVI}},
	publisher = {Robotics: Science and Systems Foundation},
	author = {Rosinol, Antoni and Gupta, Arjun and Abate, Marcus and Shi, Jingnan and Carlone, Luca},
	month = jul,
	year = {2020},
}

@article{kim_3-d_2020,
	title = {3-{D} {Scene} {Graph}: {A} {Sparse} and {Semantic} {Representation} of {Physical} {Environments} for {Intelligent} {Agents}},
	volume = {50},
	issn = {2168-2275},
	shorttitle = {3-{D} {Scene} {Graph}},
	url = {https://ieeexplore.ieee.org/document/8796400},
	doi = {10.1109/TCYB.2019.2931042},
	number = {12},
	urldate = {2024-08-01},
	journal = {IEEE Transactions on Cybernetics},
	author = {Kim, Ue-Hwan and Park, Jin-Man and Song, Taek-jin and Kim, Jong-Hwan},
	month = dec,
	year = {2020},
	note = {Conference Name: IEEE Transactions on Cybernetics},
	pages = {4921--4933},
}

@article{bavle_situational_2022,
	title = {Situational {Graphs} for {Robot} {Navigation} in {Structured} {Indoor} {Environments}},
	volume = {7},
	issn = {2377-3766},
	url = {https://ieeexplore.ieee.org/document/9826367},
	doi = {10.1109/LRA.2022.3189785},
	number = {4},
	urldate = {2024-08-01},
	journal = {IEEE Robotics and Automation Letters},
	author = {Bavle, Hriday and Sanchez-Lopez, Jose Luis and Shaheer, Muhammad and Civera, Javier and Voos, Holger},
	month = oct,
	year = {2022},
	note = {Conference Name: IEEE Robotics and Automation Letters},
	pages = {9107--9114},
}

@inproceedings{hughes_hydra_2022,
	title = {Hydra: {A} {Real}-time {Spatial} {Perception} {System} for {3D} {Scene} {Graph} {Construction} and {Optimization}},
	isbn = {978-0-9923747-8-5},
	shorttitle = {Hydra},
	url = {http://www.roboticsproceedings.org/rss18/p050.pdf},
	doi = {10.15607/RSS.2022.XVIII.050},
	language = {en},
	urldate = {2024-08-01},
	booktitle = {Robotics: {Science} and {Systems} {XVIII}},
	publisher = {Robotics: Science and Systems Foundation},
	author = {Hughes, Nathan and Chang, Yun and Carlone, Luca},
	month = jun,
	year = {2022},
}

@article{gomez_object-based_2020,
	title = {Object-{Based} {Pose} {Graph} for {Dynamic} {Indoor} {Environments}},
	volume = {5},
	copyright = {https://ieeexplore.ieee.org/Xplorehelp/downloads/license-information/IEEE.html},
	issn = {2377-3766, 2377-3774},
	url = {https://ieeexplore.ieee.org/document/9134440/},
	doi = {10.1109/LRA.2020.3007402},
	number = {4},
	urldate = {2024-07-08},
	journal = {IEEE Robotics and Automation Letters},
	author = {Gomez, Clara and Hernandez, Alejandra C. and Derner, Erik and Barber, Ramon and Babuska, Robert},
	month = oct,
	year = {2020},
	pages = {5401--5408},
}

@misc{matsuki_gaussian_2024,
	title = {Gaussian {Splatting} {SLAM}},
	url = {http://arxiv.org/abs/2312.06741},
	doi = {10.48550/arXiv.2312.06741},
	urldate = {2024-07-05},
	publisher = {arXiv},
	author = {Matsuki, Hidenobu and Murai, Riku and Kelly, Paul H. J. and Davison, Andrew J.},
	month = apr,
	year = {2024},
	note = {arXiv:2312.06741 [cs]},
}

@article{rosinol_kimera_2021,
	title = {Kimera: from {SLAM} to {Spatial} {Perception} with {3D} {Dynamic} {Scene} {Graphs}},
	copyright = {Creative Commons Attribution 4.0 International},
	shorttitle = {Kimera},
	url = {https://arxiv.org/abs/2101.06894},
	doi = {10.48550/ARXIV.2101.06894},
	urldate = {2024-03-12},
	publisher = {[object Object]},
	author = {Rosinol, Antoni and Violette, Andrew and Abate, Marcus and Hughes, Nathan and Chang, Yun and Shi, Jingnan and Gupta, Arjun and Carlone, Luca},
	year = {2021},
	note = {Version Number: 3},
}

@inproceedings{kummerle_g2o_2011,
	title = {G2o: {A} general framework for graph optimization},
	shorttitle = {G2o},
	doi = {10.1109/ICRA.2011.5979949},
	author = {Kümmerle, Rainer and Grisetti, Giorgio and Strasdat, Hauke and Konolige, Kurt and Burgard, Wolfram},
	month = jun,
	year = {2011},
	pages = {3607--3613},
}

@misc{kerbl_3d_2023,
	title = {{3D} {Gaussian} {Splatting} for {Real}-{Time} {Radiance} {Field} {Rendering}},
	url = {http://arxiv.org/abs/2308.04079},
	urldate = {2023-08-20},
	publisher = {arXiv},
	author = {Kerbl, Bernhard and Kopanas, Georgios and Leimkühler, Thomas and Drettakis, George},
	month = aug,
	year = {2023},
	note = {arXiv:2308.04079 [cs]},
}

@article{bescos_dynaslam_2018,
	title = {{DynaSLAM}: {Tracking}, {Mapping}, and {Inpainting} in {Dynamic} {Scenes}},
	volume = {3},
	issn = {2377-3766, 2377-3774},
	shorttitle = {{DynaSLAM}},
	url = {https://ieeexplore.ieee.org/document/8421015/},
	doi = {10.1109/LRA.2018.2860039},
	language = {en},
	number = {4},
	urldate = {2023-03-01},
	journal = {IEEE Robotics and Automation Letters},
	author = {Bescos, Berta and Facil, Jose M. and Civera, Javier and Neira, Jose},
	month = oct,
	year = {2018},
	pages = {4076--4083},
}

@article{bescos_dynaslam_2021,
	title = {{DynaSLAM} {II}: {Tightly}-{Coupled} {Multi}-{Object} {Tracking} and {SLAM}},
	volume = {6},
	issn = {2377-3766, 2377-3774},
	shorttitle = {{DynaSLAM} {II}},
	url = {https://ieeexplore.ieee.org/document/9385844/},
	doi = {10.1109/LRA.2021.3068640},
	language = {en},
	number = {3},
	urldate = {2023-03-01},
	journal = {IEEE Robotics and Automation Letters},
	author = {Bescos, Berta and Campos, Carlos and Tardos, Juan D. and Neira, Jose},
	month = jul,
	year = {2021},
	pages = {5191--5198},
}

@inproceedings{armeni_3d_2016,
	title = {{3D} {Semantic} {Parsing} of {Large}-{Scale} {Indoor} {Spaces}},
	issn = {1063-6919},
	doi = {10.1109/CVPR.2016.170},
	booktitle = {2016 {IEEE} {Conference} on {Computer} {Vision} and {Pattern} {Recognition} ({CVPR})},
	author = {Armeni, Iro and Sener, Ozan and Zamir, Amir R. and Jiang, Helen and Brilakis, Ioannis and Fischer, Martin and Savarese, Silvio},
	month = jun,
	year = {2016},
	pages = {1534--1543},
}

@article{wen_semantic_2021,
	title = {Semantic visual {SLAM} in dynamic environment},
	volume = {45},
	issn = {1573-7527},
	url = {https://doi.org/10.1007/s10514-021-09979-4},
	doi = {10.1007/s10514-021-09979-4},
	language = {en},
	number = {4},
	urldate = {2022-08-22},
	journal = {Autonomous Robots},
	author = {Wen, Shuhuan and Li, Pengjiang and Zhao, Yongjie and Zhang, Hong and Sun, Fuchun and Wang, Zhe},
	month = may,
	year = {2021},
	pages = {493--504},
}

@article{chang_matterport3d_2017,
	title = {{Matterport3D}: {Learning} from {RGB}-{D} {Data} in {Indoor} {Environments}},
	shorttitle = {{Matterport3D}},
	url = {http://arxiv.org/abs/1709.06158},
	language = {en},
	urldate = {2022-04-29},
	journal = {arXiv:1709.06158 [cs]},
	author = {Chang, Angel and Dai, Angela and Funkhouser, Thomas and Halber, Maciej and Nießner, Matthias and Savva, Manolis and Song, Shuran and Zeng, Andy and Zhang, Yinda},
	month = sep,
	year = {2017},
	note = {arXiv: 1709.06158},
}

@inproceedings{wu_scenegraphfusion_2021,
	address = {Nashville, TN, USA},
	title = {{SceneGraphFusion}: {Incremental} {3D} {Scene} {Graph} {Prediction} from {RGB}-{D} {Sequences}},
	isbn = {978-1-66544-509-2},
	shorttitle = {{SceneGraphFusion}},
	url = {https://ieeexplore.ieee.org/document/9578559/},
	doi = {10.1109/CVPR46437.2021.00743},
	language = {en},
	urldate = {2022-04-20},
	booktitle = {2021 {IEEE}/{CVF} {Conference} on {Computer} {Vision} and {Pattern} {Recognition} ({CVPR})},
	publisher = {IEEE},
	author = {Wu, Shun-Cheng and Wald, Johanna and Tateno, Keisuke and Navab, Nassir and Tombari, Federico},
	month = jun,
	year = {2021},
	pages = {7511--7521},
}

@article{ravichandran_hierarchical_2021,
	title = {Hierarchical {Representations} and {Explicit} {Memory}: {Learning} {Effective} {Navigation} {Policies} on {3D} {Scene} {Graphs} using {Graph} {Neural} {Networks}},
	shorttitle = {Hierarchical {Representations} and {Explicit} {Memory}},
	url = {http://arxiv.org/abs/2108.01176},
	language = {en},
	urldate = {2022-03-02},
	journal = {arXiv:2108.01176 [cs]},
	author = {Ravichandran, Zachary and Peng, Lisa and Hughes, Nathan and Griffith, J. Daniel and Carlone, Luca},
	month = aug,
	year = {2021},
	note = {arXiv: 2108.01176},
}

@article{yang_cubeslam_2019,
	title = {{CubeSLAM}: {Monocular} 3-{D} {Object} {SLAM}},
	volume = {35},
	doi = {10.1109/TRO.2019.2909168},
	number = {4},
	journal = {IEEE Transactions on Robotics},
	author = {Yang, Shichao and Scherer, Sebastian},
	year = {2019},
	pages = {925--938},
}

@article{rosinol_kimera_2020,
	title = {Kimera: {An} {Open}-{Source} {Library} for {Real}-{Time} {Metric}-{Semantic} {Localization} and {Mapping}},
	issn = {9781728173955},
	doi = {10.1109/ICRA40945.2020.9196885},
	journal = {Proceedings - IEEE International Conference on Robotics and Automation},
	author = {Rosinol, Antoni and Abate, Marcus and Chang, Yun and Carlone, Luca},
	year = {2020},
	pages = {1689--1696},
}




\end{document}